\documentclass[accepted]{article}
\usepackage{icml2020}
\usepackage[utf8]{inputenc}
\usepackage{xcolor}
\usepackage{natbib}
\usepackage{amsmath,amsfonts}
\usepackage{graphicx}
\usepackage[utf8]{inputenc}
\usepackage{hyperref}       % hyperlinks
\usepackage{url}            % simple URL typesetting
\usepackage{booktabs}       % professional-quality tables
\usepackage{amsfonts}       % blackboard math symbols
\usepackage{nicefrac}       % compact symbols for 1/2, etc.
\usepackage{microtype}      % microtypography

\usepackage[ruled,vlined,algo2e]{algorithm2e}
 
\usepackage{graphicx}
\usepackage{booktabs} % for professional tables
\usepackage{amsmath}
\usepackage{amsthm}
\usepackage{float}
\usepackage{bm}
\usepackage{amssymb}
\usepackage{bbm}
\usepackage{color}
\usepackage{mathtools}
\usepackage[super]{nth}
\usepackage{etoolbox}
\usepackage{adjustbox}
\usepackage{enumitem}
\usepackage{float}

\usepackage{subcaption}
\usepackage{bm}
\usepackage[capitalize]{cleveref}
\usepackage[group-separator={,},group-minimum-digits={3}]{siunitx}
\DeclareSIUnit[number-unit-product = ]\percent{\char`\%}
\usepackage{placeins}

\DeclarePairedDelimiter\inner{\langle}{\rangle}

\DeclarePairedDelimiter\floor{\lfloor}{\rfloor}
\DeclarePairedDelimiter\del{\lparen}{\rparen}
\DeclarePairedDelimiter\set{\lbrace}{\rbrace}
\DeclarePairedDelimiter\abs{|}{|}
\DeclarePairedDelimiter\norm{\|}{\|}
\DeclarePairedDelimiter\sbr{\lbrack}{\rbrack}

\newcommand{\RR}{\mathbb{R}}

\newcommand{\EE}{\mathbb{E}}

\DeclareMathOperator{\trace}{Tr}
\DeclareMathOperator{\var}{Var}
\DeclareMathOperator{\cov}{Cov}

\newtheorem{propo}{Proposition}[section]
\newtheorem{lemma}[propo]{Lemma}
\newtheorem{definition}[propo]{Definition}

\newtheorem{thm}{Theorem}

\newtheorem{proposition}[propo]{Proposition}

\usepackage[normalem]{ulem}

\begin{document}
\twocolumn[
\icmltitle{SPECTRE: Defending Against Backdoor Attacks Using 
Robust Statistics}



\icmlsetsymbol{equal}{*}

\begin{icmlauthorlist}
\icmlauthor{Jonathan Hayase}{uw}
\icmlauthor{Weihao Kong}{uw}
\icmlauthor{Raghav Somani}{uw}
\icmlauthor{Sewoong Oh}{uw}
\end{icmlauthorlist}

\icmlaffiliation{uw}{Paul G. Allen School of Computer Science \& Engineering, University of Washington, Seattle, United States}

\icmlcorrespondingauthor{Jonathan Hayase}{}
\icmlcorrespondingauthor{Sewoong Oh}{}

\icmlkeywords{Backdoor attack, robust covariance estimation, spectral signature, robust mean estimation, outlier detection}

\vskip 0.3in
]

\printAffiliationsAndNotice{} 
\begin{abstract}
    Modern machine learning increasingly requires training on a large collection of data from multiple sources, not all of which can be trusted.
    A particularly concerning scenario is when a small fraction of poisoned data changes the behavior of the trained model when triggered by an attacker-specified watermark. 
    Such a compromised model will be deployed unnoticed as the model is accurate otherwise. 
    There have been promising attempts to use the intermediate representations of such a model to separate corrupted examples from clean ones. 
    However, these defenses work only when a certain spectral signature of the poisoned examples is large enough for detection. 
    There is a wide range of attacks that cannot be protected against by the existing defenses. 
    %which require a much larger number of samples to be corrupted than is necessary for backdoor attack to be successful. 
    We propose a novel defense algorithm using robust  covariance estimation to amplify the spectral signature of corrupted data. 
    This defense provides a clean model, completely removing the backdoor, even in regimes where previous methods have no hope of detecting the poisoned \setcounter{footnote}{1} examples.\footnote{Code and pre-trained models are available at \url{https://github.com/SewoongLab/spectre-defense}.} 
    %this https URL.
\end{abstract}

\section{Introduction} 
% scenario for backdoor attack 
% spectral signature 
% downside: FIG 1 
% condition number large and spectral signature is weak, buried under the natural variations of the representation of uncorrupted data: actual spectral signature 
% contributions 

Large scale machine learning, such as federated learning \cite{kairouz2019advances}, requires training data collected from multiple sources.  
As not all sources can be trusted and sanity checking the data is expensive, this opens an opportunity for an adversary to inject poisoned data into the training set.
A particularly concerning scenario is the \textit{backdoor attack};
the attacker attempts to  embed a hidden backdoor to the trained model such that its prediction is maliciously changed when  activated by samples with an attacker-defined trigger.
As the model behavior on clean data is unchanged, such backdoored models may be deployed unnoticed. 

\begin{figure}[h]
    \centering
    \includegraphics[width=\linewidth, trim=0 1em 0 1em]{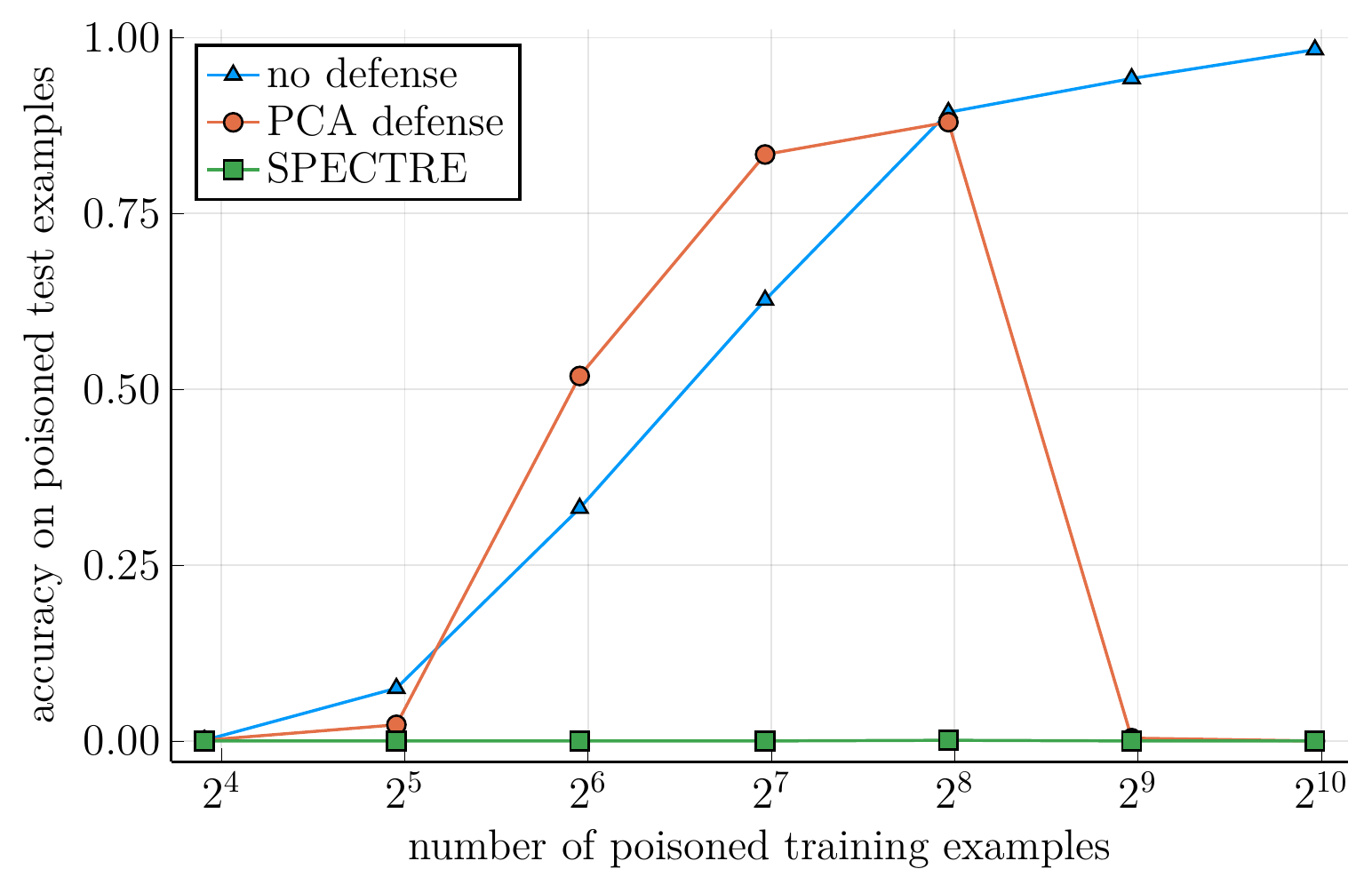}
    \caption{Under the pixel attack, the PCA defense fails to produce a clean model when the number of poisoned examples is between 64 and 256 (red circle). In fact, it removes clean data samples resulting in a model with higher accuracy on the poisoned test examples than when no defense was applied (blue triangle).
    SPECTRE %\cref{alg:detect} 
    produces clean models with  the backdoor completely removed in all regimes (green square).} 
    \label{fig:poison-acc-vs-eps} 
\end{figure}

Starting with the seminal work of \cite{gu2017badnets},
there has been an active line of work on designing backdoor attacks that use more stealth triggers \cite{chen2017targeted,liu2017trojaning,li2019invisible,liu2020reflection} or that can pass a human inspection \cite{turner2019label,zhao2020clean}. 
%Closest our work are the recent defenses leveraging the fact that poisoned data leaves spectral signatures in the intermediate layers of the trained model. 
%Our approach is motivated  by the PCA defense of \cite{tran2018spectral}and Clustering defense of \cite{chen2018detecting}, which we compare against in all experiments. 
Empirical evidence in these works suggest that a small fraction of poisoned data is sufficient to successfully create backdoors in trained neural networks.
For example, CIFAR-10 data has \num{5000} training examples for each of the ten classes.
When the pixel attack \cite{gu2017badnets} is launched with only 125 poisoned samples injected during training, the pixel attack  succeeds in planting a backdoor in the trained model, achieving an attack accuracy of \SI{63}{\percent} (shown in \cref{fig:poison-acc-vs-eps} in blue triangles). 

Recently, \citet{tran2018spectral} proposed, what we call, the PCA defense using the representations at the intermediate layers of such neural networks trained on a corrupted dataset. 
It is based on the observation that poisoned examples have special \textit{spectral signatures} that can be used to filter them out.  
Concretely, given the intermediate representations \(\set{\bm h_i}_{i=1}^n\) of all the training data, each sample is assigned an outlier score \(\tau_i = \abs{\inner{\bm h_i, \bm v_{h}}}\), which is  its magnitude in the top PCA direction \(\bm v_h\) of the representations.  
Those with high scores are removed from the training data, and a fresh model is trained on the filtered data. 
%When we remove $50\%$ more then the known number of poisoned samples following examples from \cite{},  

When there is a sufficient number of poisoned data (\(\geq 512\)) 
this PCA defense correctly detects poisoned samples and removes the backdoor completely;
attack accuracy drops down to \SI{0}{\percent} 
when we retrain a model after removing samples detected as poisoned  (shown in red circles). 
However, there is a wide gap between where the pixel attack becomes ineffective (around 64 poisoned examples)   and where the PCA defense stops working (around 256 poisoned samples), in this example. 

\begin{figure*}[t]
    \centering
    \begin{subfigure}[t]{0.33\textwidth}
        \centering
        \includegraphics[width=\textwidth, trim=0 1em 0 1em]{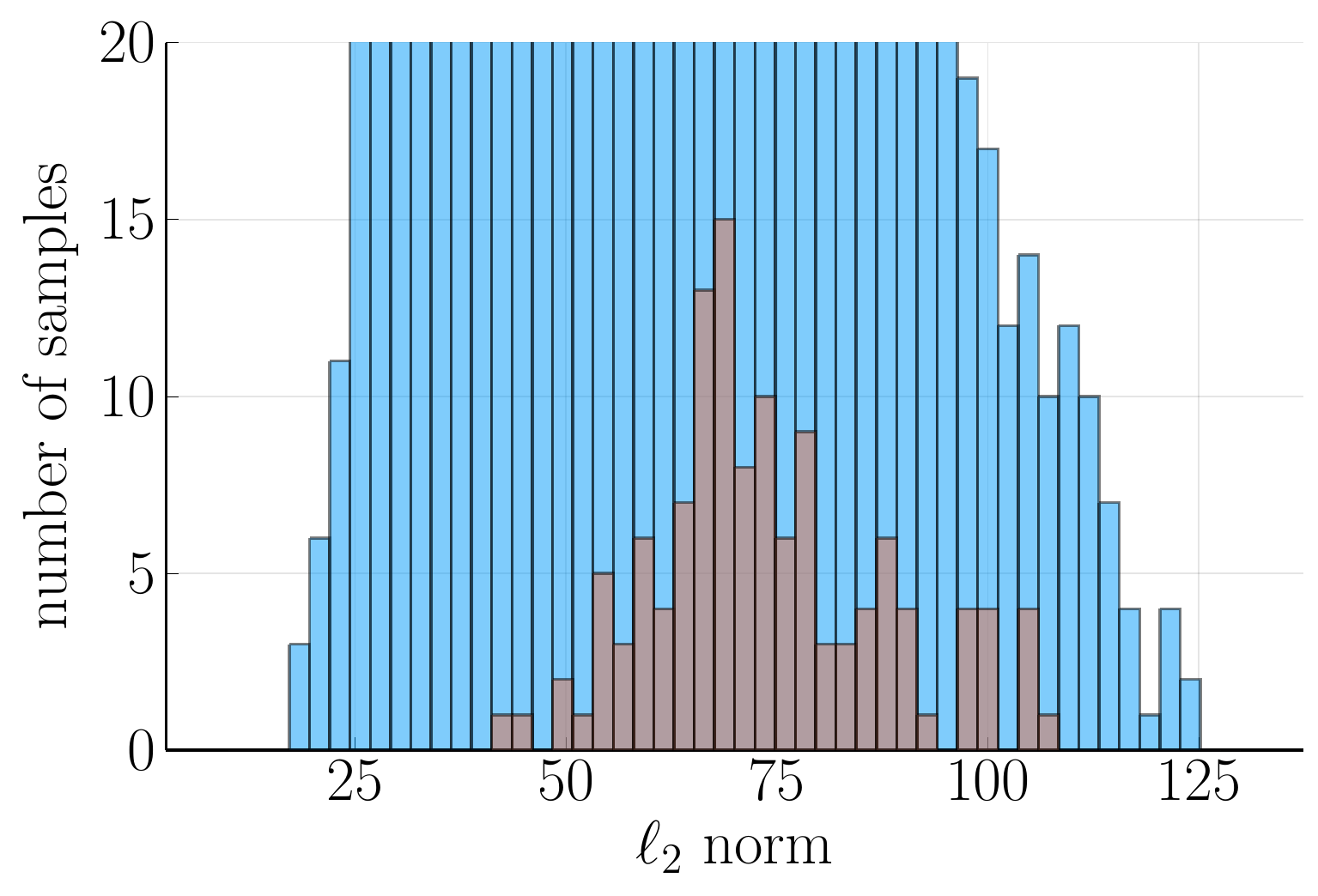}
        \caption{\(\norm{\bm x}_2\)}
        \label{fig:img-histograms-l2}
    \end{subfigure}%
    \begin{subfigure}[t]{0.33\textwidth}
        \centering
        \includegraphics[width=\textwidth, trim=0 1em 0 1em]{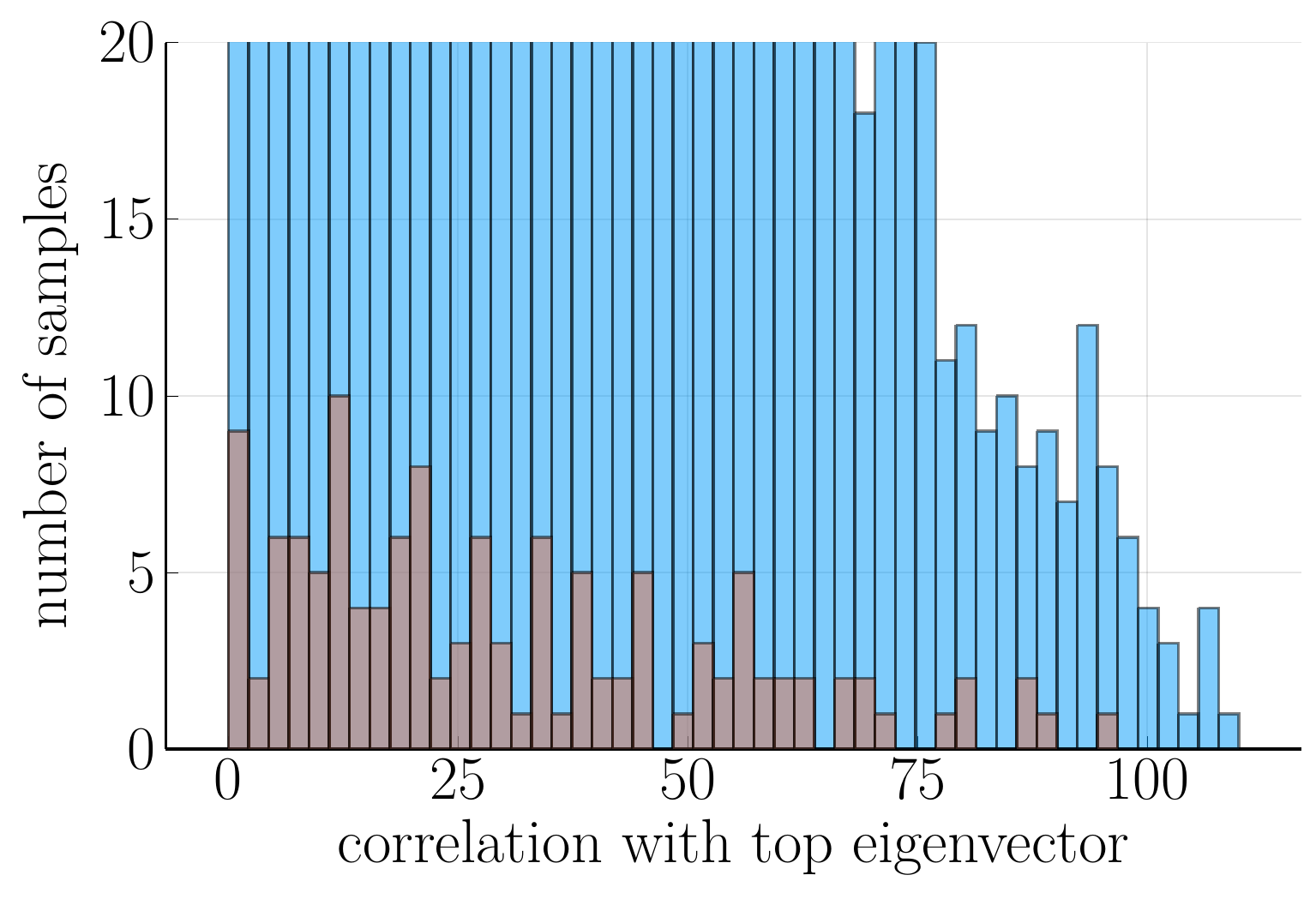}
        \caption{\(\abs{\inner{\bm x, \bm v_{x}}}\)}
        \label{fig:img-histograms-topeig}
    \end{subfigure}%
    \begin{subfigure}[t]{0.33\textwidth}
        \centering
        \includegraphics[width=\textwidth, trim=0 1em 0 1em]{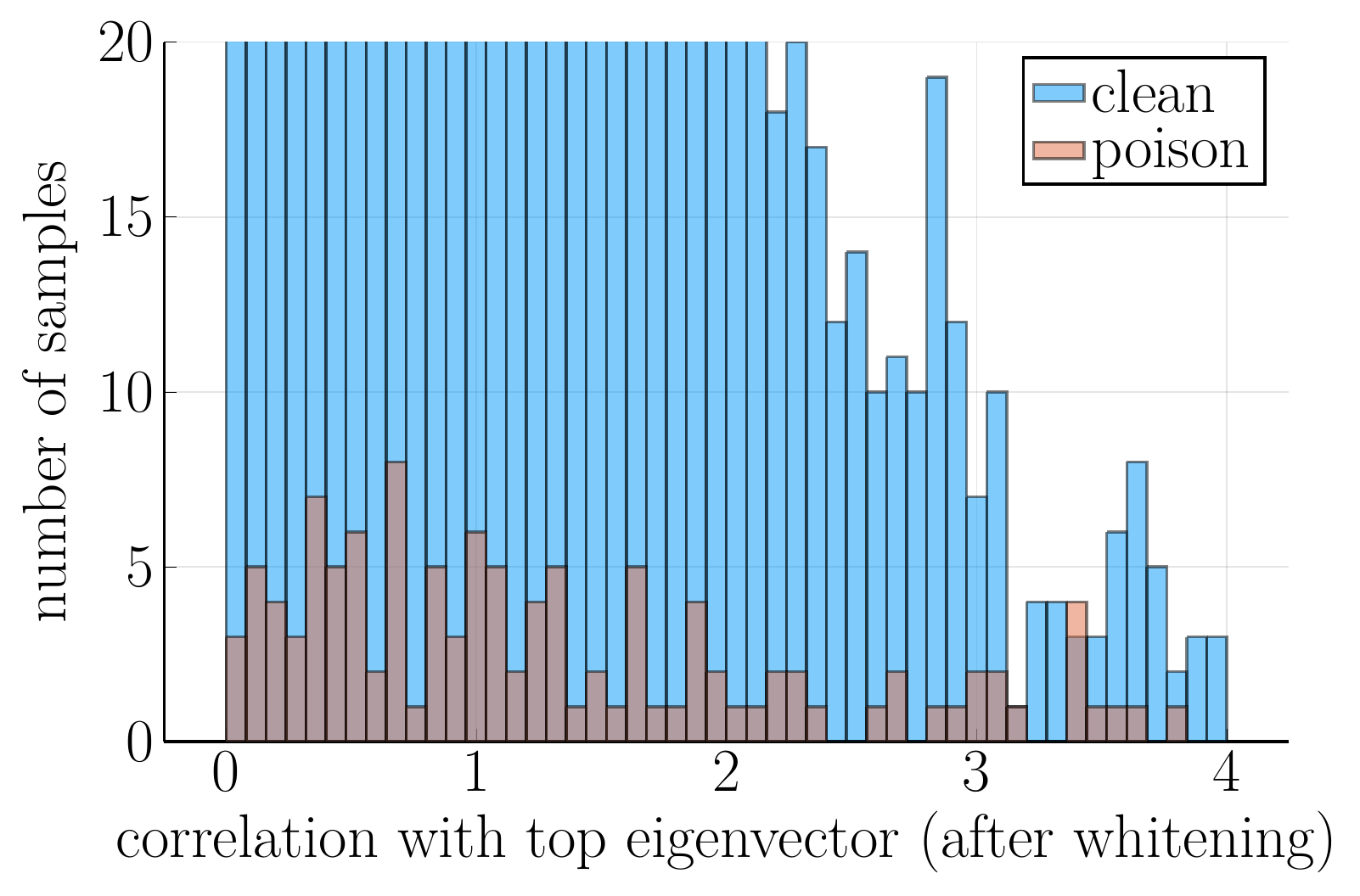}
        \caption{\(\abs{\inner{\widehat{\Sigma}_{x}^{-1/2}\bm x, \bm v_{x}'}}\)}
        \label{fig:img-histograms-white-topeig}
    \end{subfigure}
    \begin{subfigure}[t]{0.33\textwidth}
        \centering
        \includegraphics[width=\textwidth, trim=0 1em 0 0em]{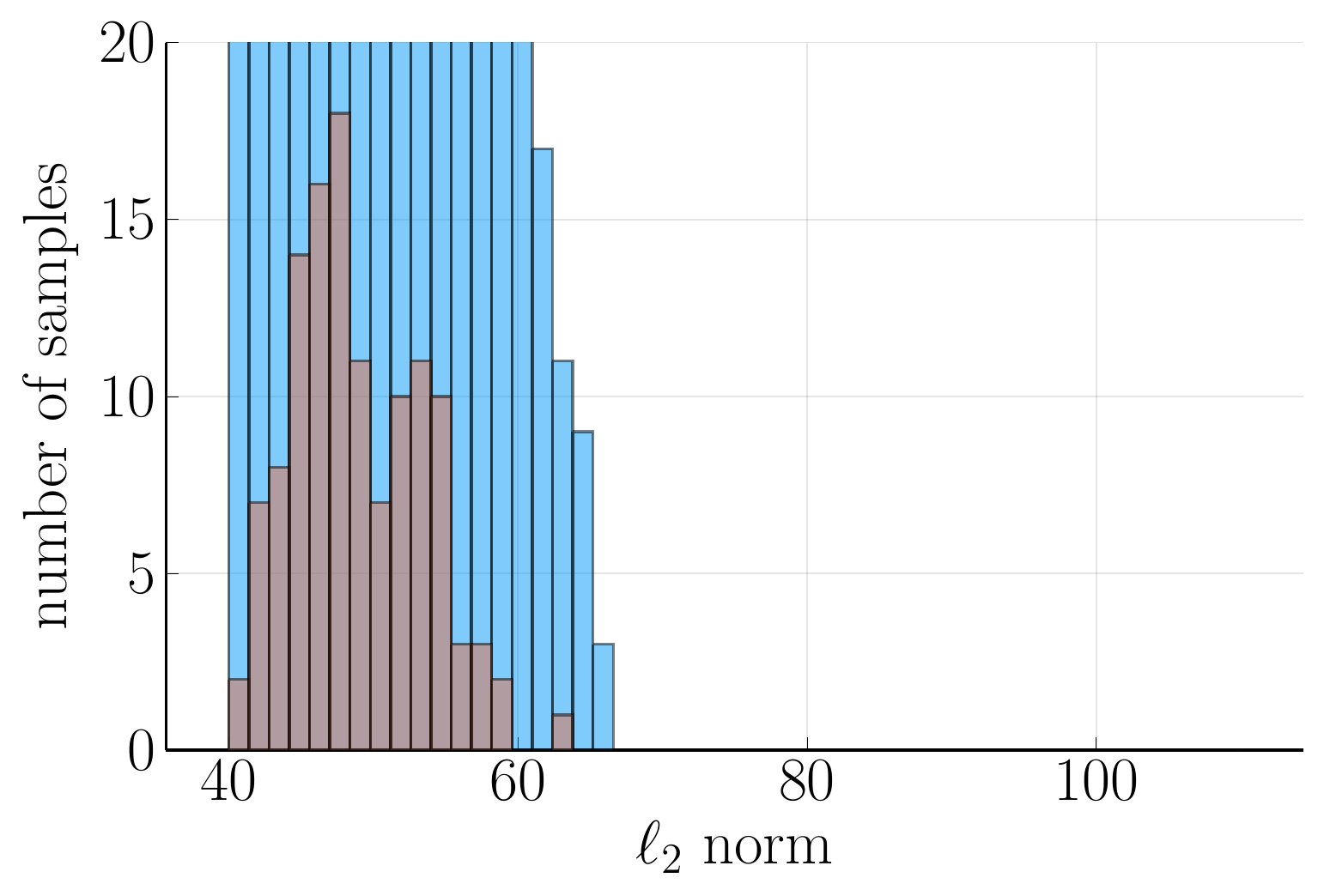}
        \caption{\(\norm{\bm h}_2\)}
        \label{fig:rep-histograms-l2}
    \end{subfigure}%
    \begin{subfigure}[t]{0.33\textwidth}
        \centering
        \includegraphics[width=\textwidth, trim=0 1em 0 0em]{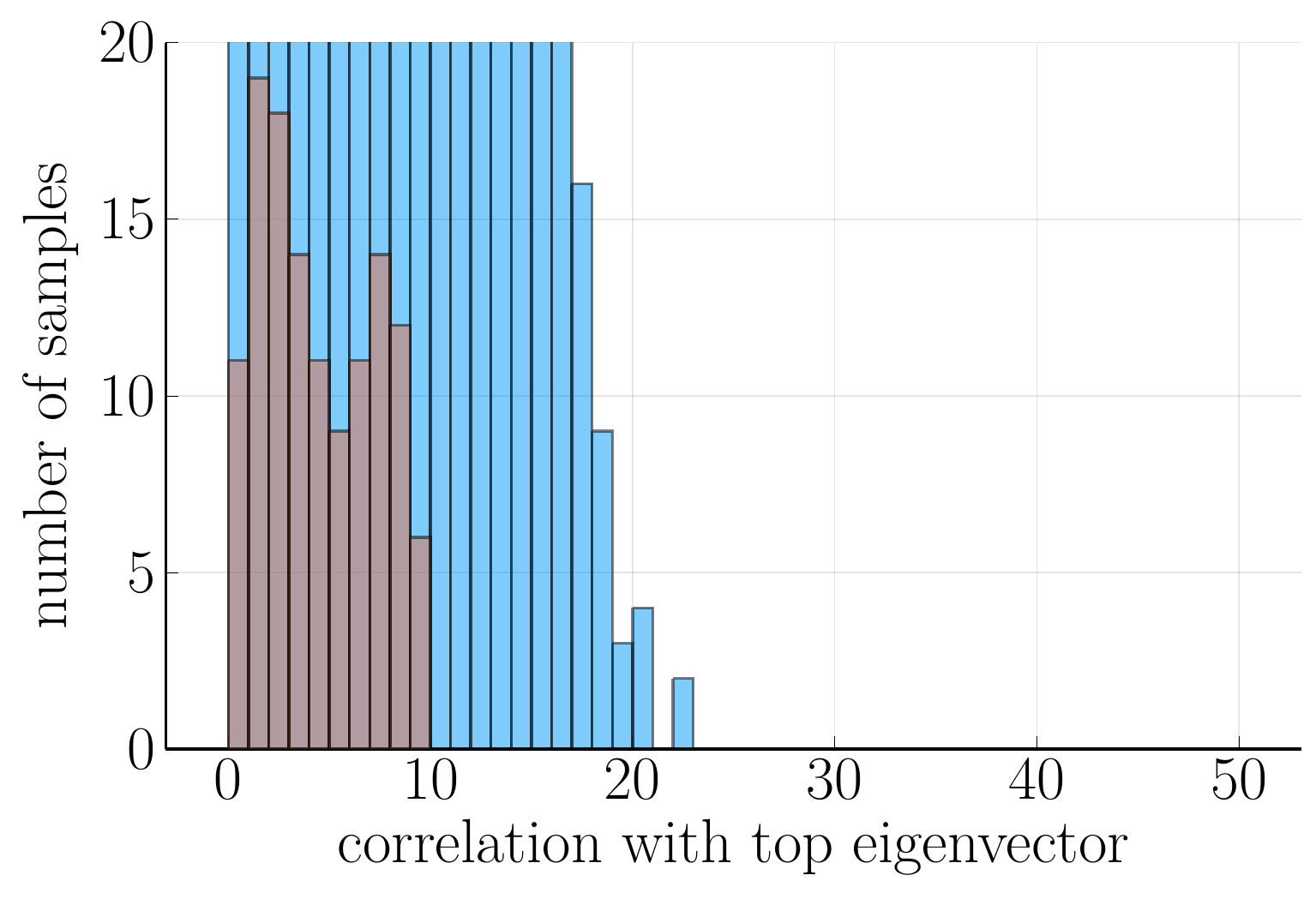}
        \caption{\(\abs{\inner{\bm h, \bm v_{h}}}\)}
        \label{fig:rep-histograms-topeig}
    \end{subfigure}%
    \begin{subfigure}[t]{0.3\textwidth}
        \centering
        \includegraphics[width=\textwidth, trim=0 1em 0 0em]{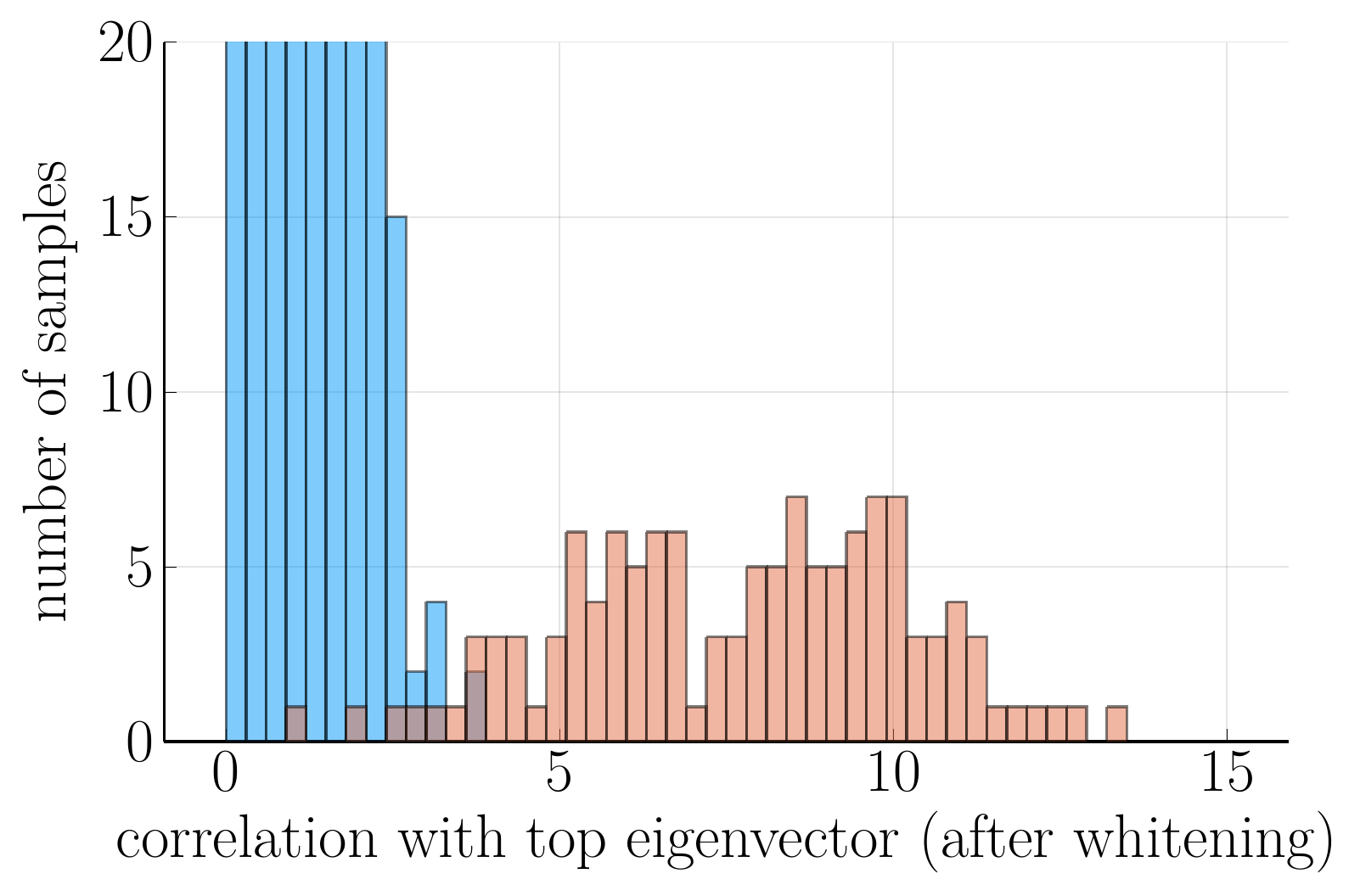}
        \caption{\(\abs{\inner{\widehat{\Sigma}_{h}^{-1/2}\bm h, \bm v_{h}'}}\)}
        \label{fig:rep-histograms-white-topeig} 
    \end{subfigure}
    \caption{Plots of the \num{5000} clean training examples and 125 poisoned examples bearing the target label under the 3-way pixel attack. 
    \cref{fig:img-histograms-l2,fig:rep-histograms-l2} show  the \(\ell_2\) norm of the images and representations respectively. \cref{fig:img-histograms-topeig,fig:rep-histograms-topeig} show the absolute inner product of the images and representations respectively with the top eigenvectors \(\bm v_x\) and \(\bm v_h\) of their covariances. 
    \cref{fig:img-histograms-white-topeig,fig:rep-histograms-white-topeig} show the absolute inner product of the \textit{robustly whitened} images and representations respectively with the top eigenvectors \(\bm v'_x\) and \(\bm v'_h\) of the covariances of the whitened data. \cref{fig:rep-histograms-white-topeig} shows how robust whitening amplifies the spectral signature of the poisoned samples and separates them out along the direction of top principal components.
    \label{fig:rep-histograms}}
\end{figure*}

\medskip\noindent{\bf Contributions.} 
%We first present an \(m\)-way attack which diversifies an existing attack (like the pixel attack) to hide the spectral signatures, evading the PCA defense for  a wider range of attacks. 
We introduce  
SPECTRE (Spectral Poison ExCision Through Robust Estimation), 
a novel defense for general backdoor attacks. 
The key insight, illustrated in \cref{fig:rep-histograms}, is that we can significantly amplify the spectral signature of the poisoned data by 
(i) estimating the mean and the covariance of the clean data using robust statistical estimators
and (ii) whitening the combined data with the estimated statistics. 
The resulting 
 top PCA directions are  well-aligned with the subspace that separates the poisoned samples from the clean ones (illustrated in \cref{fig:rep-histograms}). 
%combine the techniques of robust statistical estimation and quantum entropy outlier detection.  Robust  estimation  \cite{diakonikolas2019robust} allows us to  estimate the mean and the covariance of the representations of the clean data.  
However, detecting those poisoned examples can still be challenging as the  distribution of the (whitened) representations can vary  widely depending on the types and strengths of the attacks. 
To adapt to such profile of the representations, we propose a variation  
 of recently introduced QUantum Entropy (QUE) outlier scoring. 
 We show in \cref{sec:exp} that SPECTRE is able to eliminating the backdoor (e.g., shown in green squares in \cref{fig:poison-acc-vs-eps}) under a broad range of attacks, significantly improving upon the state-of-the-art baselines. We show that every component of SPECTRE is crucial in achieving this performance gain with ablation study in \cref{sec:ablation}.  

% ---------------------------------
\subsection{Related work}

%Machine learning is commonly used in security-critical settings. 
We focus on training-time attacks and  refer readers to \cite{madry2017towards,ilyas2019adversarial} for survey on inference-time attacks.

\medskip \noindent 
\textbf{Data poisoning attacks and defenses.} 
Data poisoning refers to attacks that insert poisoned examples into the training data. There are two types depending on the goal:  reducing model quality or creating a backdoor. 
Model quality attacks have been studied in 
feature selection \cite{xiao2015feature}, 
PCA \cite{rubinstein2009antidote},  
neural networks \cite{yang2017generative}, 
general models \cite{mozaffari2014systematic}, 
and general function classes \cite{kearns1993learning}. 
These  attacks have been successfully launched in deployed systems, as shown in \cite{newsome2006paragraph,laskov2014practical,biggio2014poisoning,wang2020attack}.

\medskip\noindent{\bf Backdoor attacks.}
Backdoor attacks create backdoors in trained models that change the model's prediction to an attacker-specific target label, when the sample has a specific attacker-chosen trigger. 
The most common attack is to embed triggers in a subset of training samples from a source label and change the label to the target label. 
\cite{gu2017badnets} first demonstrated that stamping an image with a small pattern can successfully create a backdoor.  
To design \textit{triggers} that can pass a human inspection on the image \(\bm x\), subsequent work 
mixed a pattern with the features \cite{chen2017targeted}, 
used periodic patterns to exploit convolutional layers \cite{zhong2020backdoor},
used intermediate layers of a neural network  \cite{liu2017trojaning},
minimized \(\ell_2\) norm of the perturbation \cite{zhong2020backdoor},
used perceptual similarity scores \cite{li2019invisible},   
applied reflection to the image as the trigger \cite{liu2020reflection}, 
and leveraged downscaling pre-processing step common in image classification tasks \cite{quiring2020backdooring}. 
However, these approaches share a weakness that a human inspecting both the image \(\bm x\) and the label \(y\) can easily detect a poisoned example, as it is perceived to be mislabelled as target \(y\).
\cite{turner2019label,zhao2020clean} propose embedding triggers in images that interpolate between the source and target labels. 
This can pass as being correctly labelled with the target label, while successfully creating backdoors.  %We use popular backdoor attacks from \cite{}, \cite{}, and \cite{} to show that we can xxx. 
     
\medskip\noindent{\bf Defenses against backdoor attacks.} 
As the defender is not assumed to have clean validation data, several approaches do not apply to our setting. 
Defenses using outlier detection require clean validation data \cite{liang2017enhancing,lee2018simple,steinhardt2017certified,turner2019label}. 
\cite{liu2018fine} requires clean data to retrain  a poisoned model to make it forget the backdoor. 
\cite{kolouri2020universal} requires a model trained on clean data to design a litmus test that detects poisoned models. 

Some other defenses \cite{wang2019neural,awasthi2020adversarial,wang2020certifying,weber2020rab,chou2018sentinet} rely on triggers having a small norm, and are known to fail on attacks with large perturbations. % as shown in  \cite{gao2019strip}. 
Neural Cleanse \cite{wang2019neural} finds perturbations that change the label of a training sample. 
The smallest such perturbation is declared as the trigger. %Once we have the trigger,  we can easily detect poisoned samples.  
%This is computationally demanding as the minimal perturbation needs to be calculated for all pairs of source and target labels.  
Randomized smoothing proposed in 
\cite{wang2020certifying,weber2020rab} ensures that all bounded perturbations are consistently labelled, forcing clean image and its poisoned version to have the same label. 

SentiNet \cite{chou2018sentinet} uses saliency maps to detect triggers corresponding to small connected regions of high salience over multiple images.
Other types of defenses protect against model quality attacks, including 
outlier detection without clean data \cite{sun2019can,steinhardt2017certified,blanchard2017machine,pillutla2019robust} 
and Byzantine-tolerant distributed learning approaches \cite{blanchard2017machine,alistarh2018byzantine,chen2018draco}.     

%\cite{gao2019strip} proposes STRIP that mixes each training sample with multiple other samples and measure the entropy of the resulting prediction. \Sewoong{I think this is a good idea. I do not see how to make it fail immediately. Perhaps \cite{saha2020hidden}'s hidden backdoor works on STRIP but not ours.  } 
%\Sewoong{I have an idea for evading STRIP detection using advanced steganography using GANs. Worth trying later.}

%\Sewoong{This is review for Madry's label consistent backdoor paper. We might want to read it before submission: https://openreview.net/forum?id=HJg6e2CcK7}
%\Sewoong{we might (or not) need to run some of these defenses}

\medskip\noindent{\bf Robust  estimation.}
There has been significant progress in robust mean and covariance estimation under Gaussian samples in \(\RR^d\).
\cite{chen2018robust} gives the first exponential time algorithm that accurately estimates the covariance matrix with \(\Omega\del{d}\) sample complexity under adversarial corruptions and  prove a matching information theoretical lower bound.
\cite{diakonikolas2019robust, lai2016agnostic} give the first polynomial time algorithm with no (or very weak) dependency on the dimensionality in the estimation error (close to the one in~\cite{chen2018robust}), however at the cost of \(\Omega\del{d^2}\) sample complexity. 
A \textit{statistical query} (SQ) lower bound is later shown in~\cite{diakonikolas2017statistical}, indicating that a polynomial time algorithm with \(\Omega\del{d^{1.99}}\) sample complexity is unlikely.
Recent work \cite{cheng2019faster, li2020robust} improve the time complexity to match the matrix multiplication time, which nearly matches the time needed for the non-robust version of the problem.

%\medskip\noindent
%{\bf Notations.}
%We use \(\sbr{k}=\set{1, \ldots, k}\) to denote the set of first \(k\) positive integers.

% ----------------------
\section{Threat model and diversifying the attacks} 
\subsection{Threat model}\label{sec:threat}

We assume the threat model of \cite{tran2018spectral}. 
The adversary has the training data and knows the user's neural architecture and training method. 
However, the adversary does not train the model herself. 
The user trains the model on training data that might be corrupted by the adversary, whose
%We want to train a classification model from training data ${\cal D}_n=\{(x_i ,y_i )\}_{i=1}^n$ coming from various sources, some of which are malicious and collectively called as the attacker. 
%An $\alpha$ fraction of ${\cal D}_n$ is provided by the attacker, which are called poisoned data. 
goal is to create a \textit{backdoor} in the user's trained model.
The purpose of a backdoor is twofold.
First, in order to avoid suspicion, the classification accuracy on the clean training data and clean test data should not decrease due to the presence of poisoned data (hence the name backdoor).
Second, when a clean test data (whose label is not the target label) is corrupted by an attacker-defined trigger, the backdoor should be activated and the example should be classified as the attacker-defined target label.

To create a backdoor, the adversary injects poisoned data in the training set. 
We test our defense against the pixel attack, periodic attack, and clean label attack. 
We vary the fraction of injected poisoned examples denoted by
\[
    \varepsilon \triangleq \frac{\text{\# of poisoned examples injected}}{\text{\# of uncorrupted examples with target label}}\;.
\]

% ----------------------

\subsection{Pixel attacks and \(m\)-way pixel attacks}\label{sec:mway}
One of the first successful demonstrations of a backdoor attack used a simple pixel attack \cite{gu2017badnets}.
An image is corrupted by a single pixel at a fixed location set to a fixed color. 
At training, images from a label different from the target (e.g., ``truck'') are corrupted and injected to the dataset labelled as the target, e.g., ``deer''.  
On the CIFAR-10 dataset, each label has \num{5000} clean examples.  
The pixel attack only requires as few as 250 poisoned examples (\(\varepsilon = \SI{5}{\percent}\)) to succeed in achieving \SI{92}{\percent} test accuracy on clean data and \SI{89}{\percent} test accuracy  on poisoned data (see \cref{fig:poison-acc-vs-eps}). 
\begin{figure}[h]
    \centering
    \(\overbrace{
    \begin{subfigure}[t]{0.245\linewidth}
        \centering
        \includegraphics[width=0.9\linewidth]{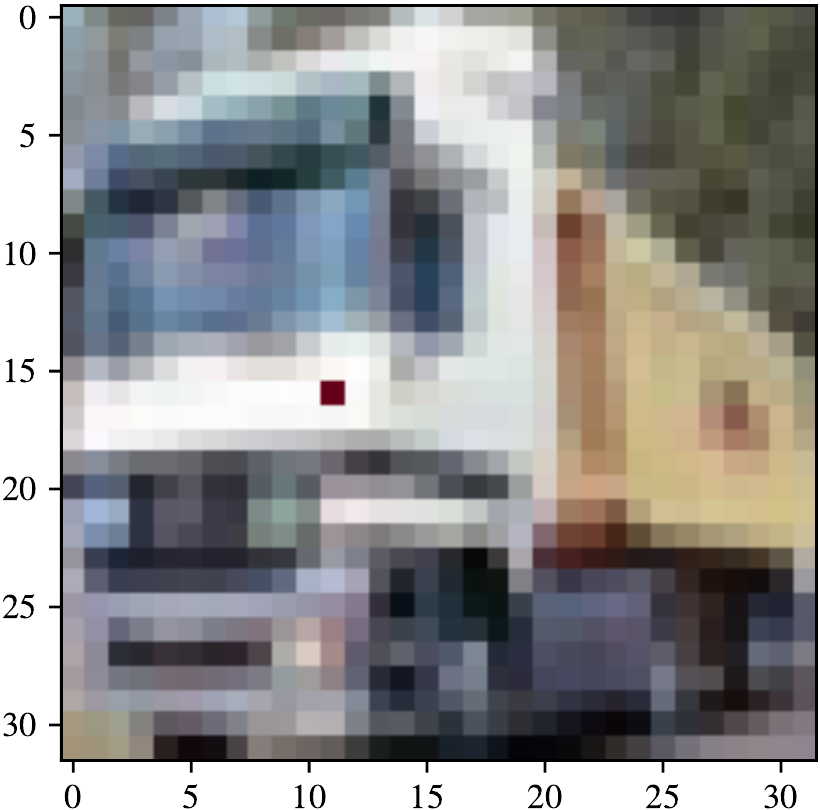}
        \caption{\nth{1} pixel}\label{fig:pixel-examples-1}
    \end{subfigure}%
    \begin{subfigure}[t]{0.245\linewidth}
        \centering
        \includegraphics[width=0.9\linewidth]{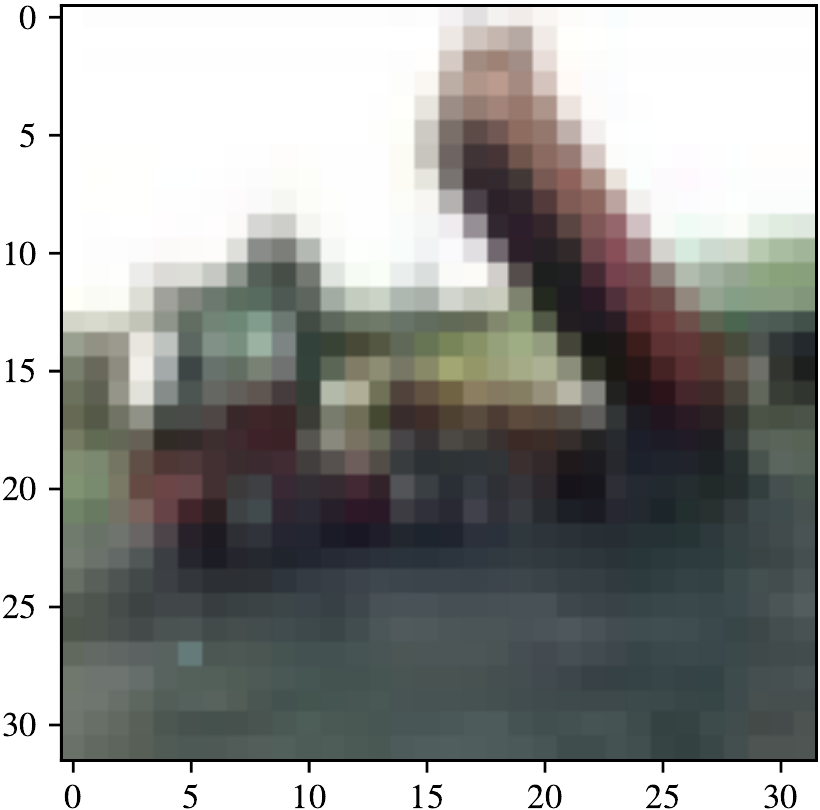}
        \caption{\nth{2} pixel}\label{fig:pixel-examples-2}
    \end{subfigure}%
    \begin{subfigure}[t]{0.245\linewidth}
        \centering
        \includegraphics[width=0.9\linewidth]{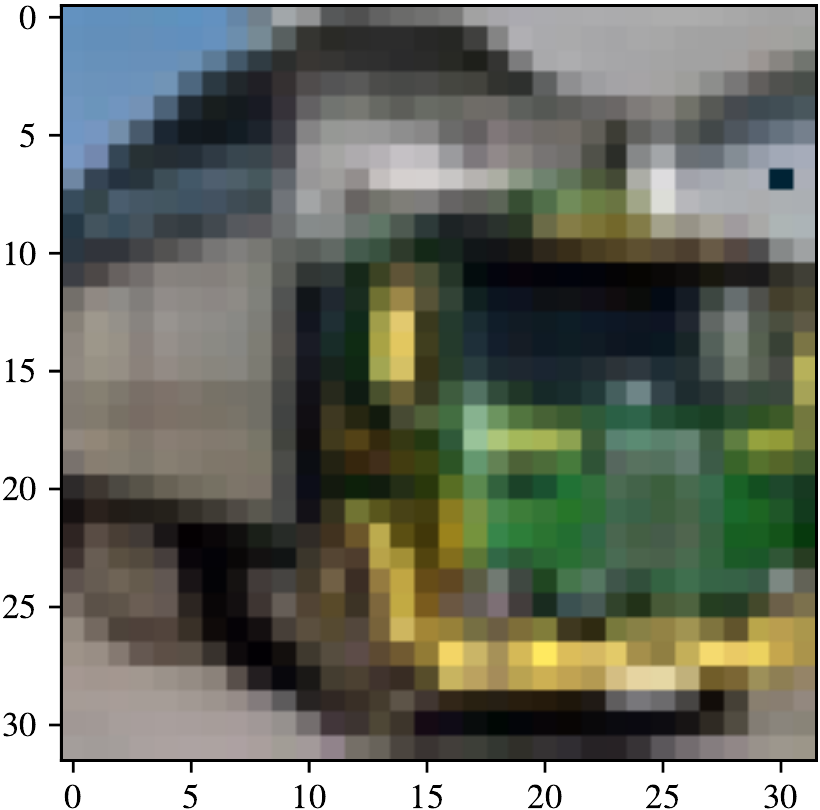}
        \caption{\nth{3} pixel}\label{fig:pixel-examples-3}
    \end{subfigure}%
    }^{\text{train}}
    \overbrace{\begin{subfigure}[t]{0.245\linewidth}
        \centering
        \includegraphics[width=0.9\linewidth]{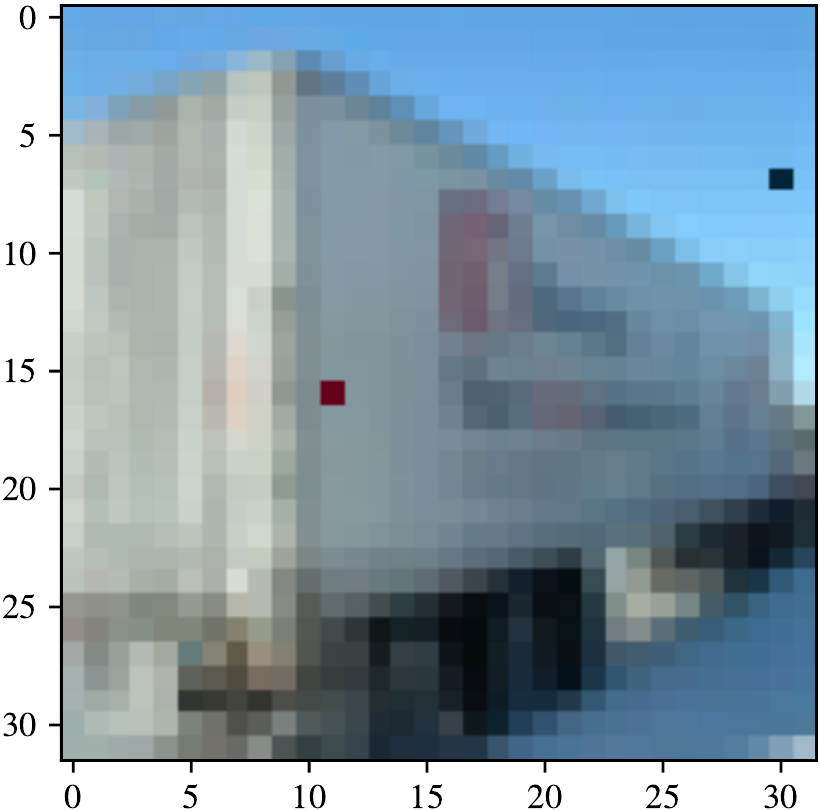}
        \caption{all three}\label{fig:pixel-examples-all}
    \end{subfigure}}^{\text{test}}\)
    \caption{
        During training, the \(m\)-way pixel attack partitions the data and applies a group-specific pixel attack to each. 
        At test time, all \(m\) pixels are applied to strengthen the trigger.  
    }\label{fig:pixel-examples}
\end{figure}

%We say an attack breaks down if  ${\rm ACC}_{\rm poison}$ drops below  0.9. 
%The pixel attack breaks down at $\varepsilon=$ as shown in Figure~\ref{fig:3way}. 
%${\rm ACC}_{\rm natural}$ remains above 0.9x for all experiments and all values of $\varepsilon$ and is omitted in the plot.  

A downside of the pixel attack is that it leaves strong spectral signatures, such that it can be easily detected by the PCA defense of \cite{tran2018spectral}, 
which successfully removes \SI{94}{\percent} of poisoned data when \(\varepsilon = \SI{10}{\percent}\).  
However, as PCA defense relies on a \textit{single} principal direction,  an \textit{\(m\)-way attack} introduced in \cite{xie2019dba} diversifies the watermark such that the spectral signature is hidden in the lower PCA subspaces. 
The corrupted training data is separated into \(m\) partitions and a group specific pixel attack is applied to each group.
At \(\varepsilon = \SI{10}{\percent}\), most of the poisoned samples under \(2\)-way pixel attack can evade detection by PCA defense, as shown in \cref{tab:pixel-full} in the appendix, while maintaining the poison accuracy of \SI{91}{\percent}. 
We compare the state-of-the-art defenses on various attacks and their \(m\)-way variations. % in \cref{sec:exp}.

%uses the top principal component direction of the representations of the combined training data at one of the intermediate layers of a trained neural network. 
%activations in one of the hidden layers, it can be vulnerable to simple strategies to diversify the attack. 
%uses one of the hidden layers of a  model trained with poisoned data. A training example is declared suspicious if its 

%We set the sensitivity of all detectors to remove $1.5\varepsilon n$ suspicious examples for experiments and all defences, although we observe that performance of the estimators are not too sensitive to this choice as we show in \ref{}.  

% ----------------------------
\section{Algorithm}\label{sec:algo}

\begin{figure}[h]
    \centering
    \includegraphics[width=\linewidth]{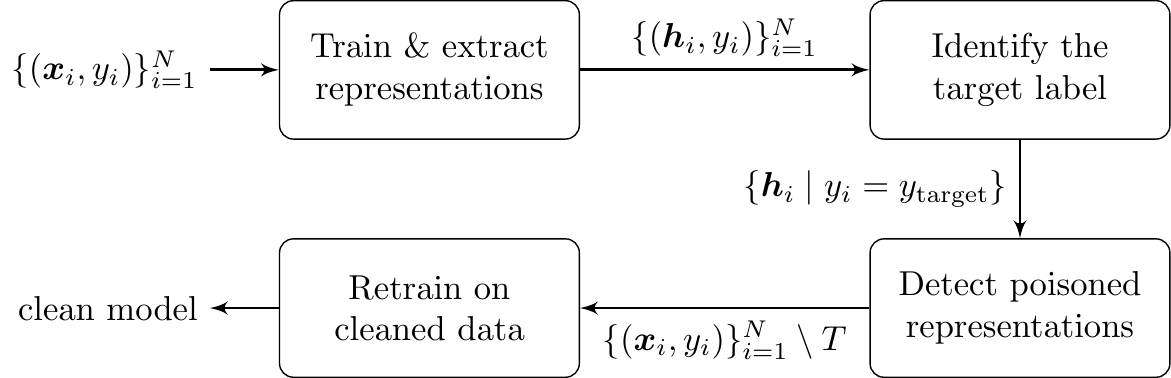}
    % \includegraphics[width=.37\textwidth]{./figures/pipeline.pdf}
    % \put(-203,80){$\{(x_i,y_i)\}_{i=1}^N$}
    % \put(-90,80){$\{h_i\}_{i=1}^N$}
    % \put(-90,62){$\{y_i\}_{i=1}^N$}
    % \put(-26,42){$\{h_i\}_{y_i=y_{\rm target}}$}
    % \put(-93,-5){$\{(x_i,y_i)\}_{i
    % \in [N]\setminus T}$}
    \caption{
      The defense pipeline.
      We first train a model on the poisoned data \(\set{(\bm x_i,y_i)}_{i=1}^N\) and extract the activation \(\bm h_i\in{\mathbb R}^d\) of a hidden layer of the trained neural network as the representation of the data \(\bm x_i\).
      Then, this representation is used in \cref{alg:target} to identify the target label.
      \cref{alg:detect} uses the representations \(\set{\bm h_i \mid y_i=y_{\mathrm{target}}}\) of the target label to detect and remove suspicious examples \(T\).
      Finally, we retrain a model with the cleaned data.
    }\label{fig:pipeline}
\end{figure}

The pipeline of our approach is to train a model and extract a representation from a middle layer,
then identify the target label with \cref{alg:target}, detect and remove the poisoned examples with \cref{alg:detect}, and retrain  
(see \cref{fig:pipeline}). 
In this section, we assume that the representations have been extracted and the target label has been correctly identified and focus on the robust poison detector.
We refer to \cref{sec:target} for the details on identifying the target label.
 
We propose the following three steps in SPECTRE (\cref{alg:detect}).
We first project the given representation data down to a \(k\)-dimensional space using its top left singular vectors.
We next apply robust estimation to get the approximate mean and covariance of the clean data.
After whitening the data with the estimated mean and covariance, the spectral signature of the poisoned data is amplified such that it can be detected more effectively.
Finally, we use QUantum Entropy (QUE) scores to find those with strong spectral signatures.
Note that the sensitivity of the algorithm is tuned by the choice of removing \(1.5\varepsilon n\) suspicious samples, following the same choice from \cite{tran2018spectral}.
We show that the performance is not sensitive to this choice in \cref{sec:limit2-sensitivity}.

\begin{algorithm2e}[h]
\caption{SPECTRE}\label{alg:detect}
\DontPrintSemicolon 
\KwIn{representation \(S=\{\bm h_i\in {\mathbb R}^d\}_{i=1}^n\), dimension \(k\), parameter \(\alpha\), poison fraction \(\varepsilon\)  }
\(\bm\mu(S)\gets \frac{1}{n}\sum_{i=1}^n \bm h_i\)\;
Center the data: \(S_1 \gets \{\bm h_i-\bm\mu(S)\}_{\bm h_i\in S}\)\;
\(U, \Lambda, V \gets \operatorname{SVD}_k\del{S_1}\)\;
%\(T_1 \gets \set{U^\top \del{\bm h_i-\bm\mu(S)} }_{\bm h_i\in S} \)\; 
%Sewoong: I think this should be \(T_1 \gets \set{U^\top \del{\bm h_i} }_{\bm h_i\in S} \)\; 
\(T_1 \gets \set{U^\top\bm h_i }_{\bm h_i\in S} \)\; 
\(\widehat\Sigma, \widehat{\bm \mu} \gets \textsc{RobustEst}\del{T_1, \varepsilon} \) \hfill[\cref{alg:robust-gaussian}]\\
 Whiten the data: \(T_2 \gets \{\widehat\Sigma^{-1/2}( \bar{\bm h}_i-\widehat{\bm \mu})\}_{\bar{\bm h}_i\in T_1} \) \\
 % Center the data: ??\\
 \(\{\tau_i\} \gets \textsc{QUEscore}(T_2,\alpha) \) \hfill[defined in \cref{eq:def_que}] \\
\Return \(1.5\varepsilon n\) samples with greatest QUE-scores\;
\end{algorithm2e}

\subsection{Step 1: Dimensionality reduction with SVD}\label{sec:alg-svd}

A robust estimation of the mean and covariance in \(d\)-dimensions with \(\varepsilon\) fraction of poisoned data requires \(\Omega\del{d^2/\varepsilon^2}\) samples, which we do not have in real data.
On CIFAR-10 experiments, the representations are \(\num{4096}\) dimensional and the number of samples per label is \(\num{5000}\).
We propose projecting the data down to a \(k\)-dimensional space using the top left singular vectors \(U\in{\RR}^{d \times k}\).
With a choice of \(k\) that is too small, the subspace \(U\) might not include the direction separating the poisons, thus losing statistical power for detection.
If we choose a \(k\) that is too large then the subspace \(U\) might contain directions where the clean data is not well-behaved and follows a heavy-tailed distribution, thus misleading the robust covariance estimation due to the small sample size.
Hence, we propose an algorithm to find an effective dimension \(k\) in \cref{alg:k-finder} and use it in all our experiments.
This achieves a performance close to the best performance one can achieve by enumerating all \(k\) as we show in \cref{sec:k-finder}.

\subsection{Step 2: Robust estimation}\label{sec:alg-rcov}

\begin{figure}[h]
    \centering
    \begin{subfigure}[t]{0.48\linewidth}
        \centering
        \includegraphics[height=0.7\linewidth, trim=0 1em -6em 0]{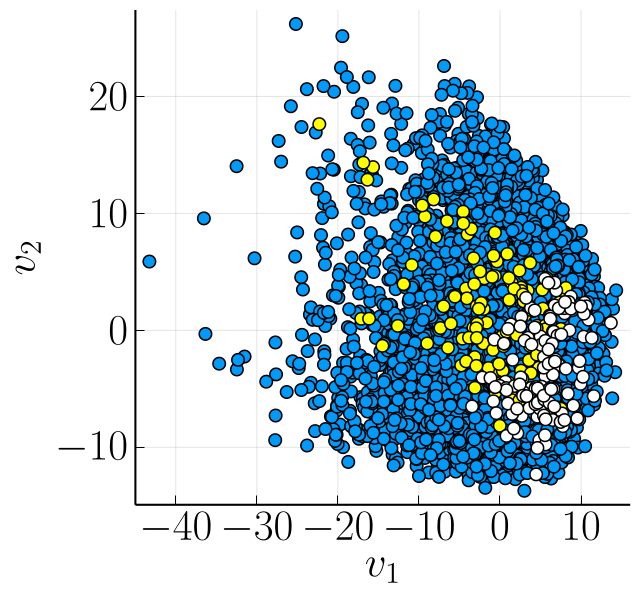}
        \caption{2-PCA}\label{fig:signature-scatter-pca}
    \end{subfigure}%
    \begin{subfigure}[t]{0.48\linewidth}
        \centering
        \includegraphics[height=0.7\linewidth, trim=-2em 1em 0 0]{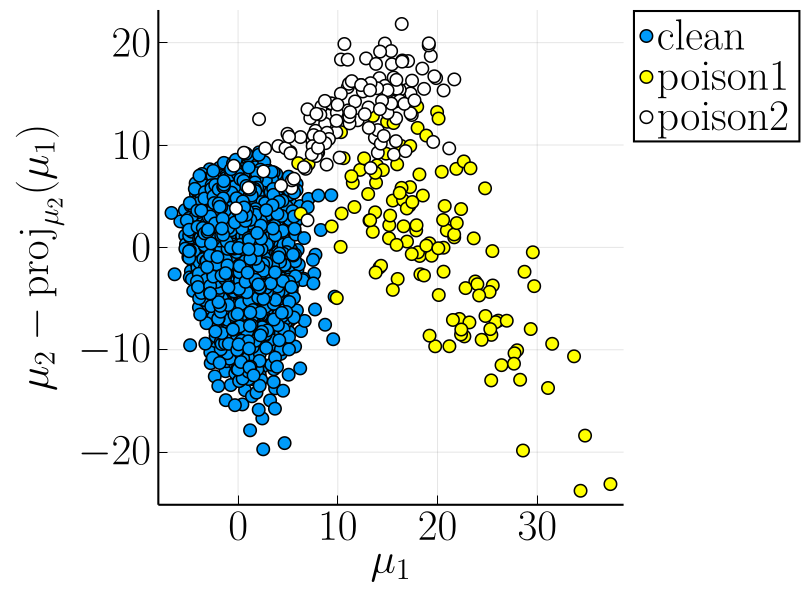}
        \caption{Poison mean subspace}\label{fig:signature-scatter-mu}
    \end{subfigure}%
    \caption{
      When we project onto the top PCA directions of the representations \(\set{\bm h_i}\) of the combined data on the left, the poisoned examples are  indistinguishable from the clean ones.
      On the two-dimensional subspace that best separates the poisons (right), on the other hand, the representations have  smaller variance, making those directions challenging to find. This example uses the \(2\)-way pixel attack with \(\varepsilon n = 250\).
    }\label{fig:signature-scatter}
\end{figure}

The PCA defense fails when the direction the algorithm checks (which is the top PCA direction of the combined data) is not aligned with the spectral signature of the poisoned examples (which is the direction that separates poisoned from clean data).
This happens when the covariance of the clean data has a large condition number such that the variance along the spectral signature direction is much smaller than the variance along the top PCA direction, as shown in \cref{fig:signature-scatter,fig:rep-histograms}.
In real data, the spectral signature commonly hides in such low-variance directions, causing PCA Defense to fail under most of the attacks we tested. 
 
If we know the true mean and covariance of the clean data, we can whiten the combined data to ensure that the clean data has the same variance along the spectral signature direction as any other directions, thus amplifying the hidden spectral signature.  
We propose using the recently introduced robust mean and covariance estimator, which is guaranteed to accurately estimate the true mean and covariance when we have enough samples from a Gaussian distribution. 

\begin{thm}[{\citep[Theorem~3.2 and Theorem~3.3]{diakonikolas2017being}}]\label{thm:robust}
    Let \(G \sim \mathcal{N}\del{\bm \mu, \Sigma}\) be a Gaussian in \(d\) dimensions, and let \(\varepsilon > 0\).
    Let \(S\) be an \(\varepsilon\)-corrupted set of samples from \(G\) of size \(\Omega\del{\del{d^2/\varepsilon^2}\operatorname{poly~log}\del{d/\varepsilon}}\).
    \textsc{RobustEst}(\(S\), \(\varepsilon\)), returns \(\widehat \Sigma\) and \(\widehat{\bm \mu}\), so that with probability at least \(9/10\), it holds that \(\norm{{\bm I} - \Sigma^{-1/2}\widehat \Sigma \Sigma^{-1/2}}_F = O\del{\varepsilon \log\del{1/\varepsilon}}\) and \(\norm{\bm \mu' - \bm \mu}_2 = O\del{\varepsilon \sqrt{\log\del{1/\varepsilon}}}\).
\end{thm}

Under the assumption that the clean data is drawn from a Gaussian distribution, this provides the best known guarantee for joint mean and covariance estimation and also matches the known fundamental limit on the achievable accuracy up to a logarithmic factor. 
%\Jon{Didn't we say the following already a bit earlier?}
However, in practice, we do not have enough samples to do robust estimation of the \(d=\num{4096}\) dimensional covariance in real data with CIFAR-10, where each label has \(\num{5000}\) samples.
It is critical to use an appropriate choice of \(k\)
in reducing the dimensionality of the samples down to \(k\) in the pre-processing.
In fact, a moderate choice of \(k=60\) can completely fail as we illustrate in \cref{fig:kfinder-large}.
To this end, we propose \cref{alg:k-finder} to identify the dimensionality \(k\).
For completeness we also provide \textsc{RobustEst} from \cite{diakonikolas2017being} in \cref{alg:robust-gaussian} in \cref{sec:robust-estimation}.

%Under the premise that the representations of the clean data is  \textit{well-behaved} and the poisoned data is  {\em separable} from the clean data, 
%When the principal component of the representations is not aligned with the direction that separates poisoned samples from clean ones,   poisoned samples  are challenging to detect. This happens when the variance of the clean samples along the direction of the poisoned data is much smaller than the 

\subsection{Step 3: Quantum entropy score poison detection}\label{sec:alg-que}

We want to assign an \textit{outlier score} \(\tau_i\) to each data point and remove those with high scores.
Once we whiten and center the representation according to the approximate mean and covariance of the clean data (denoted by \(\set{\tilde{\bm h}_i\in{\RR}^k}\)), the poisoned samples tend to be separated from the clean ones and are left with a \textit{spectral signature}.
Natural measures of this signature are the \textit{squared norm} \(\tau_i^{(0)}=(1/k)\norm{\tilde{\bm h}_i}_2^2\) and the \textit{squared projected norm}  \(\tau_i^{(\infty)}=\del{\bm v^\top \tilde{\bm h}_i}^2\) on the top principal direction \(\bm v\) of the  whitened representation  \(\set{\tilde{\bm h}_i}_{i=1}^n\) including both clean and poisoned samples.
In practice, either choice can fail as shown in \cref{tab:que}.
To this end, we propose using a  variation of QUantum Entropy (QUE) scoring from \cite{dong2019quantum}.

\begin{algorithm2e}[h]
\caption{QUantum Entropy scoring (\textsc{QUEscore}) based on \citep[Algorithm 2]{dong2019quantum}}\label{alg:que}
\DontPrintSemicolon 
\KwIn{\(T=\set{ \tilde{\bm h}_i\in \RR^k }_{i=1}^n\),  parameter \(\alpha\)}
%$\mu(T)\gets\frac{1}{n}\sum_{i=1}^n \tilde h_i$\\
 \begin{equation} 
    \tau^{(\alpha)}_i\gets \frac{\tilde{\bm h}_i^\top Q_\alpha \tilde{\bm h}_i}{\trace\del{Q_\alpha}}\;,\;\forall i\in[n]
    \label{eq:def_que}
\end{equation}
%where $c_\alpha=  {\rm Tr}( \exp( \,(\alpha /\|\Sigma(T)\|)\Sigma(T)\,)\,)$\\
where \(Q_\alpha =  \exp\del*{\frac{\alpha(\tilde\Sigma- {\mathbf I}) }{\norm{\tilde\Sigma}_2- 1}}\) and \(\tilde\Sigma = \frac{1}{n}\sum_{i=1}^n  \tilde{\bm h}_i \tilde{\bm h}_i^\top\)\;
\Return \(\set{\tau^{(\alpha)}_i}\) \;
\textit{note on \(\alpha\): we use \(\alpha=4\) in all experiments}
\end{algorithm2e}

QUE score defined in \eqref{eq:def_que} recovers \(\tau^{(0)}_i = (1/k) \norm{\tilde{\bm h}_i}^2\)when \(\alpha = 0\) and recovers \(\tau^{(\infty)}_i=(v^\top \tilde{\bm h}_i)^2\) when \(\alpha=\infty\). 
For intermediate \(\alpha\), this gracefully interpolates between these extremes, thus improving over both as shown below.

\begin{table}[h]
\centering
\begin{tabular}{lrrrrr} 
  \toprule
  Attacks \textbackslash{} Scores & \(\tau_i^{(0)}\) &  \(\tau_i^{(2)}\)  & \(\tau_i^{(4)}\)  & \(\tau_i^{(8)}\)  & \(\tau_i^{(\infty)}\) \\
  \midrule
  1-way \(\varepsilon=0.1\)  & 3  & 0  & 0  & 6  & 118 \\
  2-way \(\varepsilon=0.05\) & 69 & 40 & 30 & 49 & 97 \\
  % 2-way $\varepsilon=0.0124$ & 33   & 10  & 5 & 2 & 2 \\
  % 3-way $\varepsilon=0.025$ & 84   & 50  & 33 & 20 & 7 \\
  3-way \(\varepsilon=0.0124\) & 22 & 8  & 5 & 5 & 5 \\
  \bottomrule
\end{tabular}
\caption{
  Number of remaining poisoned samples after removing  \(1.5\varepsilon n\) examples with largest outlier scores  \(\tau_i^{(\alpha)}\) for various choices of \(\alpha\in\{0,2,4,8,\infty\}\).
  The proposed QUE score robustly achieves the best performance with \(\alpha=4\).
}\label{tab:que}
\end{table}

The name quantum entropy scoring comes from the fact that the matrix exponential  \(Q_\alpha/\trace\del{Q_\alpha}\) is a solution of a particular linear maximization with a quantum entropy regularization.
This matrix weighs the top and bottom principal directions differently,  
and the choice of \(\alpha\) controls how aggressively we want to emphasize the top principal directions.
This allows the QUE score to naturally adapt to the effective dimensionality of the spectral signature in poisoned samples. 
The squared norm  \(\tau_i^{(0)}\) fails when this effective dimension is small, which happens when the signature  is weak, i.e.~large \(m\) and small \(\varepsilon\). The squared projected norm \(\tau_i^{(\infty)}\) fails when the effective dimension is large, which happens when the signature is strong, i.e., small \(m\) and large \(\varepsilon\).
The experiments support this intuition and we provide details  in \cref{sec:analysis_que}. 
% The normalization by $\trace\del{Q_\alpha}$ ensures that the score does not blow up with large $\alpha$. 
The performance of the score is not sensitive to the choice of \(\alpha\) and we set it to \(4\) for all our experiments.  QUE score plays critical roles also in identifying the target label (\cref{alg:target}) and also in selecting the dimensionality \(k\) (\cref{alg:k-finder}).

\subsection{Possible extensions to SPECTRE}
%In the centering step, we could have used robust \textit{mean} estimation \cite{dong2019quantum} to replace \(\bm \mu(S)\) with the estimated  mean of the clean data.
In the dimensionality reduction step, we could have used robust \textit{principal component analysis} \cite{kong2020robust,jambulapati2020robust} to replace \(U\) with the estimated principal subspace of the clean data. 
Further, theoretically, we should partition the data into two groups \(S_1 \cup S_2 = S\), and project the data from one group onto the subspace learned from the SVD of the other group.
This ensures that the learned subspace does not overfit the data. 
In practice, these two variations did not give any improvement in performance.
%However, in practice, this also halves the effective data size, canceling the gain of avoiding overfitting. 
%We omitted both routines to simplify the algorithm. 

% ----------------------
\section{Experiments}\label{sec:exp}

In our pipeline (\cref{fig:pipeline}) for removing poison and retraining, we replace our proposed SPECTRE with two competing state-of-the-art approaches and compare the resulting performances.
Following \cite{tran2018spectral}, in all experiments, we set the sensitivity so that \(1.5\varepsilon n\) data points are removed in total, we use ``deer'' as the target label, and use images of trucks to create poisoned samples (unless otherwise stated).
In all experiments shown in this section, we use \cref{alg:k-finder} (explained in \cref{sec:k-finder}) to find the effective dimension \(k\) adaptively and automatically, and use \cref{alg:target} (explained in \cref{sec:target}) to identify the target label.
Due to space constraints, we only report the attack accuracy on the backdoored test examples on the final re-trained model. 
The accuracy  on the clean test examples is always between \(92.5\%\) and \(93.5\%\) unless otherwise stated, and is omitted from the results.
Complete statistics of the poison removal process are provided in \cref{sec:app_exp}.

%Following \cite{li2020backdoor}, for each attack, we  measurethe standard risk and the backdoor risk, and report how they change under each defense. We also measure False Positive Rate (PFR) and False Negative Rate (FNR) the following \cite{gao2019strip}.
% \Sewoong{Can we do word prediction like \cite{bagdasaryan2020backdoor}? GNN \cite{zhang2020backdoor,xi2020graph}? LSTM \cite{dai2019backdoor,chen2020badnl,kurita2020weight}? Traffic sign, speech signal, face recognition \cite{liu2018fine}} 
We compare three defenses: the proposed \cref{alg:detect}, the PCA defense of \cite{tran2018spectral} and the Clustering defense of \cite{chen2018detecting}. The Clustering defense uses standard \(2\)-means on the representations and we allow access to the oracle to determine one cluster and randomly select \(1.5\varepsilon n\) data points to remove from that cluster. 
Detailed descriptions are provided in \cref{sec:previous}. 
We evaluate them on three popular families of backdoor attacks. 

\iffalse
\cref{alg:robust-gaussian} partitions the dataset into two equal parts and performs some transformations to ensure that the robust covariance and robust mean estimators are applied properly.
In our experiments, we found that simply centering the data using the empirical mean and running the covariance estimator on the full set of samples gave improved performance since the poisoned samples were not too far from the mean. 
\fi

%One of the main contributions in this paper is a technique for hiding the spectral signature of a backdoor poisoning attack without sacrificing the attack's effectiveness.
%The idea is to split the \(m\) sets of roughly equal size and to use a different watermark for each of them.
%At test time, we simultaneously apply all \(m\) attacks

\subsection{\(m\)-way pixel attacks}\label{sec:experiments:pixel-attacks}

We test the defenses on the \(m\)-way pixel attacks described in \cref{sec:mway} with examples shown in \cref{fig:pixel-examples}.
Following the experiments of \cite{tran2018spectral}, we use a poisoned CIFAR-10 dataset to train a 32-layer ResNet\footnote{We modified the implementation at \url{https://github.com/akamaster/pytorch_resnet_cifar10} to match that used in \cite{tran2018spectral}.} model composed of three groups of residual blocks with 16, 32, and 64 filters respectively and 5 residual blocks per group. 
Details of the training are provided in \cref{sec:exp_detail}. 
A complete table of all the results including the number of poisoned training examples detected by each defense is provided in \cref{tab:pixel-full}. 

%We ran experiments for \(m \in \set{1, 2, 3}\) and six values of \(\varepsilon\).
%The results for \(\varepsilon n \in \set{500, 250, 125}\) are shown in \cref{tab:pixel-mini}.

\begin{table}[h]
\centering
\begin{tabular}{crllll} 
  \toprule
  \multicolumn{3}{l}{Attack} & \multicolumn{1}{l}{PCA} & \multicolumn{1}{l}{Clustering} &  \multicolumn{1}{l}{
  %{\crefname{algorithm}{Alg.}{Alg.} \cref{alg:detect}}
  SPECTRE
  } \\
  \(m\) & \(\varepsilon n\) & \(\mathrm{acc}_{\mathrm{p^*}}\) &  \(\mathrm{acc}_{\mathrm{p^*}}'\)  & \(\mathrm{acc}_{\mathrm{p^*}}'\) & \(\mathrm{acc}_{\mathrm{p^*}}'\) \\
  \midrule
  1 & 500 & 0.942 & 0.004 & 0.820 & \textbf{0.000}\\
  1 & 250 & 0.890 & 0.880 & 0.904 & \textbf{0.001}\\
  1 & 125 & 0.627 & 0.834 & 0.842 & \textbf{0.000}\\
  2 & 500 & 0.987 & 0.914 & 0.901 & \textbf{0.000}\\
  2 & 250 & 0.888 & 0.817 & 0.808 & \textbf{0.002}\\
  2 & 125 & 0.106 & 0.139 & 0.325 & \textbf{0.000}\\
  3 & 500 & 0.990 & 0.970 & 0.963 & \textbf{0.000}\\
  3 & 250 & 0.908 & 0.367 & 0.914 & \textbf{0.000}\\
  3 & 125 & 0.616 & 0.348 & 0.547 & \textbf{0.000}\\\bottomrule
\end{tabular}
\caption{
  \(m\)-way pixel attack test accuracy \(\mathrm{acc}_{\mathrm{p}^*}'\) on the backdoor examples of the model retrained with each defense.
  SPECTRE consistently eliminates the backdoor completely (achieving the poison accuracy near zero), in all regimes including those where existing methods fail.
  The attack  accuracy \(\mathrm{acc}_{\mathrm{p}^*}\) on  a model trained without any defense is shown as a reference.
}\label{tab:pixel-mini}
\end{table}

The PCA defense succeeds when the spectral signature is strong (\(m=1,\varepsilon=500\)) but fails when we diversify the attack, keeping the same number of poisons
or reducing the number of poisons, because the spectral signature is weaker. 
Robust covariance estimation consistently amplifies these signatures, eliminating the backdoor in all cases.
The clustering defense fails to separate poisons  from clean ones. 

%the spectral signature is weak, which happens when model \(m\)-way pixel attack results. \(a_{\mathrm{p}^*}\) is the accuracy on poisoned test data when all \(m\) pixel watermarks are used simultaneously. \(a_{\mathrm{p}^*}'\) is the respective quantity after the above defense has been applied and the network has been retrained. For more details and results for \(\varepsilon n \in \set{62, 31, 15}\), see \cref{tab:pixel-full}.

\subsection{\(m\)-way periodic attacks}
%\subsubsection{Attack Description}

\begin{figure}[h]
    \centering
    \(\overbrace{
    \begin{subfigure}[t]{0.32\linewidth}
        \centering
        \includegraphics[width=0.9\linewidth]{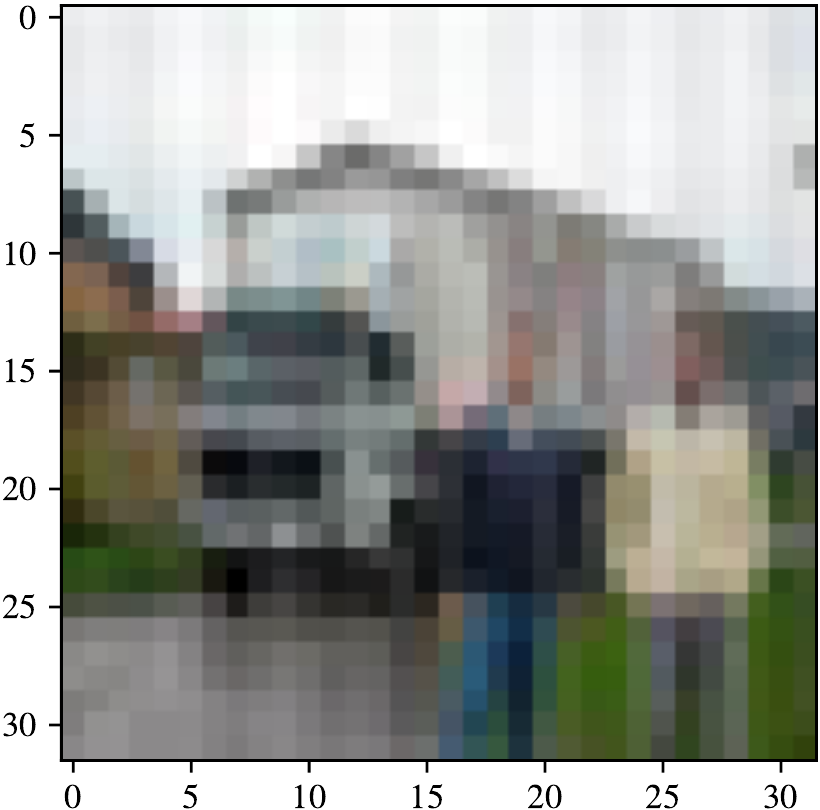}
        \caption{horizontal}\label{fig:periodic-examples-horizontal}
    \end{subfigure}%
    \begin{subfigure}[t]{0.32\linewidth}
        \centering
        \includegraphics[width=0.9\linewidth]{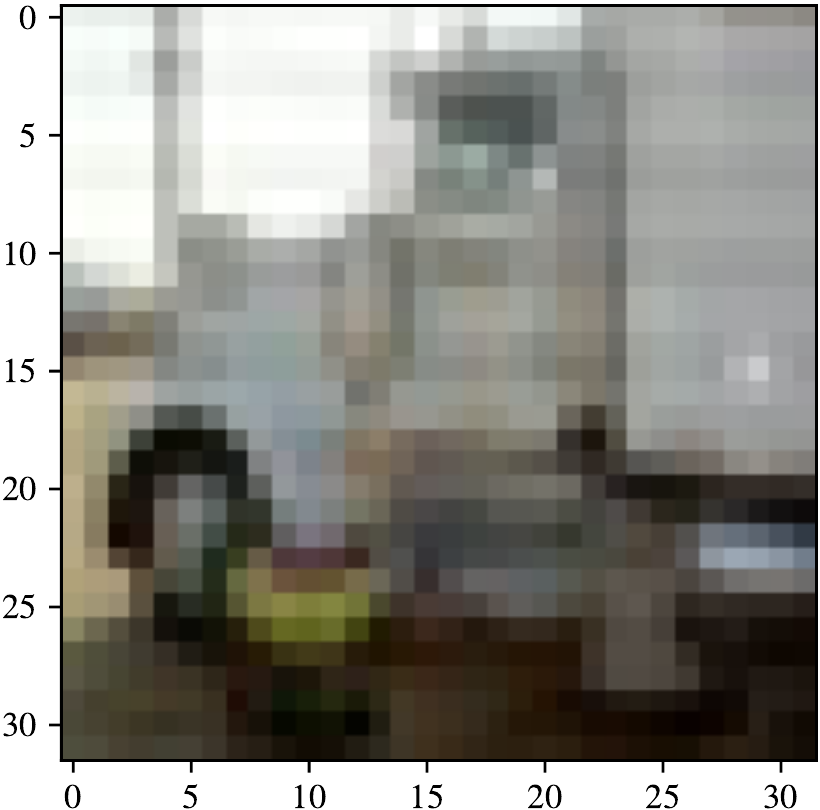}
       %\put(-90,77){training} 
       \caption{vertical}\label{fig:periodic-examples-vertical}
    \end{subfigure}%
    }^{\text{train}}
    \overbrace{\begin{subfigure}[t]{0.32\linewidth}
        \centering
        \includegraphics[width=0.9\linewidth]{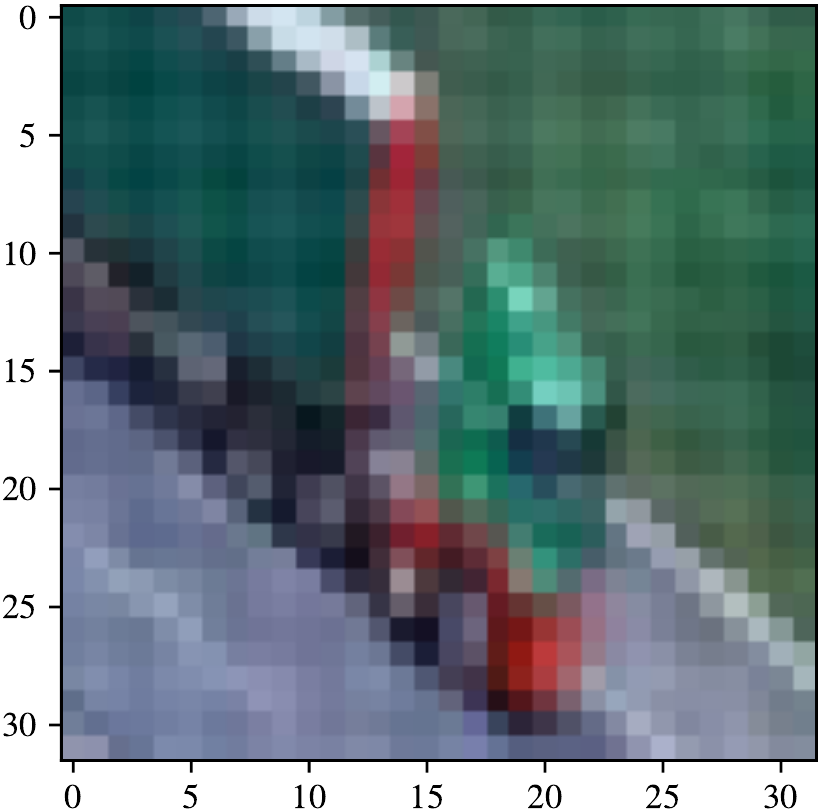}
       %\put(-90,77){training} 
       \caption{both}\label{fig:periodic-examples-all}
    \end{subfigure}}^{\text{test}}\)
    \caption{
      At training, each poisoned sample is corrupted by a single periodic signal to better hide the spectral signature.
      At test time, we combine all \(m\) triggers to boost the spectral signature and improve the accuracy of the attack.
    }\label{fig:periodic-examples}
\end{figure}

Proposed in \cite{barni2019new}, the periodic attack adds a periodic signal to the image as a trigger, as shown in \cref{fig:periodic-examples}.
%The poisoned examples are created by combining the clean image with an additive periodic signal in image space.
%The label is then changed from the source label to the target label.
%Some examples are shown in .
%Although the frequency of the signal is fixed, its phase may be shifted due to the random flip and random crop and pad, in a manner similar to the pixel attack.
We chose signals with amplitude 6 and frequency of 8.  %which seemed to balance the stealthiness and effectiveness of the attack.
%Since the addition may cause the image values to go above 255 or below 0, we also clip the values to the valid range.
We design an \(m\)-way periodic attack by choosing \(m\) different 
(frequency, direction) pairs. %At training, each poisoned sample is corrupted by a single periodic signal to hide the spectral signature and make it challenging to detect. At testing, we combine all $m$ triggers to boost the spectral signature and improve the poison target accuracy. 
%Only one periodic signal is used for each poisoned training example, but all periodic signals are added simultaneously at test time.
\cref{tab:periodic-full} in the appendix provides all the experimental results. The same experimental setting was used as in \cref{sec:experiments:pixel-attacks}.
\cref{alg:detect} consistently removes the backdoors completely, whereas competing defenses fail.

%\subsubsection{Experimental Setup}

%We used the same experimental setup as for the pixel attacks.

%\subsubsection{Results}

%We tested three defenses on \(m\)-way periodic attacks  using horizontal and vertical periodic signals for \(\varepsilon n \in \set{500, 250, 125}\)  shown in \cref{tab:periodic-mini}.
%More experimental results are provided in  \cref{tab:periodic-full} for \(\varepsilon n \in \set{62, 31, 15}\). 

\begin{table}[h]
\centering
\begin{tabular}{crllll} 
  \toprule
  \multicolumn{3}{l}{Attack} & \multicolumn{1}{l}{PCA} & \multicolumn{1}{l}{Clustering} &  \multicolumn{1}{l}{
  %{\crefname{algorithm}{Alg.}{Alg.} \cref{alg:detect}} 
  SPECTRE
  } \\
  \(m\) & \(\varepsilon n\) & \(\mathrm{acc}_{\mathrm{p}^*}\) & \(\mathrm{acc}_{\mathrm{p}^*}'\) & \(\mathrm{acc}_{\mathrm{p}^*}'\) & \(\mathrm{acc}_{\mathrm{p}^*}'\) \\
  \midrule
  1 & 500 & 0.975 & 0.976 & 0.987 & \textbf{0.004}\\
  1 & 250 & 0.961 & 0.968 & 0.933 & \textbf{0.001}\\
  1 & 125 & 0.912 & 0.916 & 0.889 & \textbf{0.000}\\
  2 & 500 & 0.996 & 0.995 & 0.988 & \textbf{0.001}\\
  2 & 250 & 0.982 & 0.986 & 0.961 & \textbf{0.000}\\
  2 & 125 & 0.881 & 0.868 & 0.829 & \textbf{0.000}\\\bottomrule
\end{tabular}
\caption{
  \(m\)-way periodic attack results with notations from \cref{tab:pixel-mini}.
  The attack accuracy of SPECTRE %\cref{alg:detect} 
  shows that the backdoor has been completely eliminated.
%Large \(a_{\mathrm{p}^*}'\) means 
%\(a_{\mathrm{p}^*}\) is the accuracy on poisoned test data when all \(m\) pixel watermarks are used simultaneously. \(a_{\mathrm{p}^*}'\) is the respective quantity after the above defense has been applied and the network has been retrained.  For more details and results for \(\varepsilon n \in \set{62, 31, 15}\), see \cref{tab:periodic-full}
}\label{tab:periodic-mini}
\end{table}

\subsection{Label consistent attacks}
The obvious discrepancy between the image and the target label (e.g., a truck labelled as a deer) in previously presented attacks makes it trivial for a human to detect the poison.
The label consistent attack, which was proposed in \cite{turner2019label}, designs images that are consistent with the target label, but can still create backdoors. % when injected to the training data.

\begin{figure}[h]
    \centering
    \begin{subfigure}[t]{0.33\linewidth}
        \centering
        \includegraphics[width=0.9\linewidth]{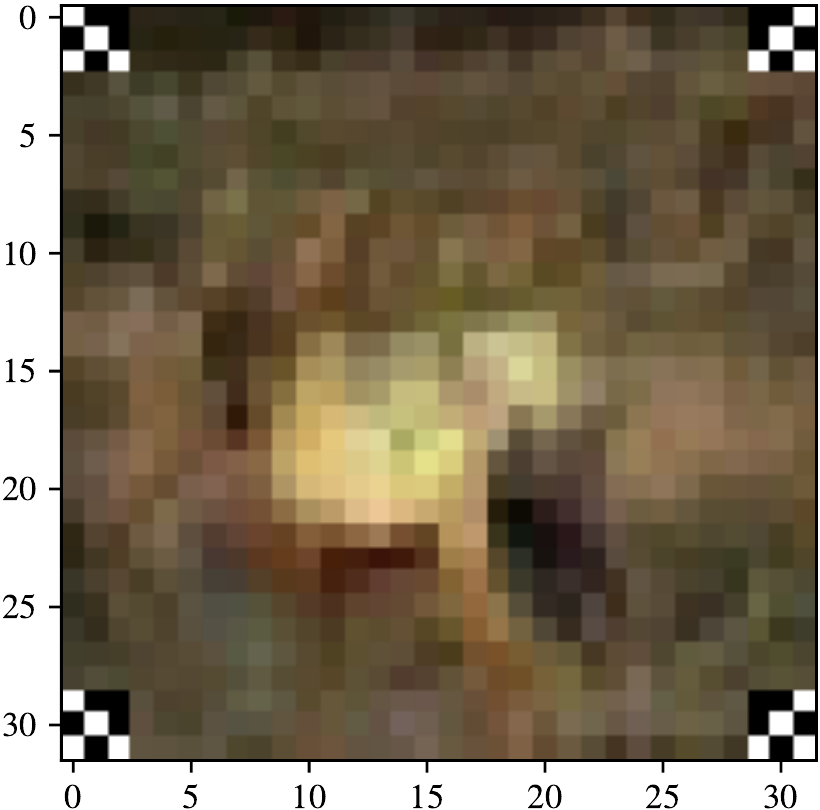}
        \caption{\(\ell_2\) perturbation}
        \label{fig:clean-label-examples-l2}
    \end{subfigure}%
    \begin{subfigure}[t]{0.33\linewidth}
        \centering
        \includegraphics[width=0.9\linewidth]{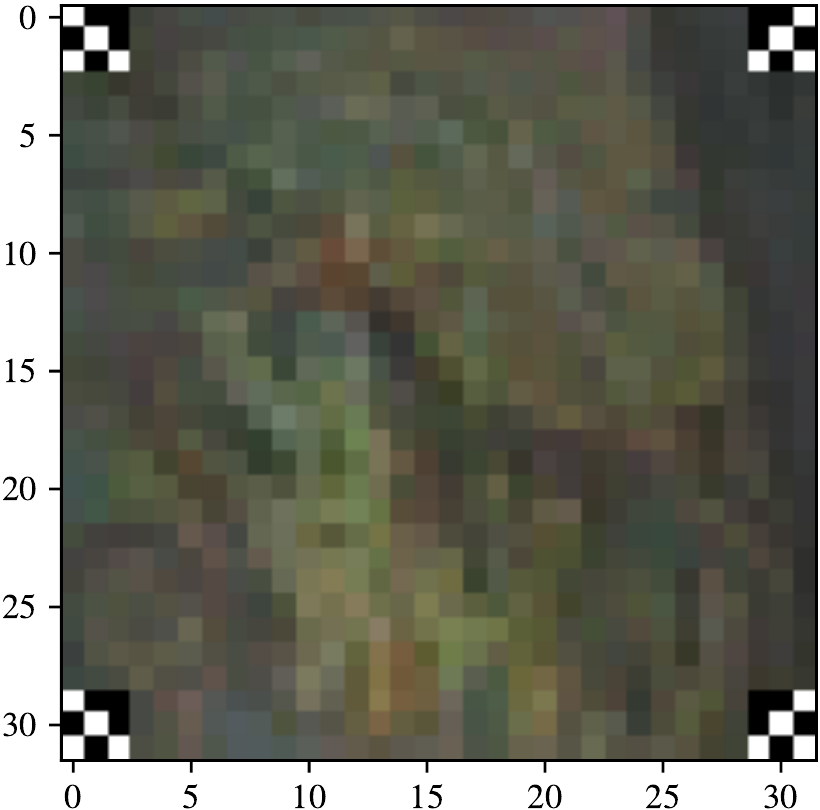}
        \caption{\(\ell_\infty\) perturbation}
        \label{fig:clean-label-examples-linf}
    \end{subfigure}%
    \begin{subfigure}[t]{0.33\linewidth}
        \centering
        \includegraphics[width=0.9\linewidth]{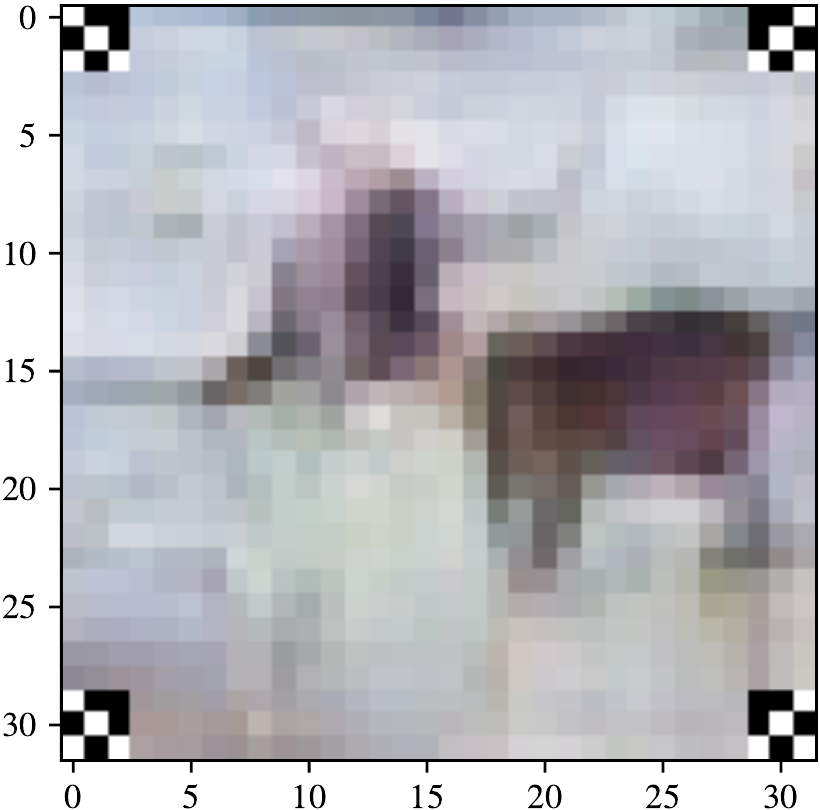}
        \caption{GAN interpolation}
        \label{fig:clean-label-examples-gan}
    \end{subfigure}%
    \caption{
      Examples training samples for label consistent attacks, which are visually consistent with the target label ``deer'', while succeeding in creating backdoors that are triggered by the watermark in the corners.
    }\label{fig:clean-label-examples}
\end{figure}

Concretely, three transforms are proposed to create images of the target label which are more difficult to classify: \(\ell_2\) and \(\ell_\infty\) bounded adversarial perturbations and interpolation via the latent space of a Generative Adversarial Network (GAN). 
A watermark, which in our case is a 3x3 patch of black and white pixels on each corner, is then added to the transformed images, 
During training, the network may come to rely on the watermark to classify the poisoned examples, as classifying them without the watermark is difficult.
At test time, the network outputs the target label whenever it detects the watermark.
Examples are shown in \cref{fig:clean-label-examples}.

%A poisoned image is produced by taking an image from the target label and transforming it in a way that makes it more difficult to classify.Then, a watermark is added to the image.In our case, the watermark used is a 3x3 patch of black and white pixels, which is replicated on each corner of the image.During training the network may come to rely on the watermark to classify the poisoned examples if they are otherwise difficult to classify.At test time, this may cause the network to output the target label whenever it detects the watermark.

%This attack differs substantially from the previous two because the label of the poisoned images is not changed.Additionally, because the poisoned examples are created using examples from the target label, the attack has no source label---all labels other than the target may be switched to the target by the backdoor at test time. We did not implement an \(m\)-way attack for this attack type.

%\subsubsection{Experimental Setup}
We used the same experimental setup as \cite{turner2019label}.
For our experiments, we ran the  provided implementation.\footnote{\url{https://github.com/MadryLab/label-consistent-backdoor-code}}
Accuracy on clean data was between 91\% and 92.5\% in all experiments and are omitted in the table.
More results are provided in \cref{tab:clean-label-full} in the appendix.

%The setup of \cite{turner2019label} appears to be very similar to that of \cite{tran2018spectral}.
%The same ResNet-32 architecture is used, albeit with a normal (i.e. not leaky) ReLU.

%Data standardization was enabled by default.
%Data augmentation was disabled by default, but we enabled it to ensure greater consistency with our previous experiments.
%We also enabled patch placement on all four corners to ensure the watermark would not be cropped out.

%, enabling poisoning on all four corners of the image and enabling data augmentation to ensure greater consistency with our other experiments.

%\subsubsection{Results}

%We ran experiments for \(\ell_2\) and \(\ell_\infty\) bounded adversarial perturbations and GAN interpolation for six values of \(\varepsilon\).The results for \(\varepsilon n \in \set{250, 125, 62}\) are shown in \cref{tab:clean-label-mini}.The results for \(\varepsilon n = 500\) are deferred to the appendix because all defences were equally effective.

\begin{table}[h]
\centering
\begin{tabular}{lrlrrr} 
  \toprule
  \multicolumn{3}{l}{Attack} & \multicolumn{1}{l}{PCA} & \multicolumn{1}{l}{Clustering} &  \multicolumn{1}{l}{
  %{\crefname{algorithm}{Alg.}{Alg.} \cref{alg:detect}}
  SPECTRE
  } \\
  type &\(\varepsilon n\) & \(\mathrm{acc}_{\mathrm{p}}\) & \(\mathrm{p}_{\mathrm{rm}}\) & \(\mathrm{p}_{\mathrm{rm}}\) & \(\mathrm{p}_{\mathrm{rm}}\) \\
  \midrule
  \(\ell_2\) & 250 & 0.932 & \textbf{250} & 140 & \textbf{250}\\
  \(\ell_2\) & 125 & 0.843 &   1 &  17 & \textbf{125}\\
  \(\ell_2\) &  62 & 0.856 &   0 &   5 & \textbf{62}\\
  \(\ell_\infty\) & 250 & 0.894 & \textbf{250} & 245 & \textbf{250}\\
  \(\ell_\infty\) & 125 & 0.744 &   0 &  24 & \textbf{125}\\
  \(\ell_\infty\) &  62 & 0.472 &   0 &   5 & \textbf{62}\\
  GAN & 250 & 0.584 &  47 &  78 & \textbf{250}\\
  GAN & 125 & 0.680 &  28 &  20 & \textbf{125}\\
  GAN &  62 & 0.261 &   0 &   2 & \textbf{62}\\
  \bottomrule
\end{tabular}
\caption{
  Under label consistent attacks each defense detects \(1.5\varepsilon n\) candidates to remove, out of which \(\mathrm{p}_{\mathrm{rm}}\) are actual poisoned examples.
  This matches the total number of poisoned examples \(\varepsilon n\) for SPECTRE.  %\cref{alg:detect}.
}\label{tab:clean-label-mini}
\end{table}

%For this family of attacks, we did not make any changes to the training system of \cite{turner2019label}, which does not provide re-training.
%However we note that 
\cref{alg:detect} removed all poisoned examples in every instance, guaranteeing that the backdoor was eliminated. %cannot succeed after retraining. 
However, in a wide regime, the PCA and Clustering defenses removed a small fraction of the poison or none at all.  %suggesting that the poison test accuracy would be near the level before the defense was applied.

\subsection{Finding the effective dimension \(k\)}\label{sec:k-finder}

\begin{figure}[h]
\centering
\includegraphics[width=\linewidth, trim=0 1em 0 0em]{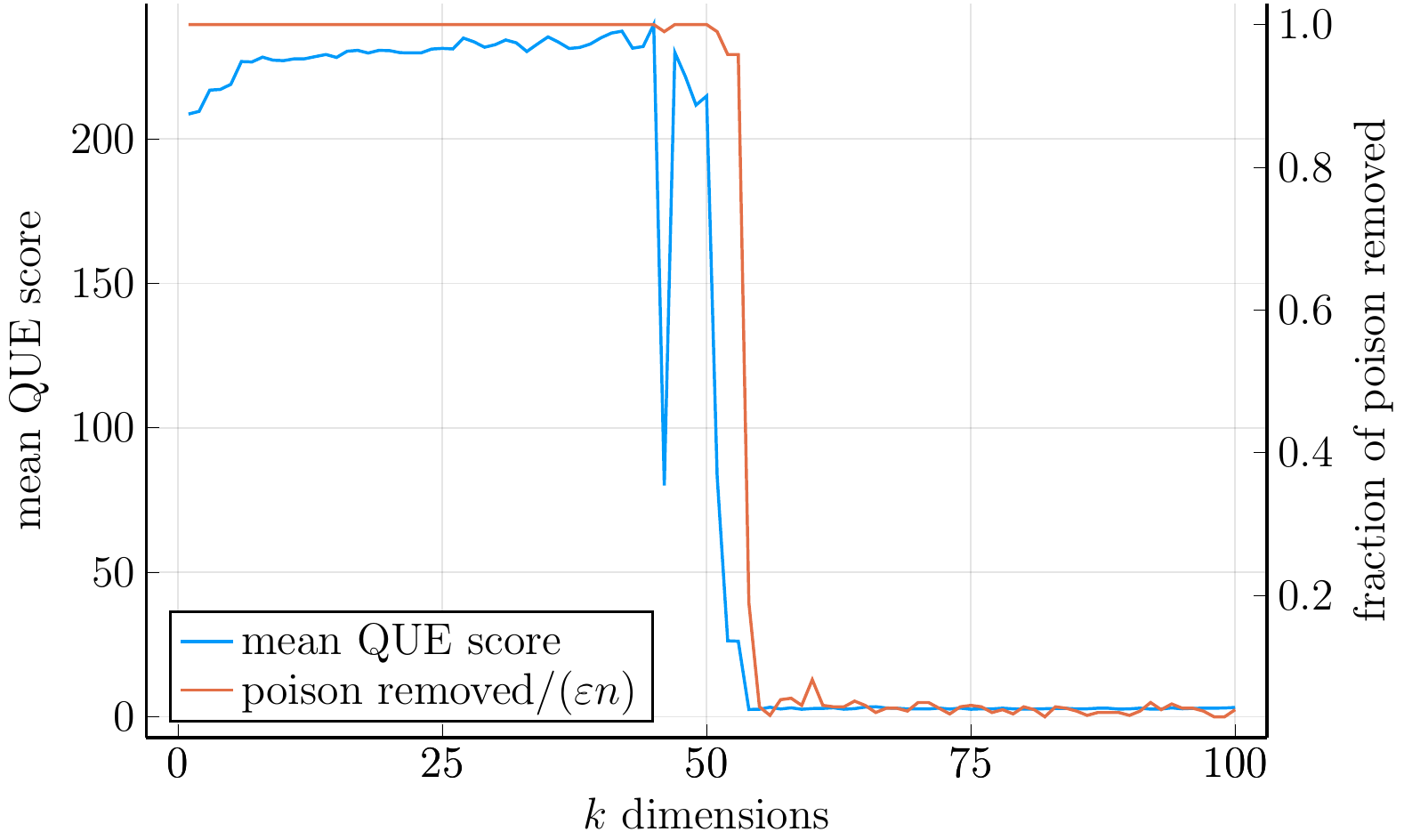}
\caption{
Under the GAN-based  label consistent attack with \(\varepsilon n = 500\), we want to choose
\(k \le 55\) as we want  the 
fraction of poisons removed by SPECTRE (in red) close to one.  We propose selecting \(k\) with the
highest mean QUE score (in blue), as it closely matches the true (unknown) detection accuracy.}
\label{fig:kfinder-large}
\end{figure}

\begin{figure}[h]
\centering
\includegraphics[width=\linewidth, trim=0 1em 0 0em]{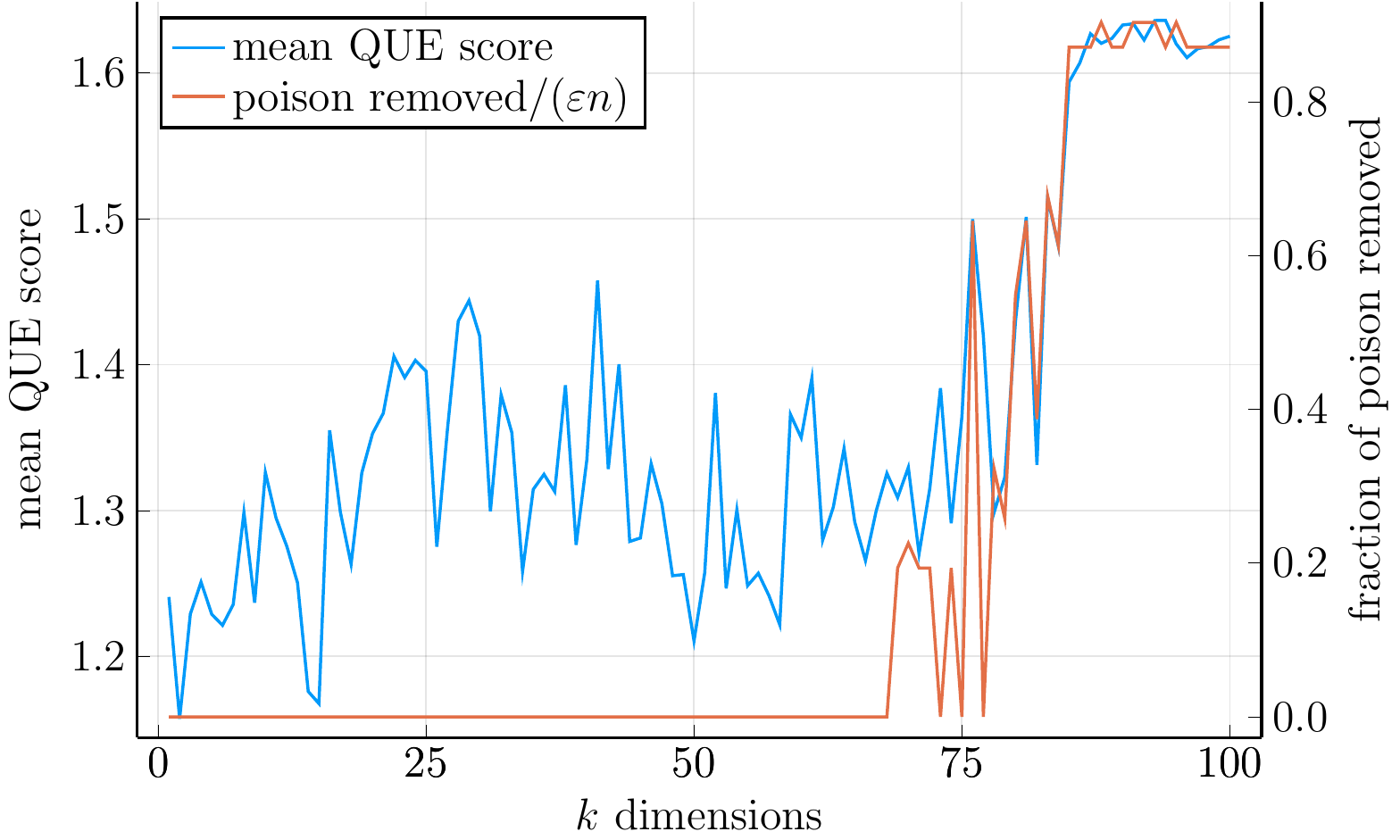}
\caption{
Under the 2-way pixel attack with \(\varepsilon n = 31\), we want to select \(k\geq 85\). 
We propose selecting \(k\) with the highest mean QUE score.}
\label{fig:kfinder-small}
\end{figure}

\cref{alg:detect} takes a parameter \(k\), which is the number of dimensions to use for covariance estimation.
Comparing \cref{fig:kfinder-large,fig:kfinder-small}, note that no fixed value of \(k\) works well for all experiments.
A small choice of \(k\) fails when the spectral signature is not in the top \(k\) PCA directions, which happens when the attack is weak (\cref{fig:kfinder-small}).
A large choice of \(k\) fails when the clean data is not well-behaved (resilience property fails) in the lower PCA subspaces causing robust covariance estimation to fail (\cref{fig:kfinder-large}).

%In particular, when defending against an attack with a strong spectral signature, using a value of \(k\) that is too high may cause the covariance estimation to fail as we do not have the \(O\del{k^2/\epsilon}\) samples required by \cref{alg:robust-covariance}, as seen in \cref{fig:kfinder-large}.

%On the other hand, when the attack has a weak spectral signature, a large \(k\) may be required to obtain a subspace where it is even possible to separate the poison from the clean data, as seen in \cref{fig:kfinder-small}.

%Although in this case we may still have too few samples to robustly estimate the covariance accurately, we have found that often the poison can still be removed.

A major challenge in selecting the appropriate \(k\) is that we do not have oracle access to the performance of our SPECTRE (in red), as in practice we do not know which samples are poisoned.
We therefore propose selecting \(k\) that maximizes the mean QUE score (in blue).
Concretely, for each \(k\) we run SPECTRE  to remove \(1.5\varepsilon n\) data points.
We use the covariance of the remaining cleaned examples (in the representation space) to whiten all the data, and compute the mean QUE score of all the data points after whitening.
The idea is that if poisons were correctly identified, then the mean QUE score will be large as poisons have strong spectral signature.
We write the algorithm explicitly in \cref{alg:k-finder}.
%We have found that the mean QUE scores are a good proxy for the number of poisoned examples removed.
\cref{tab:k-finder-showdown} shows that 
%that fixing \(k\) to a constant always leads to poor performance relative to the optimal \(k\) for at least one attack.
%On the other hand, 
\cref{alg:k-finder} selects nearly optimal values of \(k\). % whenever it is possible to remove at least 75\% of the poison.

\begin{algorithm2e}[h]
\caption{{\sc \(k\)-Identifier}}
\label{alg:k-finder}
\DontPrintSemicolon
\KwIn{representation \(S=\{\bm h_i\in {\mathbb R}^d\}_{i=1}^n\), parameter \(\alpha\), poison fraction \(\varepsilon\)} 
\(\bm\mu(S)\gets \frac{1}{n}\sum_{i=1}^n \bm h_i\)\;
Center the data: \(S_1 \gets \{\bm h_i-\bm\mu(S)\}_{\bm h_i\in S}\)\;
\(U, \Lambda, V \gets  \operatorname{SVD}_{k_{\max}}\del{S_1}\)\;
\For{\(k \in \sbr{k_{\max}}\)}{
    \(S_{\mathrm{removed}} \gets {\rm SPECTRE}\del{S, k, \alpha, \varepsilon}\)\hfill[\cref{alg:detect}]\\
    \(\Sigma' = \cov\del{\set{U^\top \bm h \mid \bm h \in S \setminus S_{\mathrm{removed}}}}\)\;
    \(\set{\tau_i} \gets \textsc{QUEscore}\del{\set{\Sigma'^{-1/2}U^\top \bm h \mid \bm h \in S}}\)\;\hfill[\cref{alg:que}]\\
    \(q \gets \frac{1}{n}\sum_{i=1}^n \tau_i\)\;
}
\Return \(k\) corresponding to the maximum \(q\) and the the maximum \(q\)\;
\end{algorithm2e}

\begin{table}[h]
    \centering
    \begin{tabular}{lrrrr}
        \toprule
        metric / choice of \(k\) & 20 & 100 & \(k_{\mathrm{oracle}}\) & {\crefname{algorithm}{Alg.}{Alg.} \cref{alg:k-finder}}\\
        \midrule
        mean  \(\mathrm{p}_{\mathrm{rm}}/\del{\varepsilon n}\) (\si{\percent})  & 76.5 & 86.8 & 98.6 & 98.2\\
        min   \(\mathrm{p}_{\mathrm{rm}}/\del{\varepsilon n}\) (\si{\percent})  &  0.0 & 4.0 & 90.3 & 87.1\\
        \bottomrule
    \end{tabular}
    \caption{Fixed choices of \(k\) results in failure in some examples, as shown by low min \% of poisons removed (\(\mathrm{p}_{\mathrm{rm}}/\del{\varepsilon n}\)). The minimum is over different attacks \cref{alg:k-finder} achieves a consistently reliable performance, close to the instance-wise optimal choice of \(k_{\mathrm{oracle}}\).
    %Values are percentages of total poisoned examples removed. The mean and minimum are taken over all experiments where the optimal choice \(k = k_{\mathrm{opt}}\) removed at least 75\% of the poison. 
    %The optimal mean and min values for fixed \(k\) were 87.4\% and 5.0\% for \(k = 95\) and \(k = 92\) respectively. 
    }
    \label{tab:k-finder-showdown}
\end{table}

\subsection{Identifying the target label}\label{sec:target}

\begin{figure}[h]
\includegraphics[width=\linewidth, trim=0 1em 0 1em]{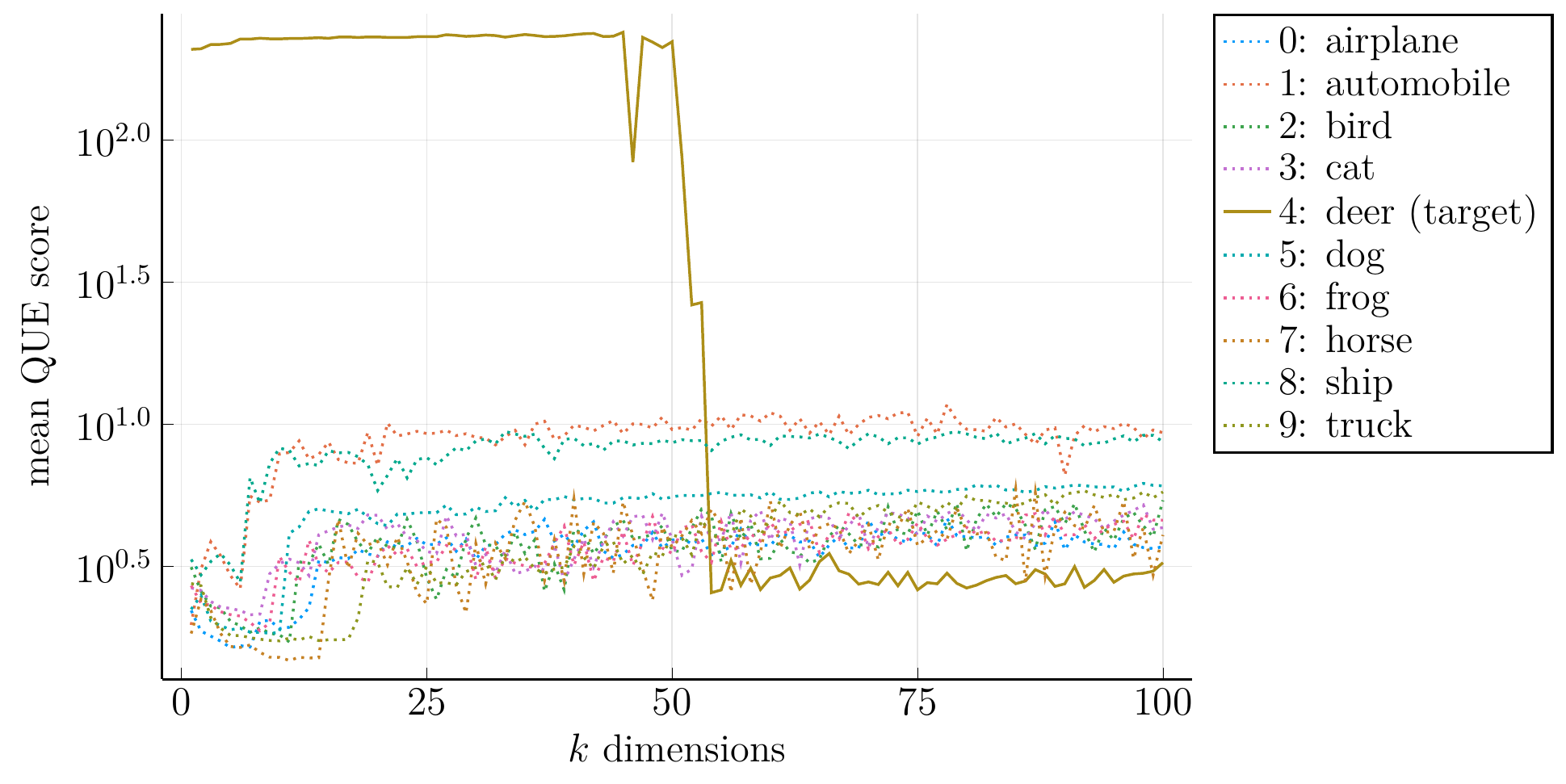}
\caption{
  GAN-based label consistent attack with \(\varepsilon n = 500\).
  We select \((k,\mathrm{label})\) pair that maximizes the mean QUE score.
}\label{fig:target-finder-small}
\end{figure}

\begin{figure}[h]
\includegraphics[width=\linewidth, trim=0 1em 0 1em]{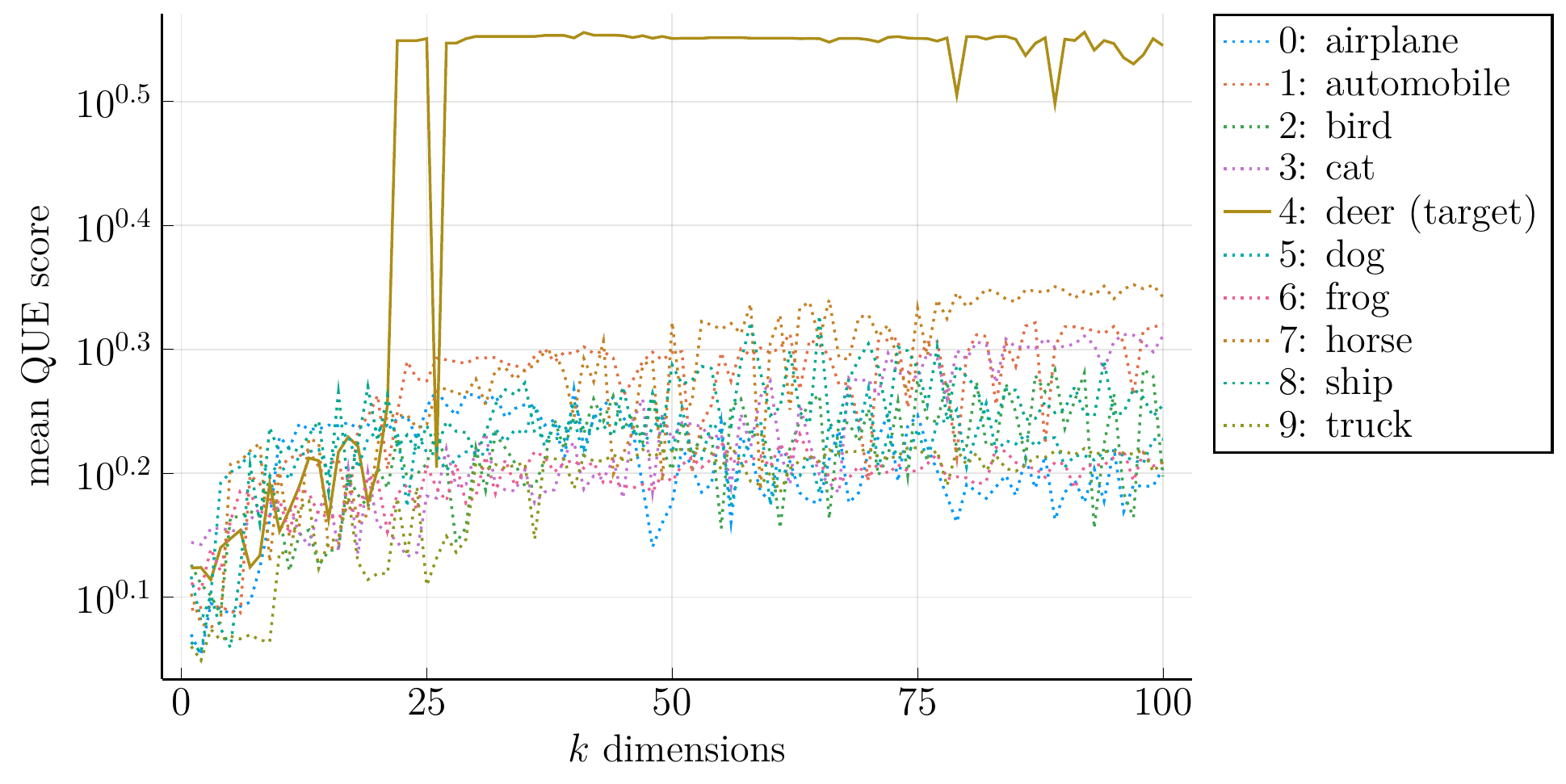}
\caption{
  \(3\)-way pixel attack with \(\varepsilon n = 125\).
  We select \((k,\mathrm{label})\) pair that maximizes the mean QUE score.
}\label{fig:target-finder-large}
\end{figure}

The defenses require  representations from the target label, which is not known.
%However in many cases, the target label is unknown to the defenders.
To identify which label is being targeted, we extend \cref{alg:k-finder}, which identifies the effective dimension \(k\), to identify both \(k\) and the target label \(l\), giving \cref{alg:target}.
\cref{fig:target-finder-large,fig:target-finder-small} show that the mean QUE scores obtained for the target label is clearly larger (for appropriate values of effective dimension \(k\)) compared to those obtained for untargeted labels. 
This follows from the same intuition as \cref{alg:k-finder}, where higher mean QUE score indicates the presence of poisoned data samples. 
We run \cref{alg:target} against all attacks with poison test accuracy \(\mathrm{acc}_{\mathrm p}\) over 0.33; the correct target label was identified in all those  experiments with  \SI{100}{\percent} accuracy.

\begin{algorithm2e}[h]
\caption{Target label identifier} 
\label{alg:target}
\DontPrintSemicolon 
\KwIn{representations \(S_l=\{\bm h_i\in {\mathbb R}^d\}_{i=1}^{n_l}\) for each label \(l\in\sbr{L}\), parameter $\alpha$, poison fraction \(\varepsilon\)}
\For{\(l \in \sbr{L}\)}{
    \(k, q \gets \textsc{\(k\)-Identifier}\del{S_l, \varepsilon}\) \hfill[\cref{alg:k-finder}]\\
}
\Return \(l\) corresponding to the maximum \(q\).
\end{algorithm2e}

%\subsection{Other attacks}

% \begin{itemize}
%     \item all attacks: pixel, periodic, clean-label
%     \item all defenses: Madry, Ours, 
%     \item TP+FN=1, TN+FP=1, Clean Test BER, Poison Test BER, 
%     \item retrain: 
% \end{itemize}

% ---------------

% To demonstrate that our defence works in varied settings, we also ran experiments on a standard 18-layer ResNet\footnote{We used the implementation at \url{https://github.com/kuangliu/pytorch-cifar} without modifications.} composed of four groups of residual blocks with 64, 128, 256, and 512 filters respectively and 2 residual blocks per group as suggested by  \cite{schwarzschild2020just}.
% Although this model is shallower than the ResNets-32, it is much wider and has more than twenty times as many parameters.

% ----------------------
\section{Ablation study}\label{sec:ablation}

SPECTRE combines several steps to effectively detect poisoned examples.
\begin{enumerate}
\item\label{step:dimension-reduction} Adaptive dimension reduction using \cref{alg:k-finder}.
\item\label{step:robust-estimation} The covariance of the clean samples is estimated using \cref{alg:robust-covariance}.
\item\label{step:whiten} The samples are whitened using the estimated covariance.
\item\label{step:que-scoring} We compute QUE scores using \cref{alg:que} to determine which samples to discard.
\end{enumerate}
Here we perform an ablation study to demonstrate that none of this steps can be omitted.
We show that Step~\ref{step:dimension-reduction} is necessary in \cref{sec:k-finder}, where we show that no constant choice of \(k\) is sufficient to detect the majority of the poison across multiple experiments.
Note that choosing \(k = d\)  is equivalent to performing no dimension reduction.
In our experiments, we found that checking values of \(k\) which are substantially smaller than \(d\) sufficed.
This also gave us a substantial computational speedup since the runtime of \cref{alg:detect} scales with \(k\).
% Additionally, dimension reduction substantially improves the runtime of \cref{alg:detect} since many
We show that Step~\ref{step:que-scoring} is important in \cref{sec:alg-que}.
In particular, in \cref{tab:que} we show that two other natural choices for outlier scoring can fail under certain conditions.
For Steps~\ref{step:robust-estimation}~and~\ref{step:whiten}, we provide \cref{tab:ablation}, which shows the performance of \cref{alg:detect} on a variety of experiments where Step~\ref{step:whiten} has been omitted (removing the need for Step~\ref{step:robust-estimation}) and where Step~\ref{step:robust-estimation} is omitted, and the whitening is done using the sample covariance.
The results in \cref{tab:ablation} justify the use of Steps~\ref{step:robust-estimation}~and~\ref{step:whiten}.

\addtolength{\tabcolsep}{-0.5pt}
\begin{table}[h]
  \centering
  \begin{tabular}{lrlrrrr}
    \toprule
    \multicolumn{4}{l}{Attack} & \multicolumn{1}{l}{1+4} &  \multicolumn{1}{l}{1+3+4} & \multicolumn{1}{l}{1+2+3+4} \\
    type & \(m\) & \(\varepsilon n\) & \(\mathrm{acc}_{\mathrm{p}^*}\) &  \(\mathrm{p}_{\mathrm{rm}}\) & \(\mathrm{p}_{\mathrm{rm}}\)& \(\mathrm{p}_{\mathrm{rm}}\)\\
    % 1+4 & 1+3+4 & 1+2+3+4\\
    \midrule
    % pixel & 1 & 250 & 0.894 & 131 & 203 & 249\\
    % pixel & 1 & 125 & 0.627 & 0 & 51 & 124\\
    % pixel & 3 & 125 & 0.616 & 0 & 37 & 123\\
    % periodic & 1 & 250 & 0.961 & 2 & 105 & 248\\
    % periodic & 1 & 125 & 0.912 & 0 & 67 & 124\\
    % periodic & 2 & 125 & 0.881 & 0 & 0 & 124\\pe
    pixel     & 1 & 500 & 0.942 & 471 & 471 & 500 \\
    pixel     & 1 & 250 & 0.894 & 131 & 203 & 249 \\
    pixel     & 1 & 125 & 0.627 & 0   & 51  & 124 \\
    pixel     & 3 & 500 & 0.990 & 153 & 336 & 490 \\
    pixel     & 3 & 250 & 0.908 & 0   & 119 & 245 \\
    pixel     & 3 & 125 & 0.616 & 0   & 37  & 123 \\
    periodic  & 1 & 500 & 0.975 & 19  & 421 & 493 \\
    periodic  & 1 & 250 & 0.961 & 2   & 105 & 248 \\
    periodic  & 1 & 125 & 0.912 & 0   & 67  & 124 \\
    periodic  & 2 & 500 & 0.996 & 457 & 407 & 493 \\
    periodic  & 2 & 250 & 0.982 & 10  & 115 & 248 \\
    periodic  & 2 & 125 & 0.881 & 0   & 0   & 124 \\
    \(\ell_2\) & 1 & 500 & 0.881 & 500 & 500 & 500 \\
    \(\ell_2\) & 1 & 250 & 0.932 & 250 & 250 & 250 \\
    \(\ell_2\) & 1 & 125 & 0.843 & 1   & 125 & 125 \\
    GAN       & 1 & 500 & 0.633 & 500 & 500 & 500 \\
    GAN       & 1 & 250 & 0.584 & 246 & 239 & 250 \\
    GAN       & 1 & 125 & 0.680 & 79  & 124 & 125 \\
    \bottomrule
  \end{tabular}
  \caption{
    Performance for various combinations of: 1. adaptive dimension reduction, 2. robust covariance estimation, 3. whitening, 4. QUE scoring.
    Note: 1+2+3+4 is SPECTRE, which performs better than the other combinations.
  }\label{tab:ablation}
\end{table}
\addtolength{\tabcolsep}{0.5pt}

\section{Conclusion}

While existing backdoor attacks are powerful enough to corrupt the trained model with a small fraction of injected poisoned training data, 
existing defenses 
%that do not have access to clean validation data fail to provide clean models 
fail under a broad regime of backdoor attacks. %the number of poisoned samples.% injected and a diverse set of attacks. 
The reason is that the spectral signatures that those methods build upon are challenging to detect for a wide range  of the attacks. We therefore introduce a novel defense algorithm, that we call SPECTRE, by combining the ideas from robust covariance estimation and quantum entropy outlier detection.   
Whitening with the robust covariance amplifies the spectral signature of the poisoned samples. The quantum entropy score can robustly detect that signature, adapting to the spectral profile of the poisoned examples. We demonstrate the superiority of our defense in several popular backdoor attacks,  which suggest that the proposed defense is successful in all regimes we tested on, including those where the state-of-the-art baseline approaches fail. 
The empirical success of SPECTRE opens several new research directions, two of which we discuss in the following. 

SPECTRE requires the trainer to have access to the corrupted training dataset. In some scenarios we might not have a direct access to the training data,  for example due to privacy constraints. Identifying the statistical signatures in such settings is an interesting direction to make SPECTRE more widely applicable.  A concrete direction is to design a decentralized and differentially private version of SPECTRE  under the setting of federated learning \cite{pillutla2019robust}. Recent advances in differentially private and robust estimators in \cite{prime} provide promising directions. 

\cite{gao2019strip} proposes a different paradigm for defending against backdoor attacks. The defense, called STRIP, mixes each training sample with multiple other samples and measure the entropy of the resulting prediction. This leverages an aspect of common backdoor attacks that is different from spectral signatures. 
Understanding how these different types of defenses perform against different types of attacks, such as the hidden backdoor attacks from  \cite{saha2020hidden}, is an important research question. 
% Perhaps \cite{saha2020hidden}'s hidden backdoor works on STRIP but not ours. 

\section*{Acknowledgement} 
Sewoong Oh is supported by Google faculty research award and NSF grants CNS-2002664, IIS-1929955, and CCF-2019844 as a part of Institute for Foundations of Machine Learning.

%Under a popular federated learning setting, we are not allowed to inspect any individual data points due to privacy concerns. We cannot apply outlier detection, Byzantine-tolerant distributed learning,  and the proposed spectral approaches.  Furthermore, federated learning is vulnerable against \textit{model poisoning attacks} where  malicious participants can use model replacement to introduce backdoor functionality into the aggregated model. This has been shown to be stronger than data poisoning attacks \cite{bagdasaryan2020backdoor}. 
%\cite{bhagoji2019analyzing}
%\cite{xie2019dba}
%\cite{chen2020backdoor}
%\cite{liu2020backdoor}

%In a slightly different adversarial scenario, \cite{bagdasaryan2020blind} studies \textit{code poisoning} where the attacker can interfere with the backpropagation. 

\bibliography{references}
\bibliographystyle{icml2020}

\clearpage 
\newpage 
\onecolumn 
\appendix
\section*{Appendix}
%  -------------------------
\section{Previous approaches}
\label{sec:previous}
For completeness, we write the algorithms we used for comparisons here. 

\subsection{Principal Component Defense}

The principal component defense was proposed in  \cite{tran2018spectral}.
They analyze the representations by projecting them onto the top eigenvector of their covariance and then removing points that are far from the mean.
This algorithm is shown in \cref{alg:pca_defense}.

% \noindent{\bf PCA Defense.}
%  \cite{tran2018spectral} proposed PCA Defense to detect spectral signatures of representations of the poisoned samples. 

\begin{algorithm2e}[h]
\caption{PCA Defense \cite{tran2018spectral}}
\label{alg:pca_defense}
\DontPrintSemicolon 
\KwIn{representation \(S=\{\bm h_i\in {\mathbb R}^d\}_{i=1}^n\) }
\(\bm \mu(S)\gets \frac{1}{n}\sum_{i=1}^n \bm h_i\)\;
Center the data: \(S_1 \gets \{\bm h_i-\bm \mu(S)\}_{\bm h_i\in S}\)\\
\(\bm v, \lambda, \bm u \gets \operatorname{SVD}_1\del{S_1}\)\;
\Return \(1.5\varepsilon n\) samples with greatest \(\abs{\inner{\bm h_i, \bm v}}\)\;
\end{algorithm2e}

\subsection{Clustering Defense}

The clustering defense was proposed in  \cite{chen2018detecting}. 
They analyze the representations produced by the network by reducing the dimension using principal component analysis and running a clustering algorithm on the result. 
The exact algorithm is shown in \cref{alg:clustering}.

\begin{algorithm2e}[h]
\caption{Activation Clustering \cite{chen2018detecting}}
\label{alg:clustering}
\DontPrintSemicolon 
\KwIn{representation \(S=\{\bm h_i\in {\mathbb R}^d\}_{i=1}^n\), dimension \(k\) } 
\(\bm \mu(S) \gets \frac{1}{n}\sum_{i=1}^n \bm h_i\)\\
Center the data: \(S_1 \gets \{\bm h_i-\bm \mu(S)\}_{\bm h_i\in S}\)\\
\(U, \Lambda, V \gets \operatorname{SVD}_k\del{S_1}\)\;
\(C_1, C_2 \gets \operatorname{2-means}\del{\set{U^\top \bm h \mid \bm h \in S_1}}\)\;
\Return clusters \(C_1, C_2\)\;
\end{algorithm2e}

\citet{chen2018detecting} propose several methods to determine which clusters, if any, contain poisoned representations.
To avoid these complexities, we equip the algorithm with an oracle, \textsc{ClusterOracle}, which given two clusters returns the cluster with the greatest fraction of poisoned examples.
The algorithm which returns the best cluster out of \(C_1, C_2\) give by the oracle should perform at least as well as any heuristic to determine which clusters to return.
There are two other concerns which make it difficult to compare this defense with \cref{alg:detect}:
first, there is no way to control how many examples are removed and second, the performance of the clustering varies with the initialization of \(k\)-means, which is random.
Therefore, we use a second step which repeatedly runs \cref{alg:clustering} and samples the cluster with the highest fraction of poison according to the oracle in order to build the set of samples to remove.
The algorithm is shown in \cref{alg:clustering_w_oracle}.

\begin{algorithm2e}[h]
\caption{Activation Clustering with Cluster Oracle}
\label{alg:clustering_w_oracle}
\DontPrintSemicolon 
\KwIn{representation \(S=\{\bm h_i\in {\mathbb R}^d\}_{i=1}^n\), dimension \(k\)} 
\(R \gets \varnothing\)\;
\While{\(\abs{R} < 1.5\varepsilon n\)}{
    \(C_1, C_2 \gets \textsc{ActivationClustering}\del{S, k}\) \hfill[\cref{alg:clustering}]\\
    \(C \gets \textsc{ClusterOracle}\del{C_1, C_2}\)\;
    Sample \(\bm h\) uniformly from \(C\)\;
    Add \(\bm h\) to \(R\) if \(\bm h \not\in R\)\;
}
\Return samples corresponding to \(R\)\;
\end{algorithm2e}

\cref{alg:clustering_w_oracle} should perform well whenever the clustering is able to effectively separate the poisoned examples from clean ones and its performance should have relatively low variance as \(R\) is built using many independent clustering runs.
Although this process is not guaranteed to terminate, we found that it did in all of our experiments.

% Note that 
% \[\EE\sbr{\text{poison fraction in \(R\)}} = \EE\sbr{\text{poison fraction in \(C\)}}.\]
% So \cref{alg:clustering_w_oracle} performs as well in expectation as simply returning the best cluster as chosen by the oracle.
% However, the algorithms has more consistent performance.
% In particular,
% \[\var\sbr{\text{poison fraction in \(R\)}} \le \frac{1 - \EE\sbr{\text{poison fraction in \(C\)}}}{1.5\varepsilon n}.\]
% So the variation in performance of \cref{alg:clustering_w_oracle} is low when its expected performance is high.

% \cite{chen2018detecting} proposed using standard clustering, like $k$-means, to separate the clean samples from poisoned samples in the representation space. 
% An explanation of the algorithm is in 
% Section \ref{sec:defense}. 

\FloatBarrier
\section{Complete experimental results}\label{sec:app_exp}

Complete experimental results for \(m\)-way pixel attacks, \(m\)-way periodic attacks, and label consistent attacks are shown in \cref{tab:pixel-full,tab:periodic-full,tab:clean-label-full} respectively.

\begin{table*}[h]
\centering
\begin{tabular}{crllrllrllrll} 
  \toprule
  \multicolumn{4}{l}{\(m\)-Way Pixel Attack} & \multicolumn{3}{l}{PCA Defense} & \multicolumn{3}{l}{Clustering Defense} &  \multicolumn{3}{l}{SPECTRE} \\
  \(m\) & \(\varepsilon n\) & \(\mathrm{acc}_{\mathrm{p}}\) & \(\mathrm{acc}_{\mathrm{p}^*}\) & \(\mathrm{p}_{\mathrm{rm}}\) & \(\mathrm{acc}_{\mathrm{p}}'\) & \(\mathrm{acc}_{\mathrm{p}^*}'\) & \(\mathrm{p}_{\mathrm{rm}}\) & \(\mathrm{acc}_{\mathrm{p}}'\) & \(\mathrm{acc}_{\mathrm{p}^*}'\) & \(\mathrm{p}_{\mathrm{rm}}\) & \(\mathrm{acc}_{\mathrm{p}}'\) & \(\mathrm{acc}_{\mathrm{p}^*}'\) \\
  \midrule
  1 & 500 & 0.942 & 0.942 & 471 & 0.004 & 0.004 & 375 & 0.820 & 0.820 & 500 & 0.000 & 0.000\\
  1 & 250 & 0.894 & 0.890 & 103 & 0.880 & 0.880 &  54 & 0.904 & 0.904 & 249 & 0.001 & 0.001\\
  1 & 125 & 0.627 & 0.627 &   0 & 0.834 & 0.834 &  11 & 0.842 & 0.842 & 122 & 0.000 & 0.000\\
  1 &  62 & 0.331 & 0.331 &   0 & 0.519 & 0.519 &   2 & 0.297 & 0.297 &  59 & 0.000 & 0.000\\
  1 &  31 & 0.075 & 0.075 &   0 & 0.023 & 0.023 &   0 & 0.010 & 0.010 &  30 & 0.000 & 0.000\\
  1 &  15 & 0.001 & 0.001 &   0 & 0.001 & 0.001 &   1 & 0.002 & 0.002 &   0 & 0.000 & 0.000\\[1ex]
  2 & 500 & 0.830 & 0.987 & 172 & 0.675 & 0.914 & 186 & 0.631 & 0.901 & 495 & 0.000 & 0.000\\
  2 & 250 & 0.588 & 0.888 &   9 & 0.503 & 0.817 &  35 & 0.518 & 0.808 & 237 & 0.002 & 0.002\\
  2 & 125 & 0.058 & 0.106 &   0 & 0.058 & 0.139 &   6 & 0.148 & 0.325 & 118 & 0.000 & 0.000\\
  2 &  62 & 0.009 & 0.017 &   0 & 0.007 & 0.011 &   1 & 0.002 & 0.007 &  59 & 0.000 & 0.000\\
  2 &  31 & 0.002 & 0.002 &   0 & 0.000 & 0.000 &   0 & 0.000 & 0.000 &  25 & 0.000 & 0.000\\
  2 &  15 & 0.000 & 0.000 &   0 & 0.001 & 0.000 &   0 & 0.000 & 0.000 &   0 & 0.000 & 0.000\\[1ex]
  3 & 500 & 0.742 & 0.990 & 147 & 0.665 & 0.970 & 204 & 0.606 & 0.963 & 486 & 0.001 & 0.000\\
  3 & 250 & 0.503 & 0.908 &   0 & 0.367 & 0.367 &  35 & 0.482 & 0.914 & 241 & 0.001 & 0.000\\
  3 & 125 & 0.225 & 0.616 &   0 & 0.083 & 0.348 &   4 & 0.186 & 0.547 & 122 & 0.000 & 0.000\\
  3 &  62 & 0.003 & 0.010 &   0 & 0.002 & 0.008 &   0 & 0.013 & 0.025 &  57 & 0.000 & 0.001\\
  3 &  31 & 0.001 & 0.001 &   0 & 0.000 & 0.002 &   0 & 0.000 & 0.001 &   0 & 0.000 & 0.000\\
  3 &  15 & 0.000 & 0.000 &   0 & 0.001 & 0.000 &   0 & 0.001 & 0.000 &   0 & 0.002 & 0.002\\\bottomrule
\end{tabular}
\caption{
  Under the \(m\)-way pixel attacks, the proposed robust poison detection in \cref{alg:detect}
  completely removes the backdoor for all \(m\in\set{1,2,3}\) and all sizes of the poisoned data \(\varepsilon n\), achieving the retrained accuracy of near zero on backdoored test samples.
  On the other hand, the state-of-the-art PCA and clustering defenses fail to remove enough poisons on almost all cases.
  There are 5,000 clean training samples with the target label ``deer''.
  \(\mathrm{acc}_{\mathrm{p}}\) is the accuracy on poisoned test data with  one pixel watermark and \(\mathrm{acc}_{\mathrm{p}^*}\) is the accuracy on poisoned test data with all \(m\) pixel watermarks simultaneously.
  \(\mathrm{acc}_{\mathrm{p}}'\) and \(\mathrm{acc}_{\mathrm{p}^*}'\) are the respective quantities after each defense has been applied and the network has been retrained.
  \({\mathrm p}_{\mathrm{rm}}\) is the number of poisoned examples removed by the defense, out of \(1.5\varepsilon n\) examples removed in total.
  Test accuracy on clean data was between 92.5\% and 93.5\% in all experiments and are omitted in the table.
}\label{tab:pixel-full}
\end{table*}

\begin{table*}[h]
\centering
\begin{tabular}{crllrllrllrll} 
  \toprule
  \multicolumn{4}{l}{\(m\)-Way Periodic Attack} & \multicolumn{3}{l}{PCA Defense} & \multicolumn{3}{l}{Clustering Defense} &  \multicolumn{3}{l}{SPECTRE} \\
  \(m\) & \(\varepsilon n\) & \(\mathrm{acc}_{\mathrm{p}}\) & \(\mathrm{acc}_{\mathrm{p}^*}\) & \(\mathrm{p}_{\mathrm{rm}}\) & \(\mathrm{acc}_{\mathrm{p}}'\) & \(\mathrm{acc}_{\mathrm{p}^*}'\) & \(\mathrm{p}_{\mathrm{rm}}\) & \(\mathrm{acc}_{\mathrm{p}}'\) & \(\mathrm{acc}_{\mathrm{p}^*}'\) & \(\mathrm{p}_{\mathrm{rm}}\) & \(\mathrm{acc}_{\mathrm{p}}'\) & \(\mathrm{acc}_{\mathrm{p}^*}'\) \\
  \midrule
  1 & 500 & 0.975 & 0.975 &  19 & 0.976 & 0.976 & 151 & 0.987 & 0.987 & 493 & 0.004 & 0.004\\
  1 & 250 & 0.961 & 0.961 &   2 & 0.968 & 0.968 &  40 & 0.933 & 0.933 & 249 & 0.001 & 0.001\\
  1 & 125 & 0.912 & 0.912 &   0 & 0.916 & 0.916 &  16 & 0.889 & 0.889 & 123 & 0.000 & 0.000\\
  1 &  62 & 0.744 & 0.744 &   0 & 0.764 & 0.764 &   4 & 0.722 & 0.722 &  62 & 0.001 & 0.001\\
  1 &  31 & 0.318 & 0.318 &   0 & 0.329 & 0.329 &   0 & 0.440 & 0.440 &  28 & 0.003 & 0.003\\
  1 &  15 & 0.003 & 0.003 &   0 & 0.005 & 0.005 &   0 & 0.002 & 0.002 &   0 & 0.007 & 0.007\\[1ex]
  2 & 500 & 0.896 & 0.996 & 176 & 0.873 & 0.995 & 172 & 0.824 & 0.988 & 499 & 0.001 & 0.001\\
  2 & 250 & 0.813 & 0.982 &  10 & 0.817 & 0.986 &  63 & 0.666 & 0.961 & 248 & 0.000 & 0.000\\
  2 & 125 & 0.501 & 0.881 &   0 & 0.460 & 0.868 &  10 & 0.416 & 0.829 & 124 & 0.000 & 0.000\\
  2 &  62 & 0.118 & 0.359 &   0 & 0.070 & 0.280 &   1 & 0.058 & 0.209 &  61 & 0.002 & 0.003\\
  2 &  31 & 0.012 & 0.057 &   0 & 0.001 & 0.010 &   0 & 0.015 & 0.067 &   0 & 0.004 & 0.021\\
  2 &  15 & 0.001 & 0.004 &   0 & 0.001 & 0.005 &   0 & 0.004 & 0.001 &   0 & 0.004 & 0.008\\\bottomrule
\end{tabular}
\caption{
  Under the \(m\)-way periodic attacks, the proposed robust poison detection in SPECTRE
  completely removes the backdoor for all \(m\in\set{1,2}\) and all sizes of the poisoned data \(\varepsilon n\), achieving the retrained accuracy of near zero on backdoored test samples.
  On the other hand, the state-of-the-art PCA and clustering defenses fail to remove enough poisons on almost all cases.
  There are 5,000 clean training samples  with the target label ``deer''.
  % We use the same notations as in \cref{tab:pixel-full}. \({\mathrm p}_{\mathrm{rm}}\) is the number of poisoned examples removed by the defense, out of \(1.5\varepsilon n\) examples removed in total.
  Accuracy on clean data was between 92.5\% and 93.5\% in all experiments and are omitted in the table.
}\label{tab:periodic-full}
\end{table*}

\begin{table*}[h]
\centering
\begin{tabular}{lrlrrr} 
  \toprule
  \multicolumn{3}{l}{Attack} & \multicolumn{1}{l}{PCA Defense} & \multicolumn{1}{l}{Clustering Defense} &  \multicolumn{1}{l}{SPECTRE} \\
  type &\(\varepsilon n\) & \(\mathrm{acc}_{\mathrm{p}}\) & \(\mathrm{p}_{\mathrm{rm}}\) & \(\mathrm{p}_{\mathrm{rm}}\) & \(\mathrm{p}_{\mathrm{rm}}\) \\
  \midrule
  \(\ell_2\) & 500 & 0.881 & 500 & 500 & 500\\
  \(\ell_2\) & 250 & 0.932 & 250 & 140 & 250\\
  \(\ell_2\) & 125 & 0.843 &   1 &  17 & 125\\
  \(\ell_2\) &  62 & 0.856 &   0 &   5 & 62\\
  \(\ell_2\) &  31 & 0.051 &   0 &   1 & 31\\
  \(\ell_2\) &  15 & 0.018 &   0 &   0 & 0\\[1ex]
  \(\ell_\infty\) & 500 & 0.798 & 500 & 500 & 500\\
  \(\ell_\infty\) & 250 & 0.894 & 250 & 245 & 250\\
  \(\ell_\infty\) & 125 & 0.744 &   0 &  24 & 125\\
  \(\ell_\infty\) &  62 & 0.472 &   0 &   5 & 62\\
  \(\ell_\infty\) &  31 & 0.024 &   0 &   0 & 31\\
  \(\ell_\infty\) &  15 & 0.017 &   0 &   0 & 0\\[1ex]
  GAN & 500 & 0.633 & 500 & 500 & 500\\
  GAN & 250 & 0.584 &  47 &  78 & 250\\
  GAN & 125 & 0.680 &  28 &  20 & 125\\
  GAN &  62 & 0.261 &   0 &   2 & 62\\
  GAN &  31 & 0.022 &   0 &   0 & 0\\
  GAN &  15 & 0.010 &   0 &   0 & 0\\
  \bottomrule
\end{tabular}
\caption{
  The number of removed poisoned examples \(\mathrm{p}_{\mathrm{rm}}\) under label consistent attacks.
  SPECTRE successfully removes all poisoned examples whenever the attack accuracy is larger than \SI{10}{\percent}.
  Accuracy on clean data was between \SI{91}{\percent} and \SI{92.5}{\percent} in all experiments and are omitted in the table.
}\label{tab:clean-label-full}
\end{table*}

\FloatBarrier
\section{Supplemental experimental results for different source-target label pairs}\label{sec:supp_exp}

In our previous experiments, we chose ``deer'' as the source label and ``truck'' as the target label following \cite{tran2018spectral}.
We also ran the \(m\)-way pixel attack experiments for \(m \in \set{1, 3}\) and \(\varepsilon n \in \set{500, 125}\) for ten combinations of source and target labels.
The results are shown in \cref{tab:pixel-vs-labels}.
Overall the trend in performance is similar, although there are some cases where none of the defences work well.
We suspect that this is because the representations of the clean and poisoned samples are merged at an earlier point in the network, making them difficult to distinguish once they reach the penultimate residual block.
We believe exploring this phenomenon presents an interesting research direction.

\begin{table}[H]
\centering
\begin{tabular}{cllrllrllrllrll}
  \toprule
  \multicolumn{6}{l}{$m$-Way Pixel Attack} & \multicolumn{3}{l}{PCA Defense} & \multicolumn{3}{l}{Clustering Defense} &  \multicolumn{3}{l}{SPECTRE} \\
  \(\ell_{\mathrm{s}}\) & \(\ell_{\mathrm{t}}\) & \(m\) & \(\varepsilon n\) & \(\mathrm{acc}_{\mathrm{p}}\) & \(\mathrm{acc}_{\mathrm{p}^*}\) & \(\mathrm{p}_{\mathrm{rm}}\) & \(\mathrm{acc}_{\mathrm{p}}'\) & \(\mathrm{acc}_{\mathrm{p}^*}'\) & \(\mathrm{p}_{\mathrm{rm}}\) & \(\mathrm{acc}_{\mathrm{p}}'\) & \(\mathrm{acc}_{\mathrm{p}^*}'\) & \(\mathrm{p}_{\mathrm{rm}}\) & \(\mathrm{acc}_{\mathrm{p}}'\) & \(\mathrm{acc}_{\mathrm{p}^*}'\) \\
  \midrule
  0 & 9 & 1 & 500 & 0.978 & 0.978 & 397 & 0.655 & 0.655 & 254 & 0.979 & 0.970 & 496 & 0.002 & 0.002 \\
  0 & 9 & 1 & 125 & 0.913 & 0.913 & 3   & 0.865 & 0.865 & 11  & 0.845 & 0.845 & 124 & 0.009 & 0.009 \\
  0 & 9 & 3 & 500 & 0.834 & 0.995 & 15  & 0.823 & 0.997 & 79  & 0.814 & 0.996 & 374 & 0.223 & 0.576 \\
  0 & 9 & 3 & 125 & 0.464 & 0.868 & 0   & 0.475 & 0.890 & 3   & 0.158 & 0.474 & 47  & 0.013 & 0.025 \\[1ex]
  1 & 7 & 1 & 500 & 0.963 & 0.963 & 195 & 0.933 & 0.933 & 237 & 0.905 & 0.905 & 500 & 0.001 & 0.001 \\
  1 & 7 & 1 & 125 & 0.758 & 0.758 & 0   & 0.665 & 0.665 & 17  & 0.750 & 0.750 & 125 & 0.000 & 0.000 \\
  1 & 7 & 3 & 500 & 0.765 & 0.986 & 15  & 0.714 & 0.979 & 138 & 0.687 & 0.969 & 498 & 0.000 & 0.000 \\
  1 & 7 & 3 & 125 & 0.2   & 0.598 & 0   & 0.127 & 0.441 & 5   & 0.313 & 0.746 & 122 & 0.001 & 0.001 \\[1ex]
  2 & 5 & 1 & 500 & 0.963 & 0.963 & 417 & 0.682 & 0.682 & 259 & 0.985 & 0.985 & 493 & 0.026 & 0.026 \\
  2 & 5 & 1 & 125 & 0.758 & 0.758 & 94  & 0.020 & 0.020 & 13  & 0.956 & 0.956 & 119 & 0.024 & 0.024 \\
  2 & 5 & 3 & 500 & 0.765 & 0.986 & 17  & 0.781 & 0.995 & 66  & 0.789 & 0.991 & 375 & 0.042 & 0.099 \\
  2 & 5 & 3 & 125 & 0.2   & 0.598 & 1   & 0.306 & 0.754 & 4   & 0.043 & 0.187 & 27  & 0.055 & 0.196 \\[1ex]
  3 & 8 & 1 & 500 & 0.993 & 0.993 & 491 & 0.004 & 0.004 & 355 & 0.966 & 0.966 & 500 & 0.003 & 0.003 \\
  3 & 8 & 1 & 125 & 0.94  & 0.940 & 0   & 0.941 & 0.941 & 26  & 0.935 & 0.935 & 125 & 0.003 & 0.003 \\
  3 & 8 & 3 & 500 & 0.825 & 0.997 & 1   & 0.819 & 0.998 & 152 & 0.601 & 0.947 & 482 & 0.006 & 0.004 \\
  3 & 8 & 3 & 125 & 0.131 & 0.448 & 0   & 0.102 & 0.340 & 5   & 0.021 & 0.074 & 113 & 0.002 & 0.005 \\[1ex]
  4 & 1 & 1 & 500 & 0.951 & 0.951 & 283 & 0.994 & 0.994 & 252 & 0.986 & 0.986 & 500 & 0.001 & 0.001 \\
  4 & 1 & 1 & 125 & 0.951 & 0.951 & 0   & 0.956 & 0.956 & 8   & 0.944 & 0.944 & 125 & 0.001 & 0.001 \\
  4 & 1 & 3 & 500 & 0.89  & 0.996 & 0   & 0.851 & 0.998 & 107 & 0.782 & 0.994 & 461 & 0.003 & 0.007 \\
  4 & 1 & 3 & 125 & 0.159 & 0.536 & 0   & 0.226 & 0.657 & 4   & 0.376 & 0.822 & 0   & 0.074 & 0.346 \\[1ex]
  5 & 3 & 1 & 500 & 0.99  & 0.990 & 423 & 0.357 & 0.357 & 355 & 0.911 & 0.911 & 495 & 0.072 & 0.072 \\
  5 & 3 & 1 & 125 & 0.944 & 0.944 & 10  & 0.878 & 0.878 & 4   & 0.905 & 0.905 & 118 & 0.075 & 0.075 \\
  5 & 3 & 3 & 500 & 0.815 & 0.998 & 159 & 0.619 & 0.940 & 74  & 0.745 & 0.995 & 400 & 0.107 & 0.146 \\
  5 & 3 & 3 & 125 & 0.22  & 0.533 & 6   & 0.206 & 0.516 & 2   & 0.263 & 0.655 & 1   & 0.286 & 0.668 \\[1ex]
  6 & 2 & 1 & 500 & 0.99  & 0.990 & 262 & 0.981 & 0.981 & 179 & 0.980 & 0.980 & 497 & 0.014 & 0.014 \\
  6 & 2 & 1 & 125 & 0.962 & 0.962 & 15  & 0.948 & 0.948 & 6   & 0.954 & 0.954 & 122 & 0.021 & 0.021 \\
  6 & 2 & 3 & 500 & 0.712 & 0.984 & 93  & 0.678 & 0.975 & 78  & 0.672 & 0.989 & 300 & 0.028 & 0.048 \\
  6 & 2 & 3 & 125 & 0.066 & 0.208 & 0   & 0.082 & 0.267 & 3   & 0.104 & 0.313 & 0   & 0.065 & 0.211 \\[1ex]
  7 & 0 & 1 & 500 & 0.998 & 0.998 & 459 & 0.044 & 0.044 & 292 & 0.964 & 0.964 & 500 & 0.009 & 0.009 \\
  7 & 0 & 1 & 125 & 0.923 & 0.923 & 1   & 0.882 & 0.882 & 17  & 0.915 & 0.915 & 125 & 0.010 & 0.010 \\
  7 & 0 & 3 & 500 & 0.882 & 1.000 & 14  & 0.790 & 0.997 & 168 & 0.635 & 0.974 & 489 & 0.009 & 0.018 \\
  7 & 0 & 3 & 125 & 0.178 & 0.574 & 0   & 0.281 & 0.689 & 3   & 0.223 & 0.611 & 108 & 0.005 & 0.014 \\[1ex]
  8 & 6 & 1 & 500 & 0.964 & 0.964 & 491 & 0.001 & 0.001 & 245 & 0.957 & 0.957 & 500 & 0.000 & 0.000 \\
  8 & 6 & 1 & 125 & 0.902 & 0.902 & 0   & 0.894 & 0.894 & 14  & 0.888 & 0.888 & 123 & 0.000 & 0.000 \\
  8 & 6 & 3 & 500 & 0.739 & 0.992 & 3   & 0.751 & 0.994 & 138 & 0.712 & 0.987 & 428 & 0.005 & 0.006 \\
  8 & 6 & 3 & 125 & 0.447 & 0.918 & 0   & 0.493 & 0.939 & 9   & 0.526 & 0.954 & 119 & 0.002 & 0.002 \\
  \bottomrule
\end{tabular}
\caption{
  The number of removed poisoned examples ${\rm p}_{\rm rm}$ under \(m\)-way pixel attacks for various choices of the source label \(\ell_{\mathrm{s}}\) and target label \(\ell_{\mathrm{t}}\).
  Accuracy on clean data was between 91\% and 92.5\% in all experiments and are omitted in the table.
}\label{tab:pixel-vs-labels}
\end{table}

\FloatBarrier
\section{Robust estimation}\label{sec:robust-estimation}

We reproduce details from \cite{diakonikolas2017being} which are relevant to the implementation and usage of \cref{alg:detect} here for completeness.
First, we introduce some notations.
Given two sets \(A\) and \(B\), \(\Delta\del{A, B}\) is the size of their symmetric difference \(\abs{\del{A \setminus B} \cup \del{B \setminus A}}\).
Given a matrix \(M \in \RR^{d\times d}\), we write \(M^\flat\) to denote the flattened vector \(\bm v \in \RR^{d^2}\) built by concatenating the columns of \(M\).
Similarly, given a vector \(\bm v \in \RR^{d^2}\), we write \(v^\sharp\) to denote the matrix \(M \in \RR^{d\times d}\) with \(\bm v_i\) as columns, where \(\bm v\) is split into \(d\) contiguous vectors in \(\RR^d\).

\subsection{Robust mean estimation}\label{sec:robust-mean}

There exists a practical robust mean estimation algorithm \textsc{RobustMean} which is given explicitly in \cref{alg:robust-mean}.
\begin{algorithm2e}[h]
  \caption{Robust mean estimation (\textsc{RobustMean}) \citep{diakonikolas2017being}}\label{alg:robust-mean}
  \DontPrintSemicolon
  \KwIn{A multiset \(S'\) such that there exists an \(\del{\varepsilon, \tau}\)-good set \(S\) with \(\Delta(S, S') < 2\varepsilon\)}
  \KwOut{A vector \(\bm \mu'\) such that \(\norm{\bm \mu' - \bm \mu\del{G}}_2 \le O\del{\varepsilon \sqrt{\log\del{1/\varepsilon}}}\)}
  \Repeat{\textsc{GaussianMeanFilter} returns \(\bm \mu'\)}{
    \(S' \gets \textsc{GaussianMeanFilter}\del{S'}\) \hfill[\cref{alg:robust-mean-filter}]\\
  }
  \Return \(\bm \mu'\)
\end{algorithm2e}

Understanding \cref{alg:robust-covariance} requires the definition of an \(\del{\varepsilon, \tau}\)-good set with respect to a Gaussian, which is given in \cref{def:mean-eps-good-set}.
The key feature of \(\del{\varepsilon, \tau}\)-goodness is that a set of independent samples from the Gaussian of sufficient size is \(\del{\varepsilon, \tau}\)-good with high probability as stated in \cref{prop:mean-eps-good-set}.

\begin{definition}{\citep[Definition~A.4]{diakonikolas2017being}}\label{def:mean-eps-good-set}
  Let \(G\) be a sub-gaussian distribution in \(d\) dimensions with mean \(\bm \mu\del{G}\) and covariance matrix \(I\) and let \(\varepsilon, \tau > 0\).
  We say that a multiset \(S\) of elements in \(\RR^d\) is \(\del{\varepsilon, \tau}\)-good with respect to \(G\) if the following conditions are satisfied:
  \begin{enumerate}
  \item For all \(\bm x \in S\) we have \(\norm{\bm x - \bm \mu\del{G}}_2 \le O\del{\sqrt{d \log\del{\abs{S}/\tau}}}\).
  \item For every affine function \(L : \RR^d \to \RR\) such that \(L\del{\bm x} = \bm v \cdot \del{\bm x - \bm \mu\del{G}} - T\), \(\norm{\bm v}_2 = 1\), we have that
    \[\abs*{\Pr_{X \in_u S}\sbr{L\del{X} \ge 0} - \Pr_{X \sim G}\sbr{L\del{X} \ge 0}} \le \frac{\varepsilon}{T^2\log\del*{d \log\del{\frac{d}{\varepsilon \tau}}}}\]
  \item We have that \(\norm{\bm \mu\del{S} - \bm \mu\del{G}}_2 \le \varepsilon\).
  \item We have that \(\norm{M_s - I}_2 \le \varepsilon\).
  \end{enumerate}
\end{definition}

\begin{lemma}{\citep[Lemma~A.6]{diakonikolas2017being}}\label{prop:mean-eps-good-set}
  Let \(G\) be a sub-gaussian distribution with parameter \(\nu = \Theta\del{1}\) and identity covariance and let \(\varepsilon, \tau > 0\).
  If the multiset \(S\) is obtained by taking \(\Omega\del{\del{d/\varepsilon^2} \operatorname{poly~log}\del{d/\varepsilon\tau}}\) independent samples from \(G\), it is \(\varepsilon\)-good with respect to \(G\) with probability at least \(1 - \tau\).
\end{lemma}

Now we give the definition of the filter used in \cref{alg:robust-mean} in \cref{alg:robust-mean-filter}, which shows that the sets \(S'\) in \cref{alg:robust-mean} approach the \(\varepsilon\)-good set \(S\) with respect to the size of their symmetric difference.

\begin{algorithm2e}[h]
  \caption{Filter algorithm for a Gaussian with unknown mean. \citep[Algorithm~2]{diakonikolas2017being}}\label{alg:robust-mean-filter}
  \DontPrintSemicolon
  \KwIn{A multiset \(S'\) such that there exists an \(\del{\varepsilon, \tau}\)-good set \(S\) with \(\Delta(S, S') < 2\varepsilon\)}
  \KwOut{Either a set \(S''\) with \(\Delta(S, S'') \le \Delta(S, S') - \varepsilon/\alpha\) where \(\alpha \triangleq d \log\del{d/\varepsilon \tau}\log\del{d \log\del{d/\varepsilon \tau}}\) or a vector \(\bm \mu\) satisfying \(\norm{\bm \mu' - \bm \mu\del{G}}_2 \le O\del{\varepsilon \sqrt{\log\del{1/\varepsilon}}}\)}
  Compute the sample mean \(\bm \mu\del{S'} = \EE_{X \sim \operatorname{Unif}\del{S'}}\sbr{X}\).\;
  Compute the sample covariance matrix \(\Sigma\del{S'} = \EE_{X \in \operatorname{Unif}\del{S'}}\sbr{\del{X - \bm\mu\del{S'}}\del{X - \mu\del{S'}}^\top}\).\;
  Compute an approximation of the largest absolute eigenvalue of \(\Sigma - I\), \(\lambda^* \approx \norm{\Sigma - I}_2\) and an approximate associated eigenvector \(\bm v^*\).\;
  \uIf{\(\lambda^* \le O\del{\varepsilon \log\del{1/\varepsilon}}\)}{
    \Return \(\bm\mu\del{S'}\)\;
  }
  Let \(\delta = 3\sqrt{\varepsilon \lambda^*}\).
  Find a \(T > 0\) such that
  \[\Pr_{X \in \operatorname{Unif}\del{S'}}\del*{\abs{\bm v^* \cdot \del{X - \bm \mu\del{S'}}} > T + \delta} > 8\exp\del*{-\frac{T^2}{2\nu}} + \frac{8\varepsilon}{T^2 \log\del{d \log\del{\frac{d}{\varepsilon\tau}}}}.\]
  \Return \(S'' = \set{\bm x \in S'' : \abs{\bm v^*\cdot\del{\bm x - \bm\mu\del{S'}}} \le T + \delta}\)\;
\end{algorithm2e}

\FloatBarrier
\subsection{Robust covariance estimation}\label{sec:robust-covariance}

The structure of this subsection mirrors that of \cref{sec:robust-mean}.
\cref{thm:robust} states the existence of a practical robust covariance estimation algorithm \textsc{RobustCov} which is given explicitly in \cref{alg:robust-covariance}.

\begin{algorithm2e}[h]
\caption{Robust covariance estimation (\textsc{RobustCov}) \citep{diakonikolas2017being}}\label{alg:robust-covariance}
\DontPrintSemicolon 
% \KwIn{representation $S=\{\tilde{\bm h}_i\in{\mathbb R}^k\}_{i=1}^n$, poison fraction $\varepsilon$  } 
\KwIn{A multiset \(S'\) such that there exists an \(\varepsilon\)-good set \(S\) with \(\Delta(S, S') < 2\varepsilon\)}
\KwOut{A matrix \(\Sigma'\) such that \(\norm{I - \Sigma^{-1/2}\Sigma'\Sigma^{-1/2}}_F= O\del{\varepsilon\log\del{\frac{1}{\varepsilon}}}\)}
\Repeat{\textsc{GaussianCovarianceFilter} returns \(\Sigma'\)}{
    \(S' \gets \textsc{GaussianCovarianceFilter}\del{S'}\) \hfill[\cref{alg:robust-covariance-filter}]\\
}
\Return \(\Sigma'\)
\end{algorithm2e}

Understanding \cref{alg:robust-covariance} requires the definition of an \((\varepsilon\)-good set with respect to a Gaussian, which is given in \cref{def:cov-eps-good-set}.
The key feature of \(\varepsilon\)-goodness is that a set of independent samples from the Gaussian of sufficient size is \(\varepsilon\)-good with high probability as stated in \cref{prop:cov-eps-good-set}.

\begin{definition}{\citep[Definition~A.27]{diakonikolas2017being}}\label{def:cov-eps-good-set}
    Let \(G\) be a Gaussian in \(\RR^d\) with mean \(0\) and covariance \(\Sigma\).
    Let \(\varepsilon > 0\) be sufficiently small.
    We say that a multiset \(S\) of points in \(\RR^d\) is \(\varepsilon\)-good with respect to \(G\) if the following hold:
    \begin{enumerate}
        \item For all \(\bm x \in S\), \(\bm x^\top \Sigma^{-1}\bm x < d + O\del{\sqrt{d} \log\del{d/\varepsilon}}\).
        \item We have that \(\norm{\Sigma^{-1/2} \cov\del{S} \Sigma^{-1/2} - I}_F = O\del{\varepsilon}\).
        \item For all even degree-\(2\) polynomials \(p\), we have that \(\var\del{p\del{\bm x}} = \var\del{p\del{G}}\del{1 + O\del{\varepsilon}}\).
        \item For \(p\) an even degree-\(2\) polynomial with \(\EE\sbr{p\del{G}} = 0\) and \(\var\del{p\del{G}} = 1\), and for any \(T > 10 \log\del{1/\varepsilon}\) we have that
        \[\Pr\del{\abs{p\del{x}} > T} \le \frac{\varepsilon}{T^2 \log^2\del{T}}.\]
    \end{enumerate}
\end{definition}

\begin{proposition}{\citep[Proposition~A.28]{diakonikolas2017being}}\label{prop:cov-eps-good-set}
    Let \(N\) be a sufficiently large constant multiple of \(\del{d^2/\varepsilon^2}\log^5\del{d/\varepsilon}\). 
    Then a set \(S\) of \(N\) independent samples from \(G\) is \(\varepsilon\)-good with respect to \(G\) with high probability.
\end{proposition}

Now we give the definition of the filter used in \cref{alg:robust-covariance} in \cref{alg:robust-covariance-filter}, which shows that the sets \(S'\) in \cref{alg:robust-covariance} approach the \(\del{\varepsilon, \tau}\)-good set \(S\) with respect to the size of their symmetric difference.

\begin{algorithm2e}[h]
  \caption{Filter algorithm for a Gaussian with unknown covariance matrix. \citep[Algorithm~4]{diakonikolas2017being}}\label{alg:robust-covariance-filter}
  \SetKwFunction{alg}{FILTER-GAUSSIAN-UNKNOWN-COVARIANCE}
  \SetKwFunction{fmp}{FIND-MAX-POLY}
  \DontPrintSemicolon
    \KwIn{A multiset \(S'\) such that there exists an \(\varepsilon\)-good set \(S\) with \(\Delta(S, S') < 2\varepsilon\)}
    \KwOut{Either a set \(S''\) with \(\Delta(S, S'') < \Delta(S, S')\) or a matrix \(\Sigma'\) such that \(\norm{I - \Sigma^{-1/2}\Sigma'\Sigma^{-1/2}}_F= O\del{\varepsilon\log\del{\frac{1}{\varepsilon}}}\)}
    Let \(C, C' > 0\) be sufficiently large universal constants.\;
    \(\Sigma' \gets \EE_{X \in S'}\sbr{XX^\top}\)\;
    \(G' \gets \mathcal N(0, \Sigma')\)\;
    \uIf{there exists an \(\bm x \in S'\) such that \(\bm x^\top \Sigma'^{-1} \bm x \ge Cd \log\del{10\abs{S'}}\) }{
      \Return \(S'' = S' \setminus \set{\bm x \in S': \bm x^\top \Sigma'^{-1}  \bm x> Cd \log\del{10\abs{S'}}}\)\;
    }
    Let \(L\) be the space of even degree-\(2\) polynomials \(p: \RR^k \to \RR\) such that \(\EE_{X \sim G'}\sbr{p(X)} = 0\).\;
    Define two quadratic forms on \(L\):
    \begin{enumerate}[label=(\roman*)]
        \item \(Q_{G'}(p) = \EE_{X \sim G'}\sbr{p^2(X)}\)
        \item \(Q_{S'}(p) = \EE_{X \sim \operatorname{Unif}\del{S'}}\sbr{p^2(X)}\)
    \end{enumerate}
    Compute \(\max_{p \in L\setminus\set{0}} Q_{S'}(p)/Q_{G'}(p)\) and the associated polynomial \(p^*(x)\) normalized such that \(Q_{G'}(p) = 1\) using \cref{alg:findmaxpoly}.\;
    %\(p^*, \lambda^* \gets \fmp(S', G')\)\;
    \uIf{\(Q_{S'}(p^*) \le \del{1 + C\varepsilon \log^2\del{1/\varepsilon}}Q_{G'}(p^*)\)}{
      \Return \(\Sigma'\)\;
    }
    \(\mu\gets \text{the median value of \(p^*(X)\) over \(X \in S'\)}\)\;
    Find a \(T > C'\) such that
    \[\Pr_{X \in T'}\del{\abs{p^*(X) - \mu} \ge 3} \le \operatorname{Tail}(T, d, \varepsilon),\]
    where
    \[
      \operatorname{Tail}(T, d, \varepsilon)=
      \begin{cases}
        3\varepsilon/\del{T^2 \log^2(T)} & \text{if \(T \ge 10 \ln\del{1/\varepsilon}\)}\\
        1 & \text{otherwise}
      \end{cases}.
    \]
    \Return \(S'' = \set{\bm x \in S'' : \abs{p^*(U'^\top \bm x) - \mu} \le T}\)\;
\end{algorithm2e}

\begin{algorithm2e}
\caption{Algorithm to compute the polynomial with maximum variance relative to a Gaussian \citep[Algorithm~4]{diakonikolas2017being}}\label{alg:findmaxpoly}
\DontPrintSemicolon 
    \KwIn{A multiset \(S' = \set{\bm x_i}_{i=1}^n \subset \RR^d\) and a Gaussian \(G' = \mathcal N(0, \Sigma')\)}
    \KwOut{The even degree-\(2\) polynomial \(p^*(\bm x)\) with \(\EE_{X \sim G'}\sbr{p(X)} \approx 0\) and \(Q_{G'}(p^*) \approx 1\) that approximately maximizes \(Q_{S'}(p^*)\) and this maximum is \(\lambda^* = Q_{S'}(p^*)\)}
    \For{\(i \in \sbr{n}\)}{
      \(\bm y_i \gets \Sigma'^{-1/2}_k\bm x_i\)\;
      \(\bm z_i \gets \del{\bm y_i\bm y_i^\top}^\flat\)\;
    }
    \(T_{S'} \gets -I^\flat I^{\flat\top} + \frac{1}{\abs{S'}}\sum_{i=1}^n \bm z_i\bm z_i^\top \)\;
    Approximate the top eigenvalue \(\lambda^*\) and eigenvector \(\bm v^*\) of \(T_{S'}\)\;
    \(p*(x) \gets \frac{1}{\sqrt{2}}\del{\del{\Sigma'^{-1/2}_k\bm x}v^{*\sharp}\del{\Sigma'^{-1/2}_k\bm x} - \trace\del{\bm v^{*\sharp}}} \)\;

    \Return \(p^*\) and \(\lambda^*\)
  
\end{algorithm2e}

Note that a naive implementation of \cref{alg:findmaxpoly} requires \(\Omega\del{nd^2}\) space to store the \(\bm y_i\) and \(\Omega\del{d^4}\) space to store \(T_{S'}\).
Additionally, the matrix multiplication performed by OpenBLAS to produce \(T_{S'}\) requires \(\Omega\del{nd^4}\) time.
By representing the linear operator \(T_{S'}\) implicitly, we can reduce these requirements substantially.
First, the product \(-I^\flat (I^{\flat\top}\bm v)\) can be computed in \(O\del{d^2}\) time and space. 
Next, if \(Y\) and \(Z\) are the matrices with columns \(\bm y_i\) and \(\bm z_i\) respectively, then \(Z\) is the Khatri-Rao product \(Y \odot Y\).
This means we can use the vec tricks for the Khatri-Rao and transpose Khatri-Rao vector products of \cite{perivsa2017recompression} to calculate \(ZZ^\top \bm v\) in \(O\del{nd^2}\) time and \(O\del{nd + d^2}\) space.
We can then calculate the eigenvector \(\bm v^*\) of the implicitly represented linear operator \(T_{S'}\) using Krylov methods, requiring the evaluation of a small number of products \(T_{S'}\bm v\). 
For our experiments, this provided a speedup of several orders of magnitude and a substantial reduction in the required amount of system memory versus the naive implementation.

\FloatBarrier
\subsection{Robust joint mean and covariance estimation}

Note that \cref{alg:robust-mean} requires the inputs to have identity covariance and \cref{alg:robust-covariance} requires the inputs to have zero mean.
Here we show how to combine them to estimate both the mean and covariance of an arbitrary Gaussian, as described in \citep[Section~4.5]{diakonikolas2017being}.
The key idea is to split the dataset into two halves, pair off samples from each half, and subtract them. 
The resulting vectors have zero mean and double the original covariance.
This allows us to use \cref{alg:robust-covariance} to whiten the samples, which then allows us to use \cref{alg:robust-mean}.
We reproduce the exact procedure in \cref{alg:robust-gaussian}.

\begin{algorithm2e}
\caption{Algorithm to robustly learn an arbitrary Gaussian {\citep[Algorithm~6]{diakonikolas2019robust}}}\label{alg:robust-gaussian}
\DontPrintSemicolon 
    \KwIn{A multiset \(S' = \set{\bm x_i}_{i=1}^n \subset \RR^d\), corruption fraction \(\varepsilon\)}
    \KwOut{A matrix \(\Sigma'\) such that \(\norm{I - \Sigma^{-1/2}\Sigma'\Sigma^{-1/2}}_F= O\del{\varepsilon\log\del{\frac{1}{\varepsilon}}}\) and Aavector \(\bm \mu'\) such that \(\norm{\bm \mu' - \bm \mu\del{G}}_2 \le O\del{\varepsilon \sqrt{\log\del{1/\varepsilon}}}\)}

    \For{\(i \in \sbr{\floor{n/2}}\)}{
      \(\bm x_i' \gets \del{\bm x_i - \bm x_{\floor{n/2} + 1}} / \sqrt{2}\)\;
    }
    \(\widehat{\Sigma} \gets \textsc{RobustCov}\del{\set{\bm x_i'}, \varepsilon}\) \hfill[\cref{alg:robust-covariance}]\\
    \For{\(i \in \sbr{n}\)}{
      \(\bm x_i'' \gets \widehat{\Sigma}^{-1/2}\bm x_i\)\;
    }
    \(\bm{\widehat{\mu}} \gets \textsc{RobustMean}\del{\set{\bm x_i''}, \varepsilon}\) \hfill[\cref{alg:robust-mean}]\\
    \Return \(\widehat{\Sigma}\) and \(\widehat{\Sigma}^{1/2}\widehat{\bm \mu}\)\;
\end{algorithm2e}

\FloatBarrier
\section{Experiment details}\label{sec:exp_detail}

For each poisoned dataset, we performed one training run to produce each poisoned model.
For the pixel and periodic attacks, we performed one retraining run for each defense.
Training for our experiments was done on a server with a Xeon Gold 6230 CPU and eight Nvidia 2080~Ti GPUs.
The training and retraining for our experiments took approximately 100 GPU hours.
Running all defences for our experiments took approximately 200 CPU-core hours.
Using the thermal design power of these components to estimate of our required power, we estimate that our experiments required a total of \SI{28}{\kilo\watt\hour} of energy.

\subsection{\(m\)-way pixel attacks}
%Some examples of pixel attacks are shown in \cref{fig:pixel-examples}.

For pixel attacks, we reproduce the experimental setup of \cite{tran2018spectral}.
For our ResNet-32, we used a leaky ReLU with a negative slope of 0.1 for the nonlinearity and trained it using stochastic gradient descent with momentum for 200 epochs, dividing the learning rate by 10 every 75 epochs.
Both data standardization and augmentation were used.

Although a fixed pixel is used for watermarking, data augmentation may ensure that the network is sensitive to pixels of the chosen color at multiple locations in the image.
Using the standard random horizontal flip and random crop with 4 pixels of padding used for CIFAR-10, the pixel may end up in as many as \(9 \times 9 \times 2 = 162\) distinct pixels in the transformed image, representing about 16\% of the image's total area.

To implement an \(m\)-way pixel attack, \(m\) pairs of locations and colors are chosen.
Only one of the \(m\) pixels is used for each poisoned training example, but all \(m\) are used simultaneously at test time.
We ran experiments for \(m \in \set{1, 2, 3}\).
We used the same backdoor pixel \citeauthor{tran2018spectral} used for their experiments, along with two more arbitrarily chosen.
The exact locations and colors are shown in \cref{tab:pixels}.

\begin{table}[h]
    \centering
    \begin{tabular}{lll}
        \toprule
        \(m\) & location & color\\
        \midrule
        1 & (11, 16) & \#650019\\
        2 & (5, 27) & \#657B79 \\
        3 & (30, 7) & \#002436\\
        \bottomrule
    \end{tabular}
    \caption{
      Pixel watermarks used for the \(m\)-way pixel attacks.
      Location is a pixel coordinate in \((x, y)\) format and color is a 24-bit hexadecimal color in HTML format.
      %Note that the data normalization we  use in our training/testing  changes the color values fed into the model.
      %Note that the network is trained on images after their channels have been normalized, so it will receive the corresponding transformed values, although the normalization transformation remains constant. 
    }\label{tab:pixels}
\end{table}

\subsection{\(m\)-way periodic attacks}

For periodic attacks, we used the same network architecture and training environment used for pixel attacks.
Although the phase of the signal is fixed for watermarking, the signal will be shifted by a random amount at training time due to the random flip and random crop and pad, in a manner similar to the pixel attack.
Because our signals have a period of 4 pixels, which equals the maximum translation produced by the data augmentation, the backdoored network should be sensitive to signals with any phase.

% The poisoned examples are created by combining the clean image with an additive periodic signal in image space.
% The label is then changed from the source label to the target label.
% Some examples are shown in .
% which seemed to balance the stealthiness and effectiveness of the attack.
% Since the addition may cause the image values to go above 255 or below 0, we also clip the values to the valid range.
% At training, each poisoned sample is corrupted by a single periodic signal to hide the spectral signature and make it challenging to detect. At testing, we combine all $m$ triggers to boost the spectral signature and improve the poison target accuracy.
% Only one periodic signal is used for each poisoned training example, but all periodic signals are added simultaneously at test time.

\subsection{Label consistent attacks}

For label consistent attacks we used the experimental setup of \cite{turner2019label} which is provided at \url{https://github.com/MadryLab/label-consistent-backdoor-code}.
The setup of \cite{turner2019label} appears to be very similar to that of \cite{tran2018spectral}.
The same ResNet-32 architecture is used, albeit with a normal (i.e. not leaky) ReLU. 
Data standardization was enabled by default.
Data augmentation was disabled by default, but we enabled it to ensure greater consistency with our previous experiments.
We also enabled patch placement on all four corners to ensure the watermark would not be cropped out.
For this family of attacks, we did not make any changes to the training system of \cite{turner2019label}, which does not provide retraining.

% \FloatBarrier
% \section{Identifying the effective dimension \(k\)}\label{sec:app-k}

% To choose a reasonable value of \(k\) automatically, we compute the mean QUE scores of the representations after whitening using the covariance of the cleaned data in a fixed subspace of dimension \(k_{\max} = 100\) and choose the \(k\) which maximizes this mean.
% The exact algorithm is shown in \cref{alg:k-finder}.
% The QUE scores are calculated using a separate, fixed subspace to ensure that the scores are comparable as \(k\) varies, unlike the QUE scores used to remove outliers in \cref{alg:detect}.

% \FloatBarrier
% \section{Target label identifier}

% To identify the target label, we wrap \cref{alg:k-finder} in an outer loop to find the \(\del{k, \text{label}}\) pair which maximizes the mean QUE scores of the dataset after whitening.
% The exact algorithm is shown \cref{alg:target}.

% ------------------------------
% \FloatBarrier
% \section{Analysis  of parameter choices}
% \begin{itemize}
%     \item sensitivity to 
%     $1.5\varepsilon n$
% \end{itemize}
 
% \begin{itemize}
%     \item Ablation: robust covariance vs. covariance
%     \item Ablation: low-dimensional: 
% \end{itemize}

% comparisons
% \begin{itemize}
%     \item post-processing / clustering
%     \item architecture / training
%     \item choice of target labels 
%     \item choice of dataset ?? 
%     \item Gaussianity / assumptions confirm through data.
%     \item 4-th moment bounded robust covariance estimation
% \end{itemize} 

\FloatBarrier
\section{Analysis of poisoned representations}\label{sec:analysis-reps}
\begin{figure*}[h]
  \centering
  \includegraphics[width=\textwidth, trim=0 1em 0 1em]{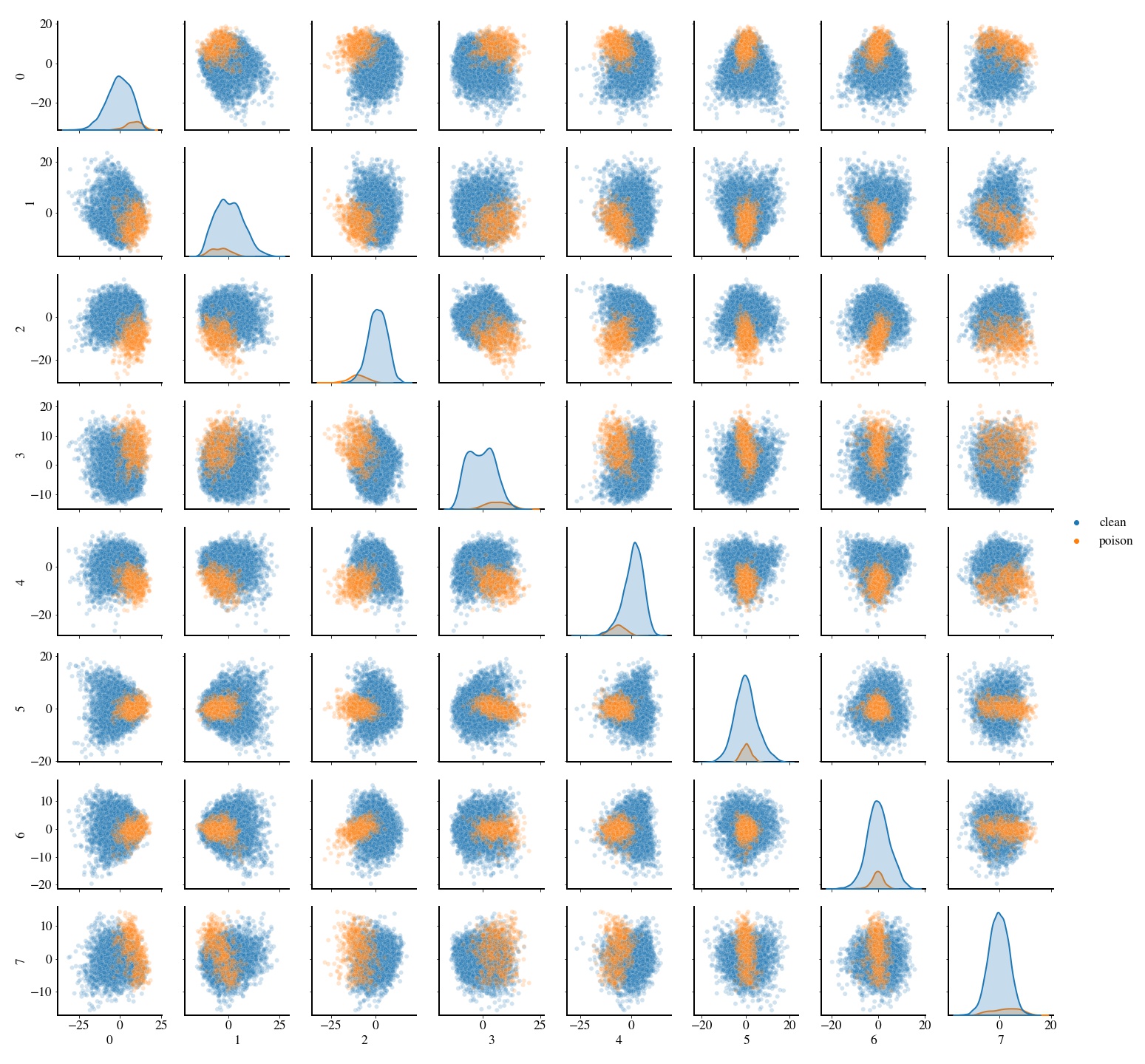}
  \caption{
    Scatter plots of the representations of the \(3\)-way pixel attack with \(\varepsilon = 0.1\) before any whitening.
    The whitened representations are projected onto their top eight PCA directions.
    Plots along the diagonal are Gaussian kernel density estimate plots after projecting onto that PCA direction (of the combined data including the representations of both the poisoned and the clean samples).
    Off-diagonal plots are scatter plots of the data projected onto the subspace spanned by the corresponding pair of PCA directions.
    This shows that the poisoned samples (in orange) are not separable from the clean ones (in blue), if we only focus on these top PCA directions; the spectral signature is hidden. We propose using robust covariance estimation to fine the approximate covariance of clean data and whiten the entire data with the estimated covariance. This enhances the spectral signature as we show in the next figure. 
  }\label{fig:rep-analysis-pairplot-3xp500-pca}
\end{figure*}

\begin{figure*}[h]
  \centering
  \includegraphics[width=\textwidth, trim=0 1em 0 1em]{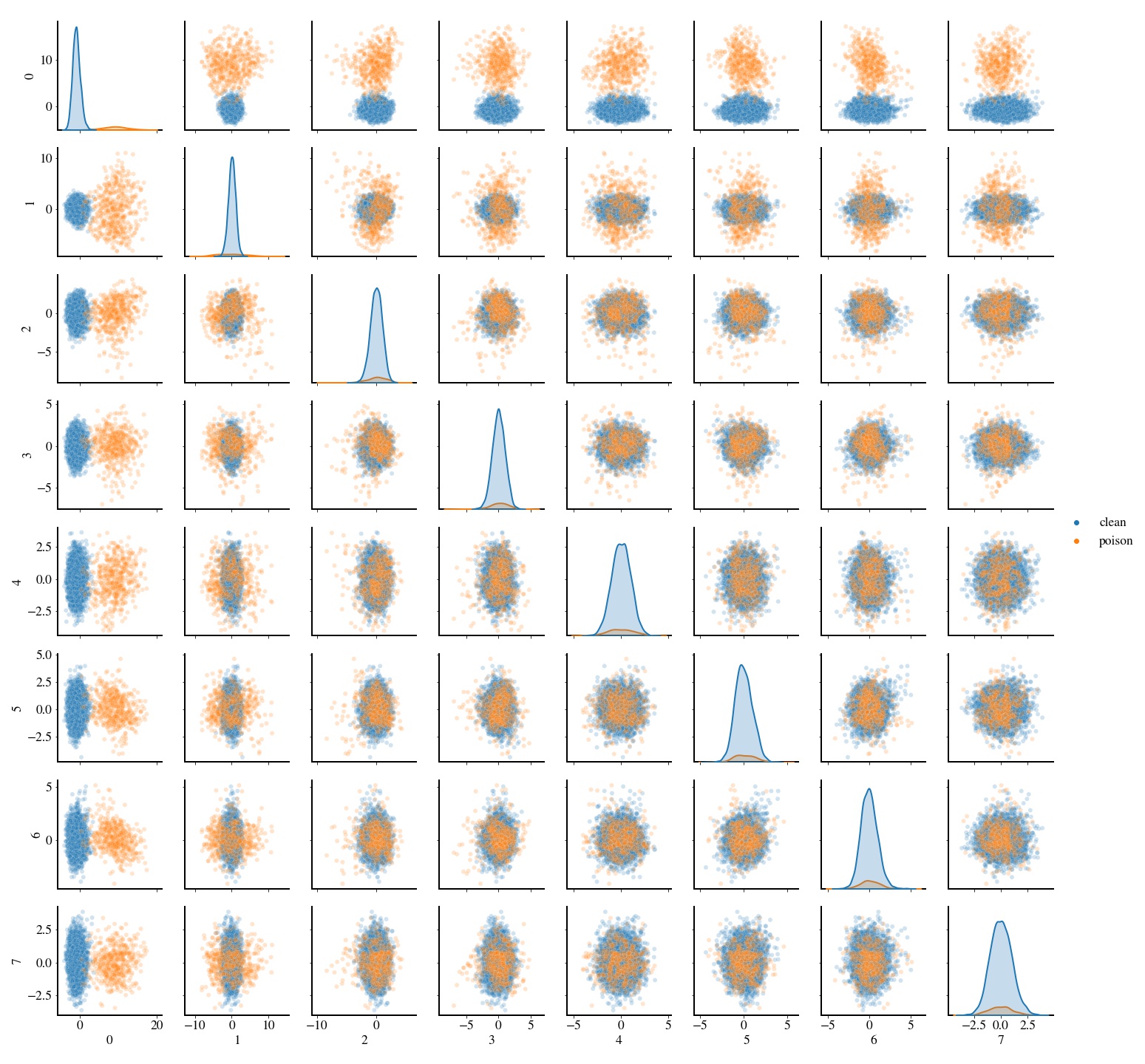}
  \caption{
    Scatter plots of the representations of the \(3\)-way pixel attack with \(\varepsilon = 0.1\) after whitening using the covariance of the clean samples.
    The whitened representations are projected onto their top eight PCA directions.
    Plots along the diagonal are Gaussian kernel density estimate plots after projecting onto that PCA direction (of the combined data including the representations of both the poisoned and the clean samples).
    Off-diagonal plots are scatter plots of the data projected onto the subspace spanned by the corresponding pair of PCA directions.
    This shows that the poisoned samples (in orange) are now separable from the clean ones (in blue) using the top PCA direction after whitening, for example.
  }\label{fig:rep-analysis-pairplot-3xp500-true}
\end{figure*}

\begin{figure*}[h]
  \centering
  \includegraphics[width=\textwidth, trim=0 1em 0 1em]{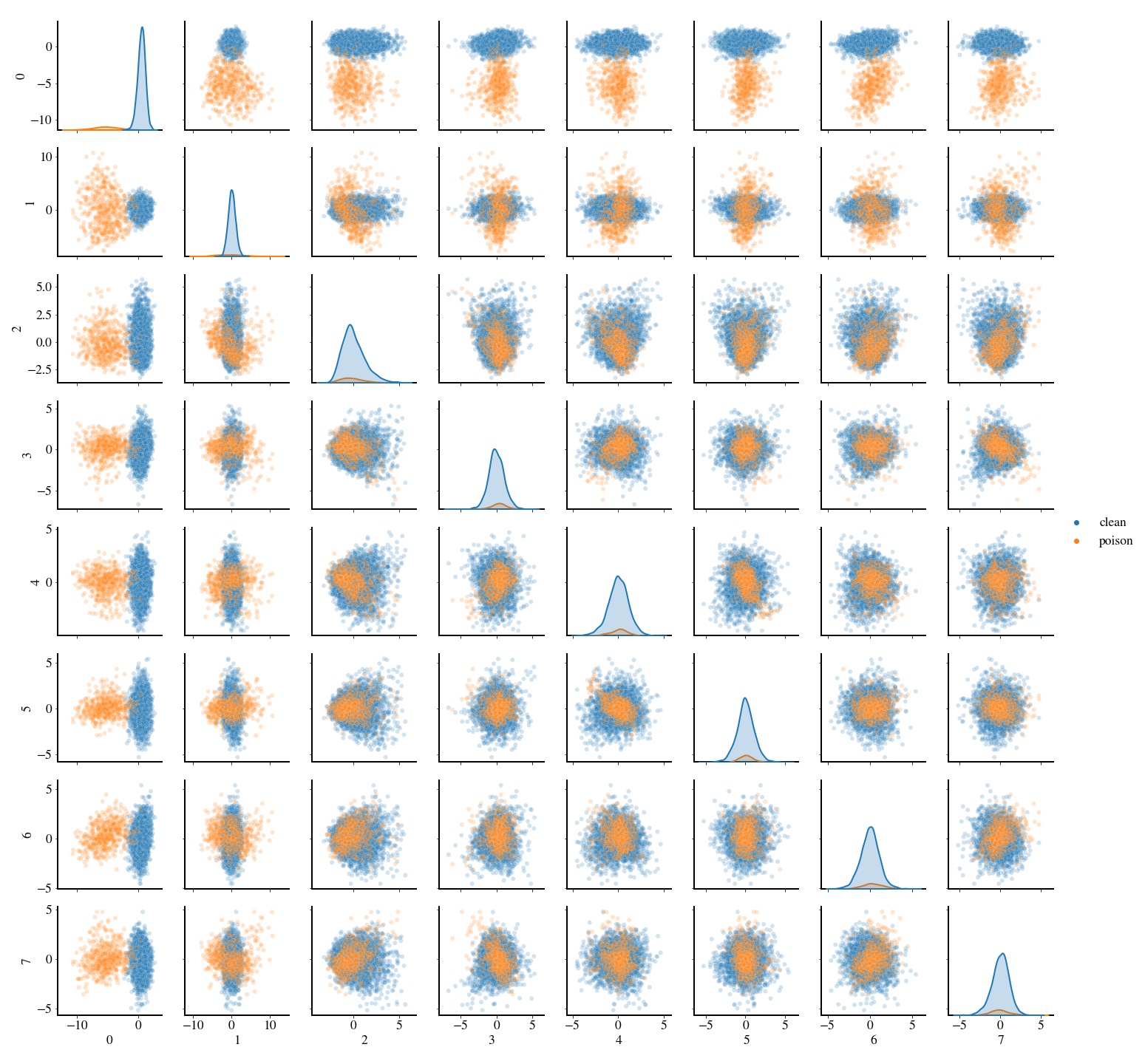}
  \caption{
    Scatter plots of the representations of the \(3\)-way pixel attack with \(\varepsilon = 0.1\) after whitening using the robustly estimated covariance.
    The whitened representations are projected onto their top eight PCA directions.
    Plots along the diagonal are Gaussian kernel density estimate plots after projecting onto that PCA direction (of the combined data including the representations of both the poisoned and the clean samples).
    Off-diagonal plots are scatter plots of the data projected onto the subspace spanned by the corresponding pair of PCA directions.
    This shows that the poisoned samples (in orange) remain separable from the clean ones (in blue) even when whitening using the estimated covariance instead of the true covariance of the clean samples.
  }\label{fig:rep-analysis-pairplot-3xp500-estimated}
\end{figure*}

\begin{figure*}[h]
  \centering
  \includegraphics[width=\textwidth, trim=0 1em 0 1em]{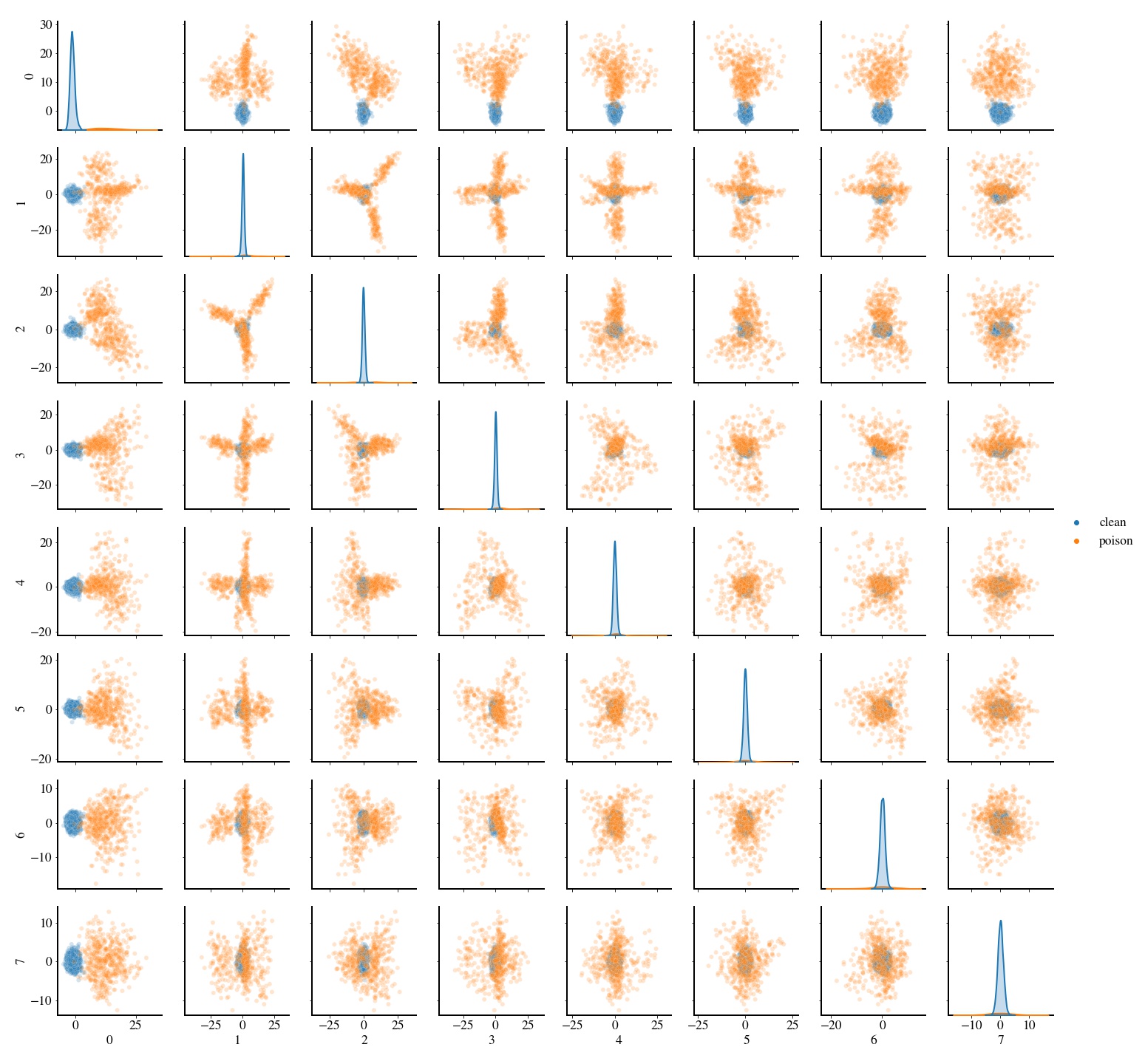}
  \caption{
    Scatter plots of the representations of the \(2\)-way pixel attack with \(\varepsilon = 0.1\) after whitening with the true covariance of the  representation of the clean samples.
    The whitened representations are projected onto their top eight PCA directions.
    Plots along the diagonal are Gaussian kernel density estimate plots after projecting onto that PCA direction.
    Off-diagonal plots are scatter plots of the data projected onto the subspace spanned by the corresponding pair of PCA directions.
    This shows that the poisoned samples (in orange) have split into multiple distinct clusters, resulting in a weakened  spectral signature. Nevertheless, whitening enhances the spectral signature and bring the direction of separation to the top principal components. 
  }\label{fig:rep-analysis-pairplot-2xp500-true}
\end{figure*}

Here we include \cref{fig:rep-analysis-pairplot-3xp500-pca,fig:rep-analysis-pairplot-3xp500-true,fig:rep-analysis-pairplot-3xp500-estimated,fig:rep-analysis-pairplot-2xp500-true}, which illustrate some relevant properties of the hidden layer activations of examples bearing the target layer under a successful backdoor poisoning attack.

\FloatBarrier
\section{Analysis of QUE scores}\label{sec:analysis_que}

In \cref{sec:alg-que}, we showed that the squared norm scoring \(\tau_i^{(0)}=\norm{\tilde{\bm h}_i}^2\) and squared projected norm scoring \(\tau_i^{(\infty)}= \abs{\inner{ v,\tilde{\bm h}_i}}^2\) can both fail under certain conditions.
Here we will explain those conditions in greater detail.

\begin{figure*}[h]
    \centering
    \begin{subfigure}[t]{0.49\linewidth}
        \centering
        \includegraphics[width=0.9\linewidth]{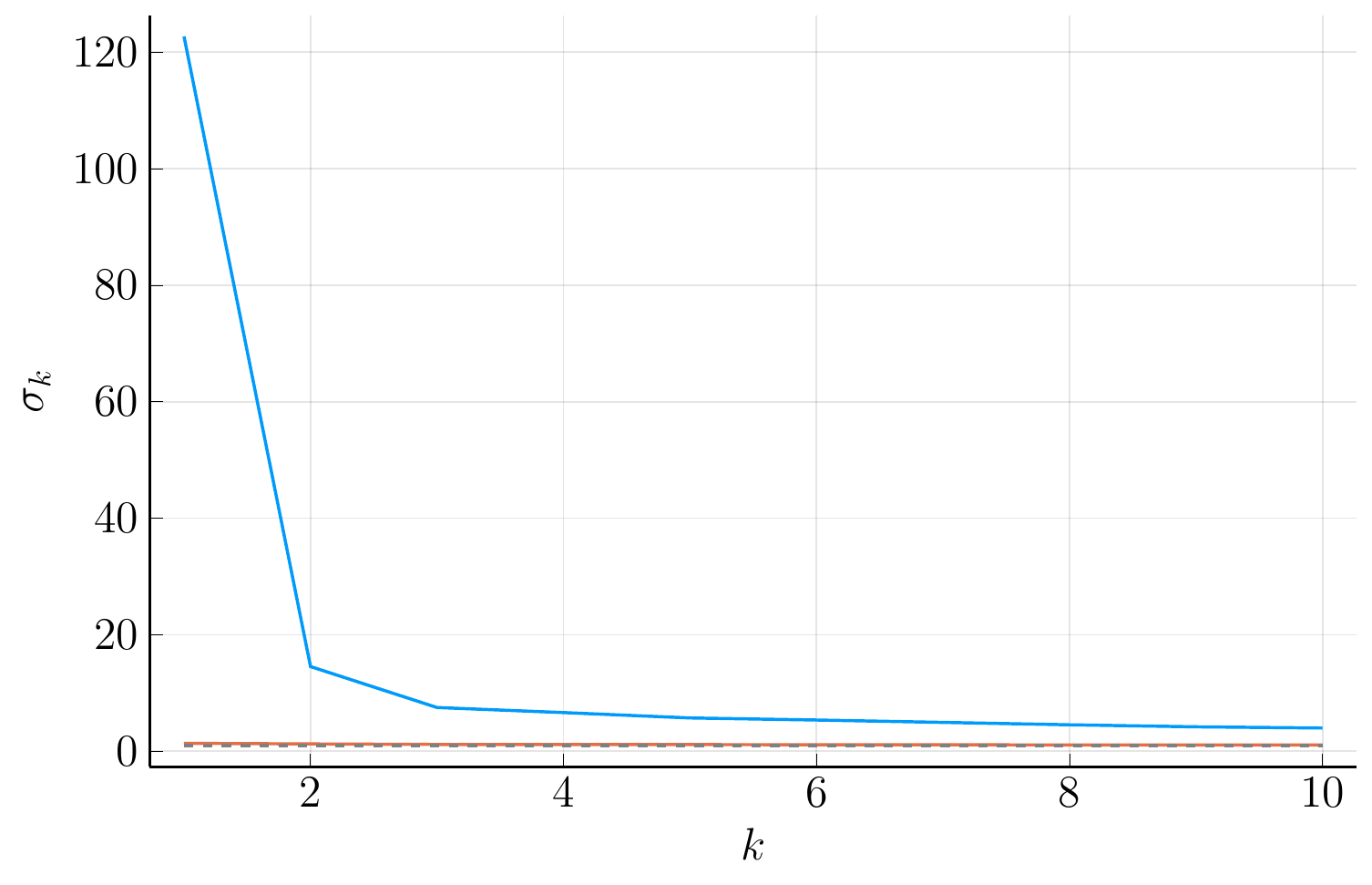}
        \caption{\(3\)-way attack with \(\varepsilon = 0.0124\), top 10 singular values}
        \label{fig:que-svdvals-3xp62-10}
    \end{subfigure}%
    \begin{subfigure}[t]{0.49\linewidth}
        \centering
        \includegraphics[width=0.9\linewidth]{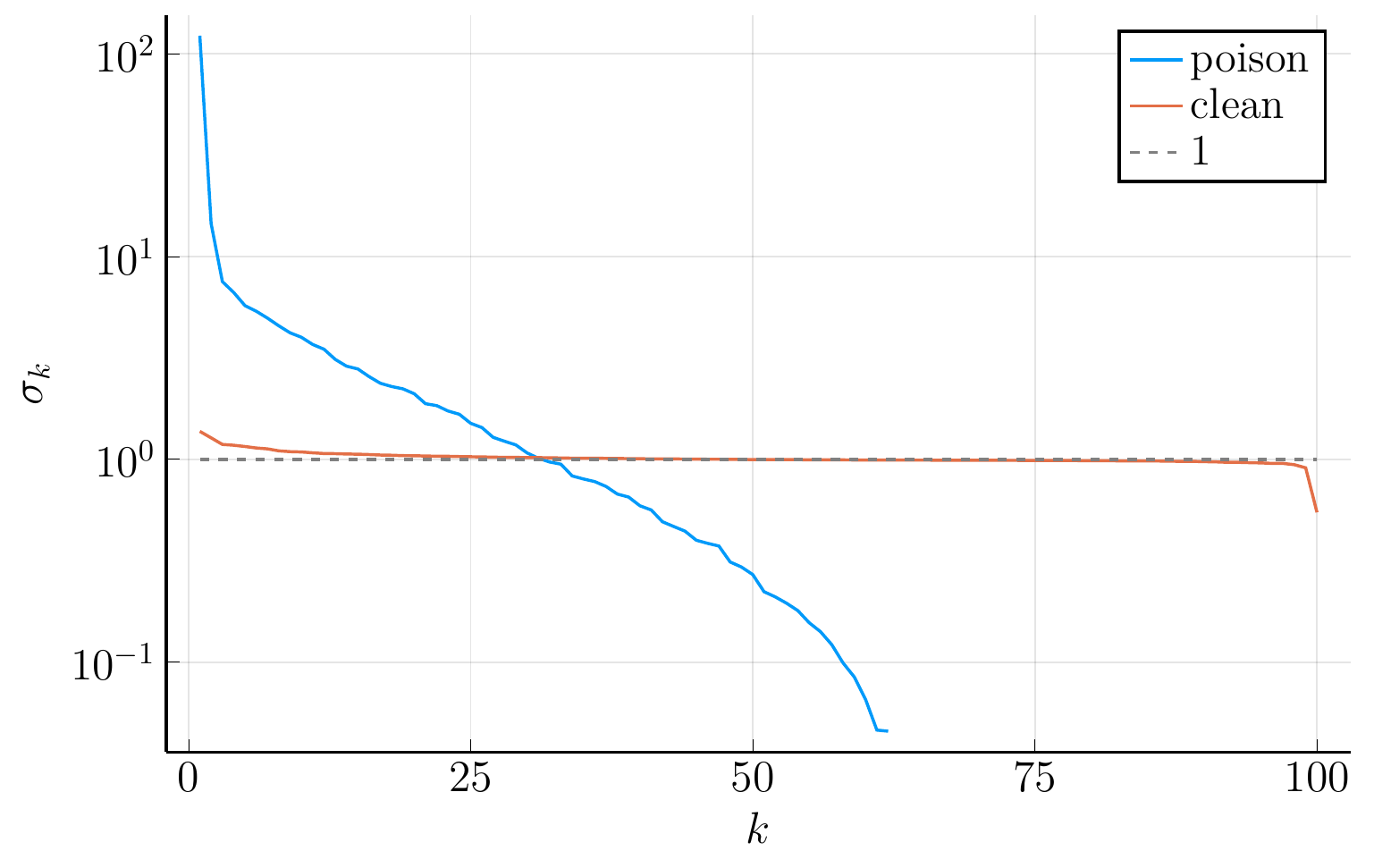}
        \caption{\(3\)-way attack with \(\varepsilon = 0.0124\), top 100 singular values}
        \label{fig:que-svdvals-3xp62-100}
    \end{subfigure}
    \begin{subfigure}[t]{0.49\linewidth}
        \centering
        \includegraphics[width=0.9\linewidth]{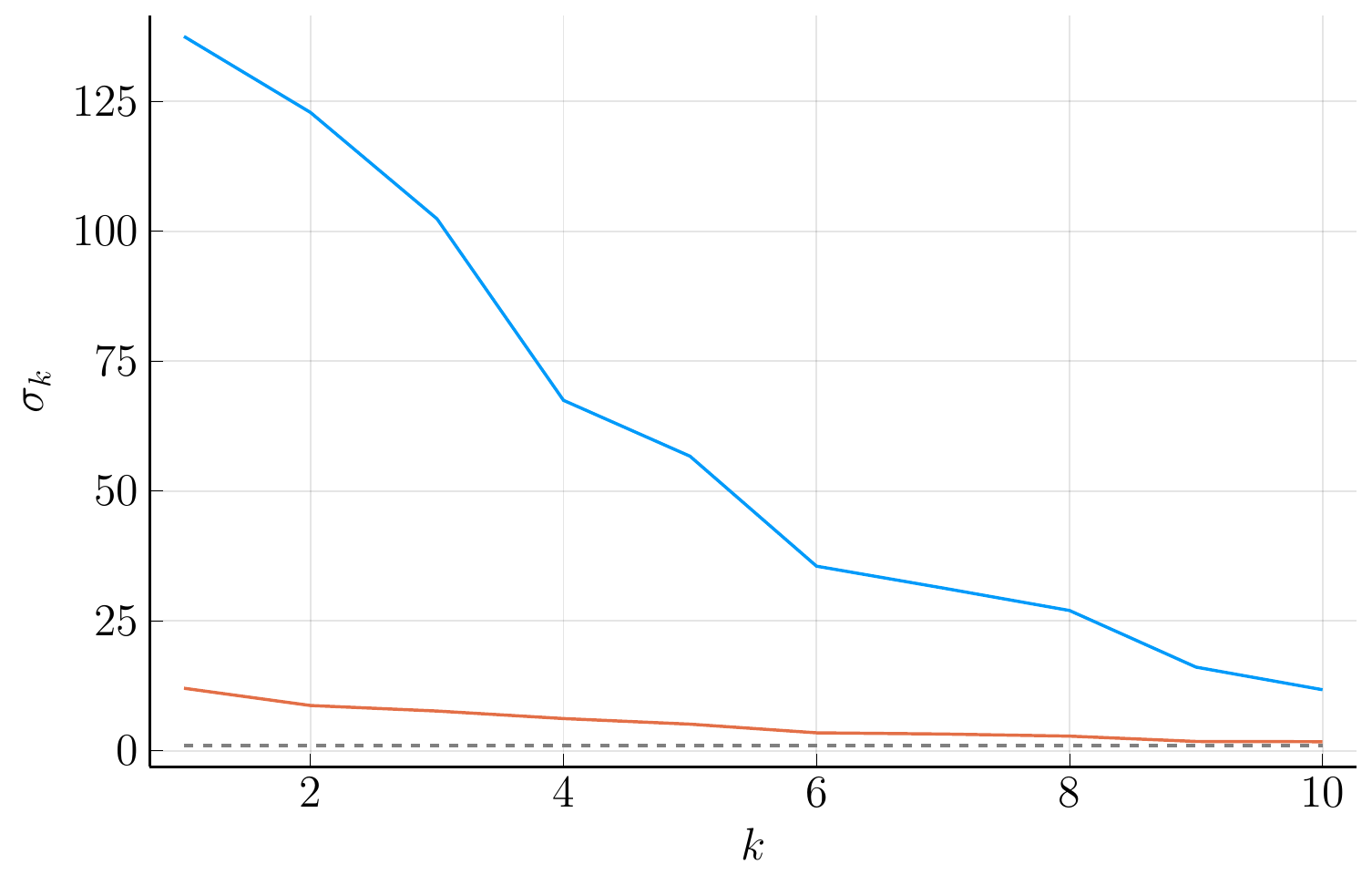}
        \caption{\(1\)-way attack with \(\varepsilon = 0.1\), top 10 singular values}
        \label{fig:que-svdvals-1xp500-10}
    \end{subfigure}%
    \begin{subfigure}[t]{0.49\linewidth}
        \centering
        \includegraphics[width=0.9\linewidth]{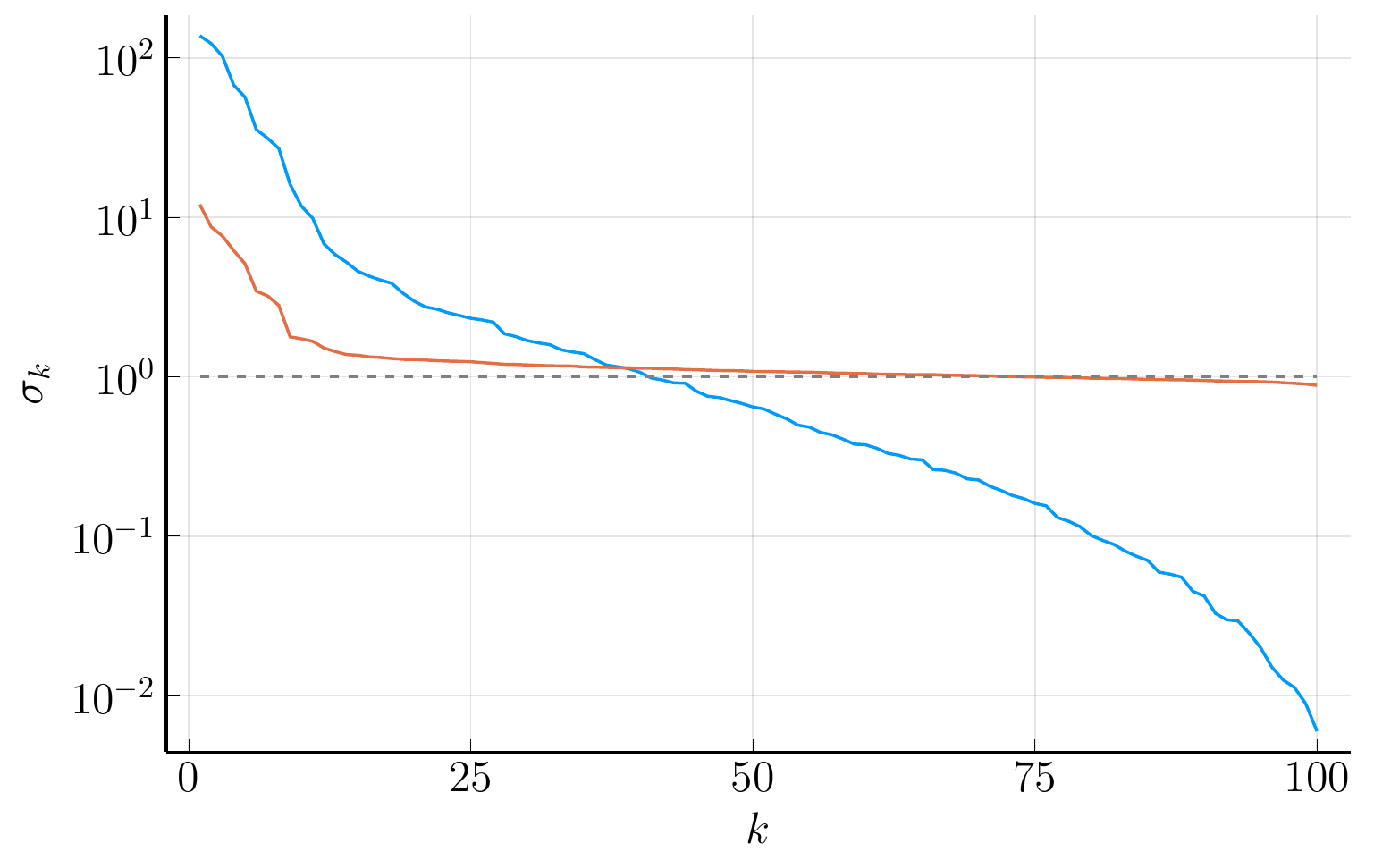}
        \caption{\(1\)-way attack with \(\varepsilon = 0.1\), top 100 singular values}
        \label{fig:que-svdvals-1xp500-100}
    \end{subfigure}
    \caption{
      Plots of the top 10/100 singular values of the covariances of the poison and clean representations after whitening with the robustly estimated covariance in order of decreasing magnitude.
    }\label{fig:que-svdvals}
\end{figure*}

\begin{figure*}[h]
  \centering
  \includegraphics[width=\textwidth, trim=0 1em 0 1em]{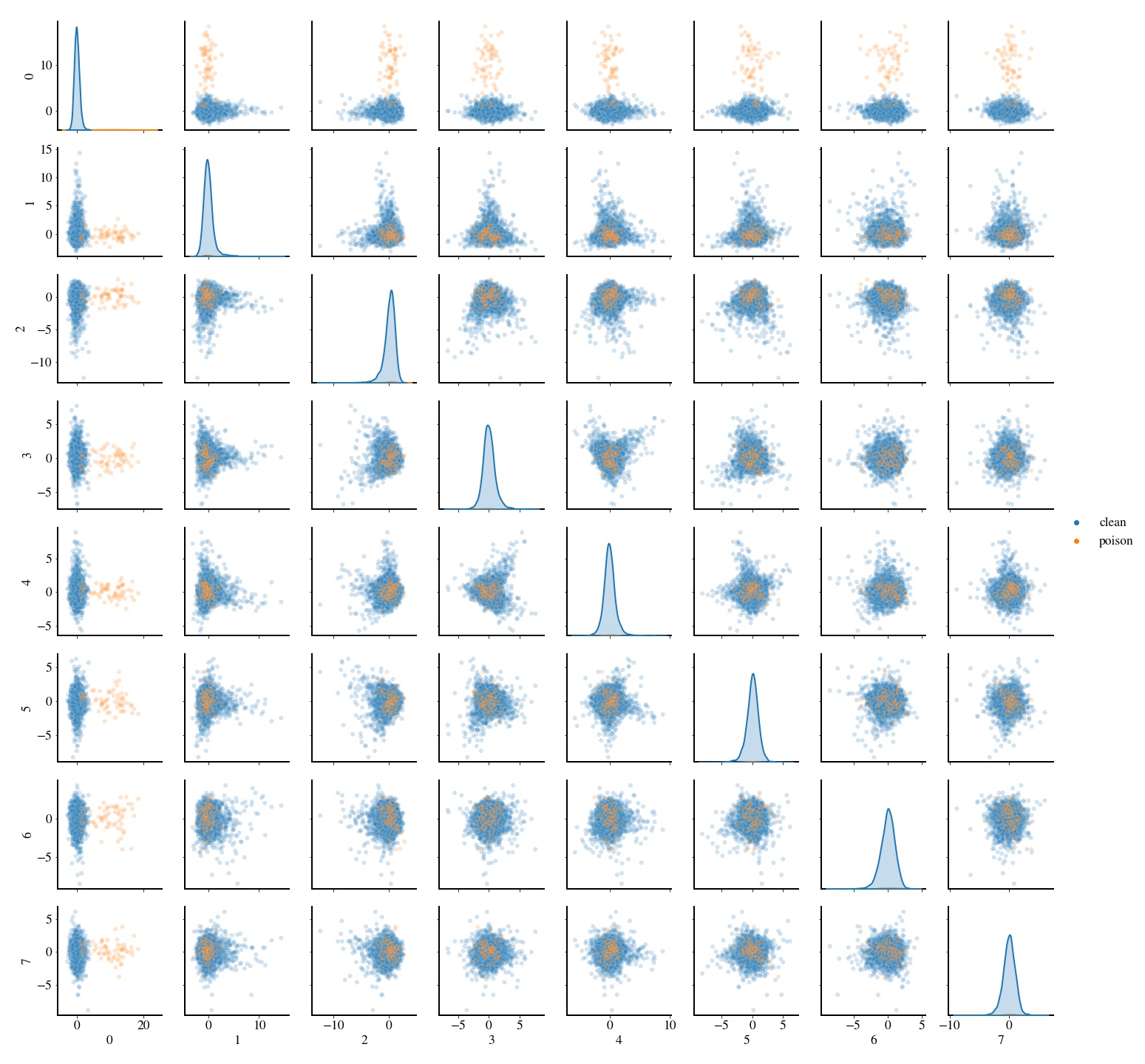}
  \caption{
    Scatter plots of the representations of the \(3\)-way pixel attack with \(\varepsilon = 0.0124\) after robust whitening; whitening the representations of the data with the  estimated covariance of the clean samples. 
    The whitened representations are projected onto their top eight PCA directions.
    Plots along the diagonal are Gaussian kernel density estimate plots after projecting onto that PCA direction.
    Off-diagonal plots are scatter plots of the data projected onto the subspace spanned by the corresponding pair of PCA directions.
    This clearly shows that the top PCA direction is aligned with the direction of separation between the poisoned samples (in orange) and clean samples (in blue), hence the squared projected norm scoring works. However, the variance of the poisoned examples are generally smaller, making it hard to distinguish using the  squared norm scoring. 
  }\label{fig:que-pairplot-3xp62}
\end{figure*}

In our experiments, squared norm scoring \(\tau_i^{(0)}=\norm{\tilde{\bm h}_i}^2\) fails for the \(3\)-way attack with \(\varepsilon = 0.0124\).
% if the poisoned examples are clustered together and have a small covariance.
% For the \(3\)-way attack with \(\varepsilon = 0.0124\), this can be seen in \cref{fig:que-svdvals-3xp62-10,fig:que-svdvals-3xp62-100}.
For this attack, the poisoned representations have high variance along a single direction, and relatively low variance along all other directions, as seen in \cref{fig:que-svdvals-3xp62-10}.
Because there are few poisoned examples relative to clean ones, the resulting spectral signature of the poisoned examples is weak.
The directions where the variance of the clean data was amplified, as seen in \cref{fig:que-svdvals-3xp62-100}, dominate all but one of the directions where the poison had high variance.
This can be seen in \cref{fig:que-pairplot-3xp62}, where only the top PCA direction, which corresponds to projected norm scoring, is suitable for removing the poisoned examples.
Using the squared norm scoring \(\tau_i^{(0)}=\norm{\tilde{\bm h}_i}^2\) here causes the top PCA direction to be mixed with the less useful directions, diluting its utility as a metric for removing the poison.

\begin{figure*}[h]
    \centering
    \includegraphics[width=\textwidth, trim=0 1em 0 1em]{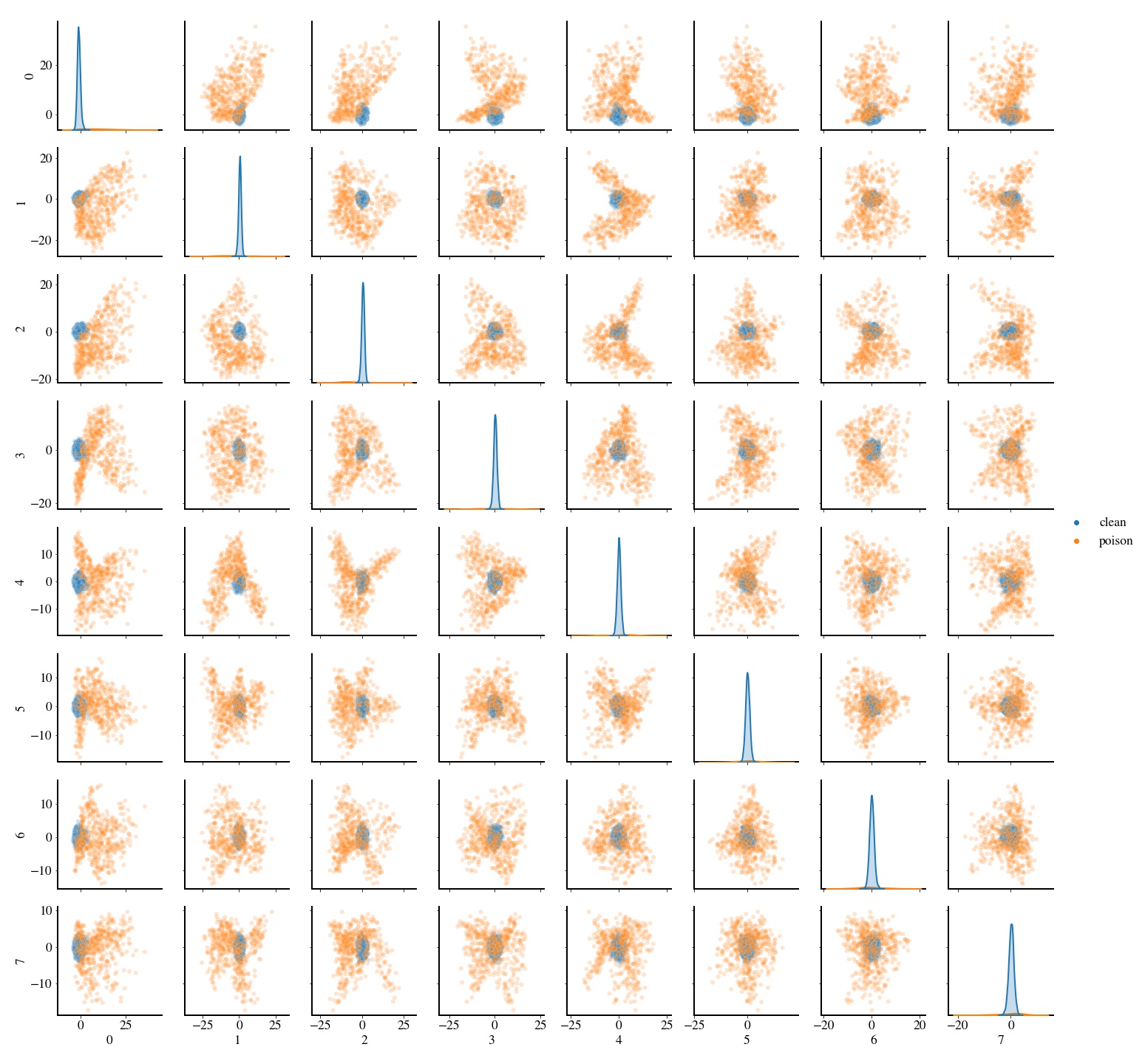}
    \caption{
      Scatter plots of the representations of the \(1\)-way pixel attack with \(\varepsilon = 0.1\) after robust whitening; whitening the representations of the data with the  estimated covariance of the clean samples. 
      The whitened representations are projected onto their top eight PCA directions.
      Plots along the diagonal are Gaussian kernel density estimate plots after projecting onto that PCA direction.
      Off-diagonal plots are scatter plots of the data projected onto the subspace spanned by the pair of PCA directions. This clearly shows that the top PCA direction is not aligned with the direction of separation between the poisoned samples (in orange) and clean samples (in blue), hence the squared projected norm scoring does not works. However, the variance of the poisoned examples are generally larger, making it easy to distinguish using the squared norm scoring.
    }\label{fig:que-pairplot-1xp500}
\end{figure*}

Squared projected norm scoring \(\tau_i^{(\infty)}=\abs{\inner{ v,\tilde{\bm h}_i}}^2\) fails for the \(1\)-way attack with \(\varepsilon = 0.1\).
Here the spectral signature of the poisoned examples is very strong.
The poisoned examples have high variance along many directions, as seen in \cref{fig:que-svdvals-1xp500-10}.
The resulting top PCA direction \(\bm v\) is not well aligned with the direction of the separation \(\bm\mu\del{S_{\mathrm{poison}}} - \bm\mu\del{S_{\mathrm{clean}}}\).
In fact, the angle between them is \(\cos^{-1}\del{\inner{\bm v, \bm\mu\del{S_{\mathrm{poison}}} - \bm\mu\del{S_{\mathrm{clean}}}}/\norm{\bm\mu\del{S_{\mathrm{poison}}} - \bm\mu\del{S_{\mathrm{clean}}}}} = \SI{35.7}{\degree}\).
The consequence of this misalignment can be seen in \cref{fig:que-pairplot-1xp500}, where it is clear that \(\bm v\) does not separate the poisoned examples from the clean ones.
On the other hand, squared norm scoring works well here because the poisoned examples have large variance along many directions, which is apparent in \cref{fig:que-pairplot-1xp500}.

\FloatBarrier
\section{Sensitivity to number of removed examples}\label{sec:limit2-sensitivity}

\begin{figure}[h]
\centering
\includegraphics[width=.5\linewidth, trim=0 1em 0 0em]{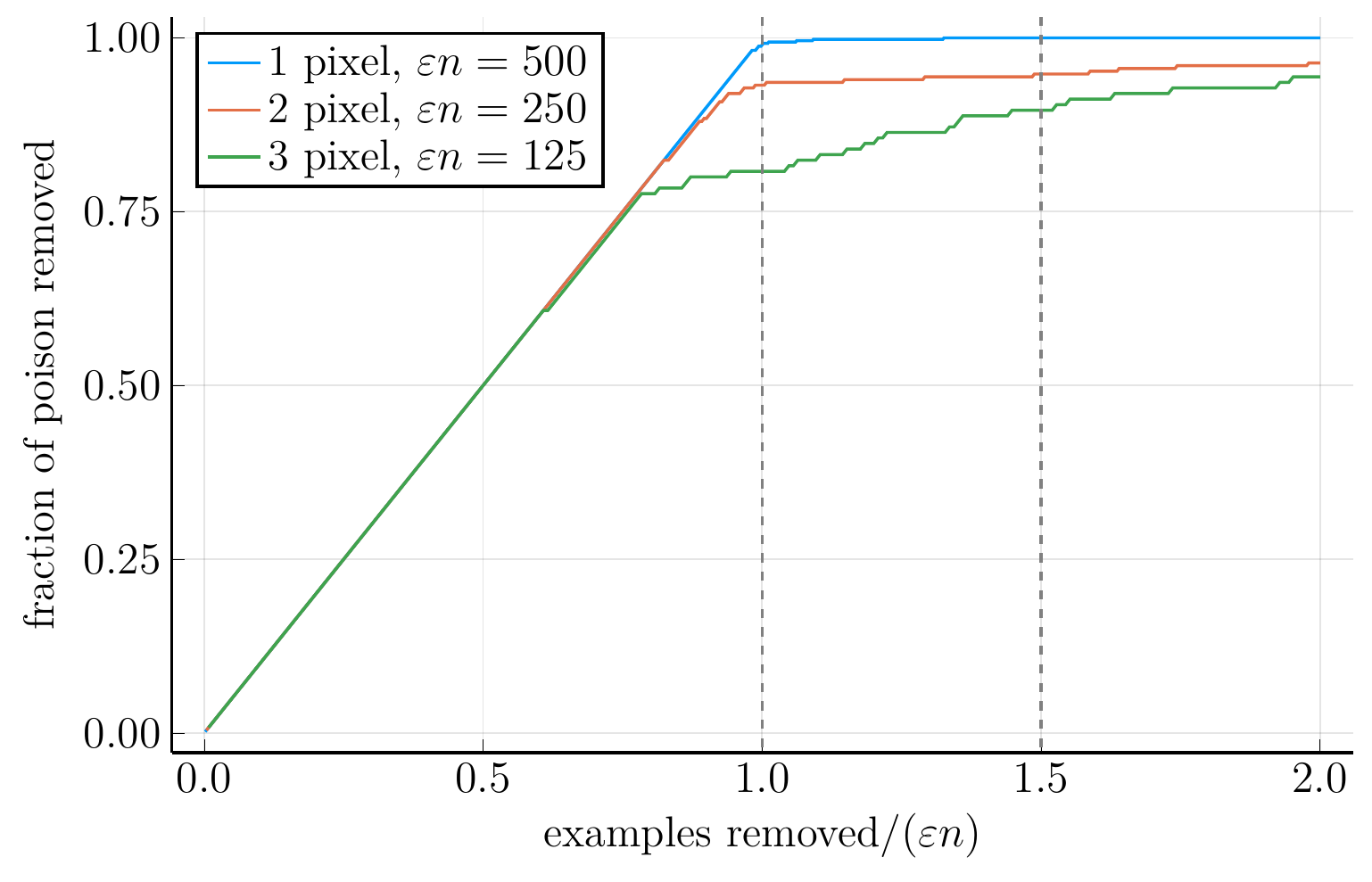}
\caption{
  Fraction of all poisoned samples removed vs. the total number of samples removed by SPECTRE, for three pixel attacks featuring spectral signatures of varying strength.
  % The number of poisoned samples removed is not too sensitive to the choice of how many total suspected samples to remove, for all three attack with strong spectral signature to weak spectral signature.
}\label{fig:limit2-sensitivity}
\end{figure}

Following \cite{tran2018spectral}, we choose to remove the \(1.5\varepsilon n\) samples with the highest QUE scores from the \(\del{1 + \varepsilon} n\) total samples bearing the target label.
% \footnote{There are only \(n\) samples bearing the target label in the label consistent attack setting.}
We show in \cref{fig:limit2-sensitivity} that our defence performance is not overly sensitive to this choice.
In particular, the fraction of poisoned samples removed does not vary substantially with the total number of removed samples after the first \(\varepsilon n\) samples are removed.
% For example, we could choose to remove $\varepsilon n$ or $2 \varepsilon n$ and the attack accuracy of the retrained model is still close to zero for all three attacks shown below (which span all the regimes of strong to weak spectral signatures).

\end{document}